%% file: main.tex
\documentclass[10pt]{article}

\usepackage[accepted]{tmlr}

\input{math_commands.tex}

\usepackage{booktabs}       
\usepackage{nicefrac}       
\usepackage{microtype}      
\usepackage{graphicx}
\usepackage{doi}
\usepackage{wrapfig}
\usepackage{float} 
\usepackage{import}
\usepackage{caption}
\usepackage{subcaption} 
\usepackage{adjustbox}
\usepackage{comment}
\usepackage{acronym}
\usepackage{xspace}
\usepackage{array}
\usepackage[mathscr]{eucal}
\usepackage{multirow}
\usepackage{amsmath,amssymb,amsfonts}
\usepackage{multicol}
\usepackage{algorithm}
\usepackage{algpseudocode}
\usepackage{multirow}
\usepackage{changepage}
\usepackage{enumitem}
\usepackage{lscape}
\usepackage{tabularx} 
\usepackage{mathtools}

\newcommand{\bluecomment}[1]{\textcolor{blue}{\Comment{#1}}}

\newcommand*{\eg}		{e.g.,\ }
\newcommand*{\ie}		{i.e.,\ }

\acrodef{ML} 		[\textsc{ML\xspace}]				{Machine Learning}
\acrodef{DL} 		[\textsc{DL\xspace}]				{Deep Learning}
\acrodef{ANN} 		[\textsc{ANN\xspace}]				{Artificial Neural Network}
\acrodef{DNN} 		[\textsc{DNN\xspace}]				{Deep Neural Network}
\acrodef{DNNs} 		[\textsc{DNNs\xspace}]				{Deep Neural Networks}
\acrodef{CNNs} 		[\textsc{CNNs\xspace}]				{Convolutional Neural Networks}
\acrodef{CNN} 		[\textsc{CNN\xspace}]				{Convolutional Neural Network}
\acrodef{MTL} 		[\textsc{MTL\xspace}]				{Multi-Task Learning}
\acrodef{RL} 		[\textsc{RL\xspace}]				{Reinforcement Learning}
\acrodef{LL} 		[\textsc{LL\xspace}]				{Lifelong Learning}
\acrodef{NLP} 		[\textsc{NLP\xspace}]				{Natural Language Processing}
\acrodef{LSTM} 		[\textsc{LSTM\xspace}]				{Long Short-Term Memory}
\acrodef{MAML} 		[\textsc{MAML\xspace}]				{Model Agnostic Meta Learning}
\acrodef{GAN}   	[\textsc{GANs\xspace}]	            {Generative Adversarial Nets}
\acrodef{VAE}   	[\textsc{VAEs\xspace}]	            {Variational Auto-Encoders}
\acrodef{NAS}   	[\textsc{NAS\xspace}]	            {Neural Architecture Search}

\usepackage{hyperref}

\title{Meta-Sparsity: Learning Optimal Sparse Structures in Multi-task Networks through Meta-learning}

\author{\name Richa Upadhyay\email richa.upadhyay@ltu.se \\
      \addr Luleå University of Technology, Sweden
      \AND
      \name Ronald Phlypo \email ronald.phlypo@grenoble-inp.fr \\
      \addr University Grenoble Alpes, France
      \AND
      \name Rajkumar Saini \email rajkumar.saini@ltu.se\\
      \addr Luleå University of Technology, Sweden
      \AND
      \name Marcus Liwicki \email marcus.liwicki@ltu.se\\
      \addr Luleå University of Technology, Sweden}

\begin{document}
 
\maketitle

\begin{abstract}
\input{sections/0_abstract}
\end{abstract}

  

\section{Introduction} \label{sec:intro}
\input{sections/1_introduction}

\section{Related work} \label{sec:related_work}
\input{sections/2_related_work}

\section{Methodology} \label{sec:method}
\input{sections/3_methodoloy}

\section{Experimental setup} \label{sec:exp_setup}
\input{sections/4_expsetup_results}

\section{Results and Discussion} \label{sec:discussion}
\input{sections/5_discussion}

\section{Conclusion and future scope} \label{sec:conclusion}
\input{sections/7_conclusion}



\subsubsection*{Acknowledgments}
The authors would like to acknowledge the computational resources provided by the National Supercomputer Centre at Linköping University, particularly the Berzelius supercomputing system, supported by the Knut and Alice Wallenberg Foundation. The authors also wish to thank the reviewers for their invaluable feedback and suggestions, which have significantly contributed to improving the quality of this paper.

\bibliography{ref}
\bibliographystyle{tmlr}

\section{Appendix}\label{appendix}
\input{sections/8_appendix}

\end{document}

%% file: math_commands.tex
\usepackage{amsmath,amsfonts,bm}

\def\eqref#1{equation~\ref{#1}}

\def\1{\bm{1}}

\DeclareMathAlphabet{\mathsfit}{\encodingdefault}{\sfdefault}{m}{sl}
\SetMathAlphabet{\mathsfit}{bold}{\encodingdefault}{\sfdefault}{bx}{n}

\DeclareMathOperator*{\argmin}{arg\,min}

%% file: sections/0_abstract.tex
This paper presents meta-sparsity, a framework for learning model sparsity, basically learning the parameter that controls the degree of sparsity, that allows deep neural networks (DNNs) to inherently generate optimal sparse shared structures in multi-task learning (MTL) setting. 
This proposed approach enables the dynamic learning of sparsity patterns across a variety of tasks, unlike traditional sparsity methods that rely heavily on manual hyperparameter tuning. 
Inspired by Model Agnostic Meta-Learning (MAML), the emphasis is on learning shared and optimally sparse parameters in multi-task scenarios by implementing a penalty-based, channel-wise structured sparsity during the meta-training phase. 
This method improves the model’s efficacy by removing unnecessary parameters and enhances its ability to handle both seen and previously unseen tasks.
The effectiveness of meta-sparsity is rigorously evaluated by extensive experiments on two datasets, NYU-v2 and CelebAMask-HQ, covering a broad spectrum of tasks ranging from pixel-level to image-level predictions.
The results show that the proposed approach performs well across many tasks, indicating its potential as a versatile tool for creating efficient and adaptable sparse neural networks. 
This work, therefore, presents an approach towards learning sparsity, contributing to the efforts in the field of sparse neural networks and suggesting new directions for research towards parsimonious models.

%% file: sections/1_introduction.tex
Model compression in \ac{DL} is a process that aims at reducing the size and complexity of neural network models while maintaining their performance. 
In the current technological trend, model compression has become essential in \ac{DL} due to its role in enabling complex neural networks to operate efficiently on devices with constrained resources by substantially improving memory usage and computation requirements. 
It facilitates the practical application of \ac{DL} models in real-world scenarios, fostering sustainability and enhancing accessibility on a broader scale (\cite{9043731}).
It is possible to reduce the size and complexity of a model by leveraging techniques such as Neural Architecture Search (NAS)(\cite{ren2021comprehensive}), model value quantization, model distillation (\cite{hinton2015distilling}), low-rank factorization (\cite{sainath2013low}), parameter sharing (\cite{desai2023defense}), sparsification (\cite{review_sparsity}), and many more. 
This work focuses on two of these techniques, \ie parameter sharing in the form of hard parameter sharing in multi-task learning and sparsification. 
However, our emphasis extends beyond simply compressing models.
We combine the concept of model sparsification with \ac{MTL}, viewing sparsification as a technique for selecting optimally shared features in a multi-task setting.
This methodology enables the strategic distribution of these features across diverse tasks during \ac{MTL}, enhancing the joint learning process.

\begin{figure}[ht]
    \centering
    \includegraphics[width = 0.9\linewidth]{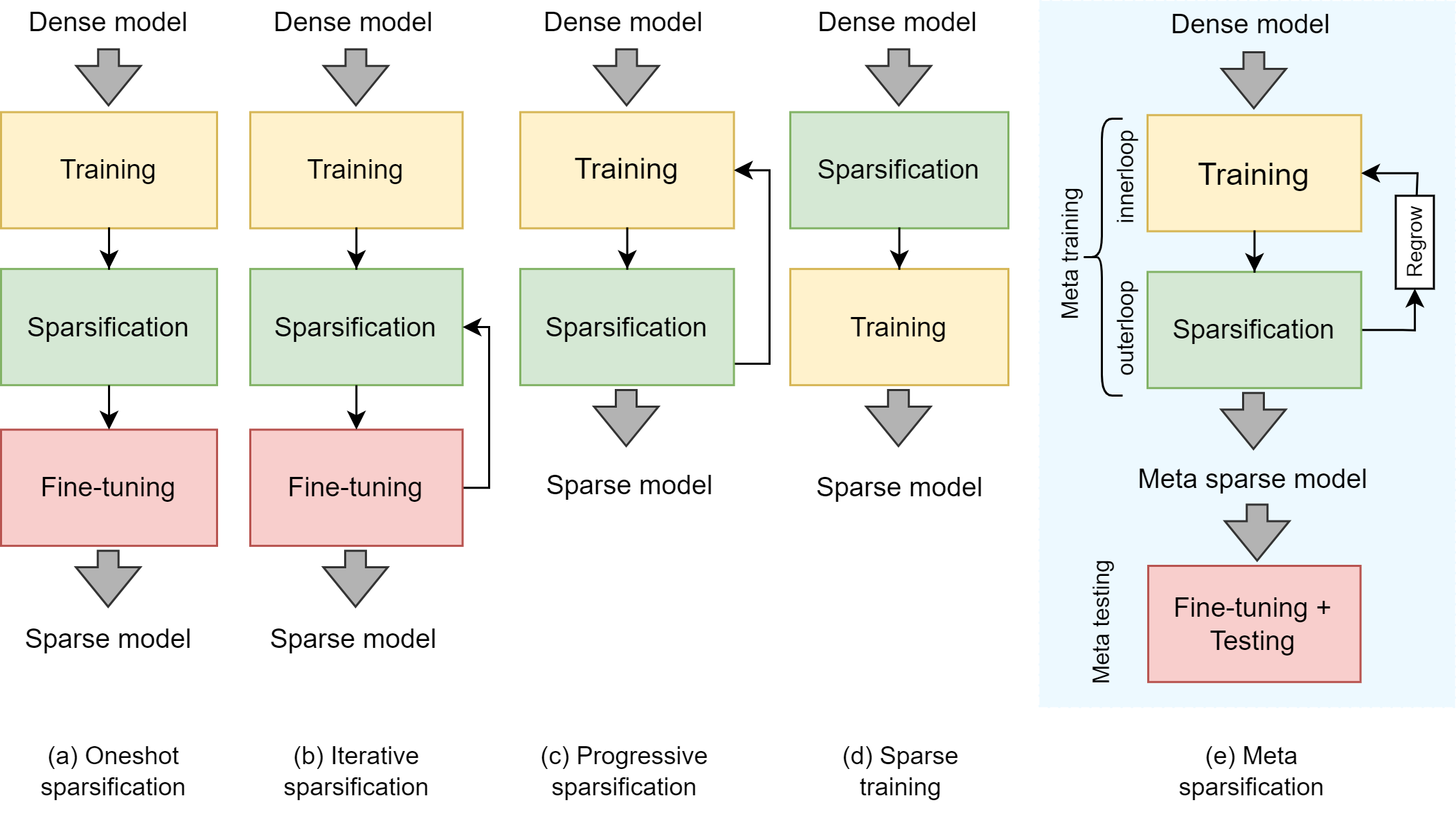}  
    \caption{This figure illustrates (a-d) a few common approaches to achieving sparse models and (e) the proposed approach of meta-sparsification.}
\label{fig:intro_fig}
\end{figure}

Sparsification\footnote{\textit{Note:}  
In the context of this work, \textit{sparsification} should not be confused with \textit{pruning}, which is also a widely used term in model compression.
This is because both are different concepts; pruning is a technique that can help to achieve sparsification by systematically removing certain elements; sparsification is a much broader concept that can be reached through various methods, including but not limited to pruning. 
So the words sparsification and pruning are \textbf{not} used interchangeably.} 
in \ac{DL} pertains to transforming a densely connected neural network or a dense model to a sparse one, where a considerable proportion of the parameters (typically model weights) are set to zero. 
This reduction of active weights leads to less computational load and memory usage during both the training and inference phases.
Additionally, sparsification not only highlights the importance of key features (sparsity-driven feature selection) but also significantly enhances the model's ability to generalize to new data (\cite{review_sparsity}).
When implementing sparsity in \ac{DNN}, three key aspects must be considered: (i) identifying what elements should be made sparse, (ii) determining when to induce sparsity, and (iii) deciding on the method \ie how to achieve sparsity.
The article by \cite{review_sparsity} offers a comprehensive survey that thoroughly explores these factors related to sparsity, among various other elements.
Several elements can be sparsified in a \ac{DNN}, like the parameters (or weights), neurons, filters in convolution layers, and heads in attention layers.
Sparsification can be scheduled post-training, during training, or as sparse training, as illustrated in Figure~\ref{fig:intro_fig}. 
Further details on these methods are provided in the subsequent text.
Various techniques can be employed to attain sparsity, including methods like magnitude pruning based on thresholds, sparsification based on input-output sensitivity, penalty-driven approaches, variational methods, and many more.
In this work, we emphasize sparsifying the model parameters during the training by applying penalty-based (regularization-based) sparsification methods.

Various sparsity-inducing strategies have been developed to facilitate the transformation of a dense model into a sparse model (\cite{review_sparsity}). 
However, a vital question that arises before implementing these strategies is determining when and how to apply sparsification. 
Figure~\ref{fig:intro_fig} broadly illustrates some of these sparsity scheduling techniques as discussed by \cite{hubens2020pruning}. 
The one-shot sparsification (Figure~\ref{fig:intro_fig}(a)) is a fundamental technique that sparsifies an already trained dense model until a desired sparsity is achieved and then fine-tunes the sparse model. 
Fine-tuning allows the model to relearn.
It helps to re-distribute the weights across the remaining parts of the network, ensuring that performance does not suffer significantly.
Figure~\ref{fig:intro_fig}(b) shows iterative sparsification where sparsity is induced gradually over many steps (\cite{han2015learning, liu2017learning}). 
Only a handful of weights are eliminated at each iteration, and the network is fine-tuned to restore performance.
This process is repeated multiple times.
\cite{zhu2017prune} presented the concept of automated progressive (or gradual) sparsification (Figure~\ref{fig:intro_fig}(c)) in their work, which is very similar to iterative sparsification.
While both methods incrementally introduce sparsity into a dense network, automated gradual sparsification is different because it sparsifies the weights during the training of the model, allegedly claiming that it eliminates the necessity for subsequent fine-tuning.
Figure~\ref{fig:intro_fig}(d) illustrates sparse training, where a sparse model is trained, and while training, it tries to retain the sparsity. 
This is often referred to as sparse-to-sparse training (\cite{dettmers2019sparse, evci2020rigging}), while the others Figure~\ref{fig:intro_fig}(a-c) are referred to as dense to sparse training.

In general, any sparsification involves the use of a few hyper-parameters that control the degree of sparsity, like thresholds in magnitude pruning, sparsification budget (or target), the regularization parameter in penalty-based sparsity approach, the dropout rate (probability), learning rate in gradual pruning, and several others. 
Often, approaches like grid search, random search, heuristic methods, etc, are employed to determine the optimal values of these crucial hyper-parameters.
We introduce a method to learn these hyper-parameters that induce sparsity through meta-learning. 
This work, therefore, proposes meta-sparsification, broadly depicted in Figure~\ref{fig:intro_fig}(e).
In this work, we refer to it as \textit{learning to learn sparsity}, or \textit{learned sparsity}, or \textit{meta-sparsity}.
To be more precise, it is learning to learn or meta-learning (\cite{pmlr-v70-finn17a}) the sparsity controlling hyper-parameter, thereby learning the sparsity pattern.

We begin with a dense multi-task model, which shares some layers \ie a shared encoder or backbone between many tasks.
To achieve a clear understanding, we begin by answering the three essential questions related to sparsification.
(i) \textit{What to sparsify?} - we aim to sparsify the parameters or weights of the model, primarily the parameters of the shared layers of a multi-task network. 
We apply channel-wise structured sparsity to the shared layers to achieve optimal shared representations across all the tasks. 
Structured sparsity is a coarse-grained approach that considers the architecture of the dense network while sparsifying \ie for a \ac{CNN}; it involves zeroing out entire filters, neurons, or channels, basically a structured block of parameters.
(ii)\textit{how to achieve sparsity?} (\ie the method) - we apply penalty-based (regularization-based) structured sparsity to the backbone of a multi-task network.
The sparsification is applied in a setting inspired by \textit{ gradient-based meta-learning} \eg \ac{MAML} (\cite{pmlr-v70-finn17a}) to learn the meta-sparsity patterns across all the tasks.
Meta-learning (\cite{thrun2012learning, huisman2021survey}), commonly referred to as `learning to learn,' involves developing algorithms that learn from a variety of tasks and then apply the accumulated knowledge to enhance future learning of new or similar tasks.
(iii) \textit{when to sparsify?} - sparsity is applied while training the network; this is also known as \textit{dynamic sparsity}.
It refers to an approach where the sparsity pattern\footnote{\textit{sparsity pattern} refers to the specific arrangement or layout of non-zero parameters within a model architecture, for example, a \ac{CNN} when structured sparsity is applied. In the context of channel-wise structured sparsity, this pattern identifies the active (non-zero) channels within the network layers from the zeroed-out (inactive) ones. These patterns highlight the selective engagement of certain parts of the network while others remain dormant.} changes over time during training, adapting to the evolving data and training requirements, while static sparsity refers to fixed sparsity patterns throughout training. 
The broader objective of this work is to explore learning the sparsity patterns across multiple learning episodes or tasks and to develop a meta-sparse model during the meta-training stage that can potentially be fine-tuned for the same tasks or unseen tasks in the meta-testing stage (as shown in Figure~\ref{fig:intro_fig}(e)).

Now, the question is why this approach is applied in a multi-task context.
Traditional \ac{MAML} often involves tasks or learning episodes that are very similar or homogeneous; for example, all are image classification tasks (\cite{meta_timothy, paper1}). 
Because of this similarity, tasks may exhibit consistent sparsity patterns, hindering the meta-model's ability to accommodate a wider variety of tasks with different sparsity requirements.
\ac{MTL}, in contrast, enables the simultaneous learning of multiple, diverse tasks. 
Therefore, we aim to identify robust sparsity patterns that are suitable for various diverse tasks by using sparsity in multi-task scenarios and learning sparse behavior across tasks through meta-learning.
The main contributions of this work are as follows -
\begin{itemize}
    \item \textit{Learned sparsity framework called Meta-sparsity}: We introduce a novel approach that allows for the dynamic learning of sparsity in \ac{DNN}. This is a significant shift from heuristic-based sparsity control to a more dynamic, task-aware method. 
    Our method is versatile, being agnostic to model type, task, and sparsity type, allowing for wide applicability.
    \item \textit{Comprehensive evaluation:} We rigorously assess our approach across a variety of tasks using two publicly available datasets, showcasing its effectiveness. 
    To provide comparative insights, we compare these results to those obtained from models using no sparsity and fixed-sparsity (\ie the fixed value of sparsity hyper-parameter) in single-task and multi-task settings.     
    \item \textit{Robustness validation:} The robustness of our meta-sparse models is suggested by their ability to perform reasonably well on previously unseen tasks during meta-testing, showcasing their potential for adaptability.
    \item \textit{Versatility in sparsity types:} Although this work focuses on channel-wise structured sparsity, we also validate the efficacy of our approach on unstructured sparsity, demonstrating its broad utility.
    \item \textit{Direction for future research:} We conclude with a discussion on potential future directions and open questions to encourage more investigation in the areas of model sparsity. 
\end{itemize}

This paper is structured as follows: Section 2 delves into the related work, setting the context and background for this study.
Section 3 describes the methodology of the proposed approach.
Section 4 details the experimental setup, while Section 5 presents the results, performance analysis, and a thorough discussion of the findings.
Finally, Section 6 concludes the work and suggests directions for future research.

%% file: sections/2_related_work.tex
This section focuses on research works that have employed the notion of learning diverse entities (\ie hyper-parameters, optimizers, loss functions, and many more) through the utilization of black-box models or alternative methodologies. 
We additionally delve into the utilization of sparsity in deep learning and subsequently focus our attention on the application of sparsity in \ac{MTL} networks.  
Finally, we position our work within the existing literature. 

\textbf{What can be learned?:} 
The concept of learning to learn, commonly known as meta-learning, has been a learning paradigm of great interest in research for many years.
This field has significantly advanced since the seminal works of \cite{schmidhuber:1987:srl, 155621, thrun2012learning}, leading to a wide range of applications.
These algorithms aim to achieve generalization by learning from experiences, with the extent of generalization depending on the accumulation of meta-knowledge.
In simple words, this generalization can be achieved by learning to learn parameters, hyper-parameters, loss functions, architectures, optimizers, and many more, which constitutes meta-knowledge.  
Many excellent studies have been conducted in this field of study; in order to set a context for our work, we will attempt to highlight and briefly discuss some of them. 

We will first begin with \textit{learning to learn parameter initialization}. 
Transfer learning centers on the concept of providing generalized initial parameters for the downstream task, enabling the fine-tuning of the new task with relatively fewer data samples. 
A possible approach to find the best initial parameters is to utilize meta-learning algorithms to train the parameters across different tasks, acquiring meta-knowledge that can aid in rapid adaption to new tasks.  
The article by \cite{pmlr-v70-finn17a} presented \ac{MAML} an algorithm that optimizes its parameters to facilitate adaptability and quick learning across a diverse range of tasks by providing a set of initial parameters.
Reptile (\cite{nichol2018reptile}) is also a similar algorithm that is mathematically similar to first-order \ac{MAML}, that learns the initialization of a network.  
First-order MAML (FOMAML) (\cite{nichol2018first}) and Almost No Inner Loop (ANIL) (\cite{raghu2019rapid}) are simplifications of \ac{MAML} that provide computational advantages compared to \ac{MAML}. 
The comprehensive survey by \cite{huisman2021survey} provides detailed insights into other related work within the field of meta-learning.

While training neural networks, one of the most tedious tasks is hyperparameter tuning. 
This is because it involves a trial-and-error process over a vast, complex search space, and often, the hyperparameters are very use-case specific; there is no guarantee of finding the best solution, and each trial can be computationally expensive and time-consuming.   
That is why \textit{learning to learn hyperparameters} is critical. 
Some works by \cite{li2017meta, xiong2022learning, meta_LR, subramanian2023learned} focused on learning the learning rate or learning the learning rate schedules to train a deep learning model. 
Many articles learn to adapt all the hyperparameters, including the learning rate. 
For example, the article by \cite{baik2020meta} adaptively generated per-step hyperparameters to improve the performance of the model. 
\cite{NEURIPS2021_bac49b87} presented an approach for hyperparameter optimization by leveraging evolutionary strategies to estimate meta-gradients.
Another similar work by \cite{franceschi2018bilevel} offered a structured bi-level programming approach where the outer level updated hyper-parameters while the inner loop focused on task-specific learning or loss minimization. 
The approach presented by \cite{franceschi2018bilevel} closely aligns with our work, with the primary distinction being our specific emphasis on learning structured sparsity within the framework of Multi-Task Learning (MTL) by employing \ac{MAML}.

The articles by \cite{wortsman2019learning, gao2021searching, bechtle2021meta, gao2022loss, raymond2023learning, raymond2023online} aim to learn parametric loss functions, thereby \textit{learning to learn a loss function}. 
On similar grounds, several works focus on \textit{learned optimizers} or black-box or parametric optimizers, like the research by \cite{155621, andrychowicz2016learning, wichrowska2017learned, lv2017learning, li2017learning, shen2020learning, harrison2022closer, metz2022velo, metz2020tasks, gao2022meta}.
The learning-to-learn concept is also used in the field of neural architecture search (\cite{elsken2019neural}) to achieve optimal architectures. 
This idea of \textit{learned architectures} is discussed by \cite{lian2019towards, shaw2019meta, elsken2020meta, Ding_2022_CVPR, rong2023across, schwarz2022metalearning}.
Numerous studies in existing literature apply learning to learn concepts across various domains, such as reinforcement learning, attention learning, and neural memory learning, among others.
However, this paper will not delve into each of these applications, as it is not intended to be a comprehensive review of the field.
Instead, our primary objective is to highlight distinct insights and developments within a more limited domain of learning sparsity.
Overall, it can be summarized that the black-box or parameterized models are increasingly replacing traditional white-box aspects of deep learning, such as loss functions, optimization algorithms, automated architecture search, and others. 
Building on this concept, this paper presents an approach to \textit{learning to learn sparsity} using meta-learning.

Sparsification is one of the approaches for model compression in Deep Neural Networks (DNNs).
It does not just reduce the complexity of the model but also leads to significant gains in performance, computation, and energy efficiency, all while strategically selecting key features that contribute to these gains.
\cite{review_sparsity} presented an elaborate study of sparsity in deep learning. 
They give an extended survey of works in this domain based on the types of sparsity, what can be sparsified, when to sparsify, and how to introduce sparsity.
Some other works that discuss model sparsity or pruning are \cite{zhu2017prune, gale2019state}.
Our work is in line with the area of \textit{learned sparsity}, in contrast to works achieving sparsity using fixed hyper-parameters. 
Most of the works in the domain focus on adaptive weight or parameter pruning (\cite{han2015learning}), a technique to induce sparsity in a network. 
\cite{pmlr-v119-kusupati20a} proposed a dynamic sparsity parameterization (STR) technique that used the sparse threshold operator to achieve sparsity in \ac{DNN} weights by learning layer-specific pruning thresholds. 
Unlike the static (fixed) pruning methods, where the sparsity hyperparameter is fixed once pruning is performed, this method dynamically adjusts the sparsity hyperparameter during training, thereby determining the optimal sparsity pattern.

Another similar work by \cite{8578988} introduced the learning-compression method, a two-stage process used by the algorithm to prune the network.
In the learning phase, the weights are optimized \eg reducing loss, while in the compression phase, network pruning is done using $l_0$ or $l_1$ constrain along with fixed pruning hyperparameters. 
Instead of manually adjusting the pruning rates of each layer, \cite{zhou2021effective} introduced ProbMask, a technique that leverages probability to determine the importance of weights across all layers and allows automatic learning of weight redundancy levels based on a global sparsity constraint. 
A combination of structured and unstructured sparsity is employed by \cite{zhou2021learning}, resulting in N:M fine-grained sparsity wherein each group of M consecutive weights of the network, there are at most N weights that have non-zero values.
So, this method does weight pruning based on the N:M budget.
Many other works focus on adaptive pruning of networks or learning structured sparsity (\cite{wen2016learning, meng2020pruning, liu2018rethinking, lee2021layeradaptive, wang2021neural, paper3}); however, none have applied meta-learning for learning to learn parameter sparsity.
The articles contributed by \cite{wen2016learning, meng2020pruning, deleu2021structured} provide comprehensive insights into the concept of structured sparsity in deep neural networks.
The idea of structured sparsity constitutes an essential aspect of our paper, and we draw significant inspiration from these works.

\textbf{Sparsity in multi-task learning: }
The domain of Multi-Task Learning (MTL) confronts two significant challenges. 
Firstly, as the number of tasks increases, the total number of trainable parameters also increases.
Secondly, it is crucial to efficiently group similar tasks or effectively share features across tasks during training to prevent adverse information transfer among tasks, which could potentially impact the performance of certain tasks. 
Sparsity is used in the literature as one of the solutions to overcome these problems in multi-task settings. 
For example, \cite{kshirsagar2017learning} constructed a regularization term to jointly optimize the task-specific parameters while also learning shared group structures (or parameters). 
Similarly, \cite{JMLR:v17:15-215} in their work introduce Multi-task Sparse Structure Learning (MSSL) for joint estimation of per-task parameters and their relationship structure parameters using Alternating Direction Method of Multipliers (ADMM) algorithm (\cite{boyd2011distributed}).
\cite{argyriou2006multi}, also assumed that tasks are related and used $l_1-l_2$ regularization with the combined loss for efficient multi-task feature learning. 
\cite{obozinski2010joint, 10.1145/1553374.1553392} used different sparse and non-sparse regularizations to learn the low-dimensional subspace, which is shared by all the tasks.
\cite{sun2020learning} proposed a method for iterative magnitude pruning of multi-task model parameters, which was inspired by the lottery ticket hypothesis (\cite{frankle2018the}).
Their method induces unstructured $l_1$ type sparsity (sparse masks for weight matrices) by training sub-nets for multiple tasks until convergence.

\textbf{Positioning our work: } In the literature stated earlier,  we looked into numerous ideas, such as learning to learn an optimizer, loss function, initial parameters, hyper-parameters, and many more.
Building upon these discussions, we introduce an approach that utilizes meta-learning to facilitate the process of \textit{learning to learn sparsity}.
From the view of \ac{MTL}, the previously mentioned studies induce sparsity in multi-task models by utilizing fixed values for regularization parameters, pruning thresholds, or sparsification budgets.
In contrast to employing a trial-and-search methodology for the fixed hyper-parameter, the present study emphasizes acquiring the ability to learn a generalized sparsity parameter across many tasks.
This approach can also be viewed as one of the strategies for optimal feature sharing among tasks within a multi-task framework facilitated by learned structured sparsity.
To the best of our knowledge, this research direction has not been previously explored, offering a unique contribution to the fields of sparsity and \ac{MTL}.



%% file: sections/3_methodoloy.tex
In this section, we outline our proposed methodology. 
First, we will look into the conventional \ac{MTL}, meta-learning with a focus on \ac{MAML}, and group (structured) sparsity in order to establish a strong foundation for \textit{meta sparsity}.
This foundational discussion sets the stage for understanding the nuances of our approach.
Furthermore, we also discuss a theoretical perspective of generalization in the context of this work.

\subsection{Multi-task Learning}\label{sec:mtl}

\begin{figure}[t]
    \centering
    \includegraphics[width = 0.9\linewidth]{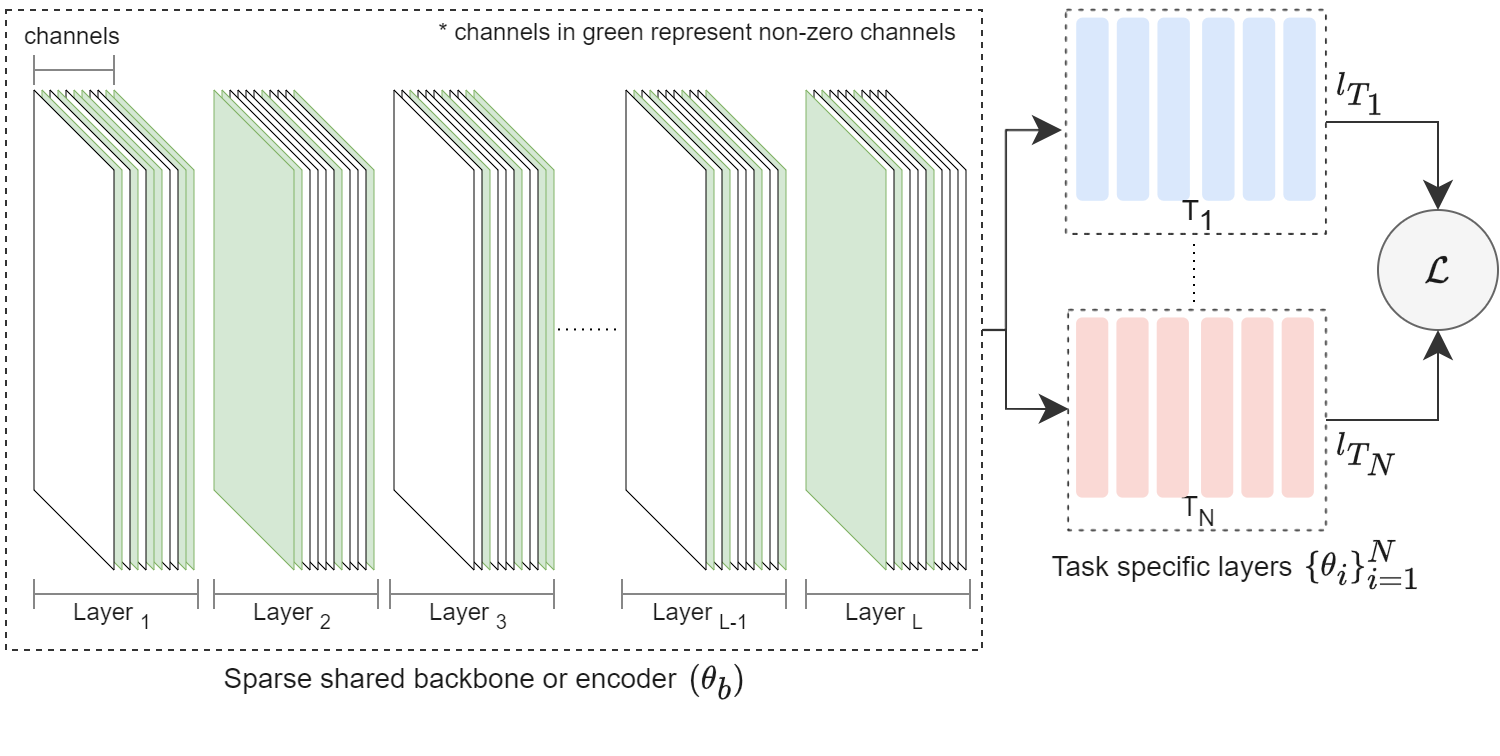}
    \caption{A schematic of the multi-task architecture used in this work (inspired by \cite{paper3}).}
    \label{fig:arch}
\end{figure}

Multi-task learning (\cite{caruana1997multitask}) is a very well-established learning paradigm wherein the aim is to jointly learn or train multiple related tasks.
The underlying theory here is that the tasks help each other to learn better due to inductive transfer between the tasks, leading to improved performance and better generalization. 
A successful \ac{MTL} can be achieved by establishing a balanced sharing between tasks, such that there is a positive transfer of information. 
This primarily depends on the parameter sharing approach used for \ac{MTL}, which, according to \cite{crawshaw2020multitask}, can be either hard parameter sharing or soft parameter sharing. 
Hard parameter sharing is a result of the shared architecture, while soft parameter sharing can be achieved by applying constraints on the model parameters.
These concepts, along with various research efforts in this area, are explored in a comprehensive survey on \ac{MTL} by \cite{crawshaw2020multitask}.
This work employs a very simple and conventional multi-task architecture, demonstrating hard-parameter sharing see Figure~\ref{fig:arch}.

Consider a task distribution $p(T)$, from which, say, $N$ non-identical yet related tasks are sampled that can be trained simultaneously in a multi-task setting, say, $\mathcal{T} = \{T_1, T_2, ..., T_N\}$. 
The objective of \ac{MTL} is to train a multi-task model, say $f(\Theta)$ so that a combined loss ($\mathcal{L}$) of the task-specific losses $\{l_{T_1}, l_{T_2}, \ldots, l_{T_N}\}$ is minimized, such that the optimal parameters are given by the Equation~\ref{eq:mtl}, 
\begin{equation}\label{eq:mtl}
    \Theta^* = \argmin_\Theta \mathcal{L}(f(\Theta)) = \argmin_{\theta_b, \theta_i \in \Theta}  \mathcal{F}_c(l_{T_i}(f_i(\theta_b,\theta_i))), ~~~~~ i = 1,2,\ldots N    
\end{equation}
where $\Theta = \{\theta_b\} \cup \bigcup_{i=1}^{N} \{\theta_i\}$, indicating that the set of multi-task parameters $\Theta$ comprises $\theta_b$,  the parameters of the shared architecture or backbone, and $\theta_i$  the parameters specific to each of the $N$ tasks. 
$f_i$ denotes the model designated for a particular task, consisting of a shared backbone ($\theta_b$) along with layers specific to the task $\theta_i$.
$\mathcal{F}_c$ represents the function that combines or, more precisely, balances the $N$ task-specific losses.
This function may be a straightforward sum of losses, a weighted average, or any other tailored function designed to suit the specific requirements of the use case.
To maintain fair learning across all tasks and to avoid any one task from dominating the learning process, loss balancing is essential in \ac{MTL}.
Several approaches to combine single-task losses to ease the multi-task optimization are discussed in \cite{crawshaw2020multitask}.

In this work, we employ the uncertainty weighting approach given by \cite{kendall2018multi}.
The core idea of this approach is to optimize learning outcomes in multi-task contexts by prioritizing or weighing the tasks based on their uncertainty.
It can mathematically be represented as,
\begin{equation}\label{eq:uncertain}
    \mathcal{L} = \mathcal{F}_c(l_{T_1}, l_{T_2}, \ldots, l_{T_N}) = \sum_{i = 1}^{N} \bigg( \frac{1}{2\sigma_{i}^2} \cdot l_{T_i} + \log \sigma_i \bigg) 
\end{equation}
where $\sigma_i$ is a learnable noise parameter for $i^{th}$ task.
This combined loss $\mathcal{L}$ undergoes backpropagation, where gradients of the loss are computed with respect to all model parameters $\Theta$, encompassing both shared and task-specific parameters. 
These gradients are then used to update $\Theta$ to minimize $\mathcal{L}$ by finding the optimized parameters, thereby enhancing the performance of all tasks.
Algorithm~\ref{algo:mtl} outlines the practical implementation of Multi-Task Learning (MTL).
\input{tables/mtl_algo}

\subsection{Meta-learning}
Meta-learning, often known as `learning to learn' (\cite{Thrun98, baxter1998theoretical}), is a learning algorithm that leverages past learning experiences to enhance the performance of a new task.
In conventional \ac{ML}, an algorithm is designed to acquire knowledge from a given dataset to perform a specific task.
However, meta-learning goes a step further; it involves algorithms that can evaluate their learning process, analyze their performance across various tasks, and leverage this insight to learn new tasks more efficiently.
Meta-learning algorithms can be classified into three categories, \ie (i)model-based, (ii)metric-based, and (iii)optimization-based or gradient-based meta-learning.
These classifications are thoroughly discussed in several survey papers, including works by  \cite{huisman2021survey, TIAN2022203, meta_timothy}, and many more, which provide in-depth analyses and comparisons of methodologies within each category.
Out of these, in this work, we focus on optimization or gradient-based meta-learning algorithms, particularly \ac{MAML} (\cite{pmlr-v70-finn17a}).
\ac{MAML} facilitates fast and efficient adaptation of a \ac{DNN} to new and distinct tasks, using only a limited amount of training data.
This is accomplished by optimizing a set of initial parameters that are highly adaptable and enable quick fine-tuning via minimal gradient updates.
This approach improves model generalizability and performance and has been thoroughly tested in a variety of settings (tasks), such as classification, regression, and reinforcement learning.
In this work, we extend its application beyond similar or homogeneous tasks to dense prediction tasks like segmentation, depth estimation, and others by incorporating \ac{MTL} in the form of multi-task learning episodes (\cite{paper2} of heterogeneous tasks \footnote{As defined by \cite{paper1}, tasks are considered homogeneous when they share similar objective, and if the objectives differ, they are classified as heterogeneous. For example, if the objective of all the tasks is image classification, they can be referred to as homogeneous tasks.}.

The \ac{MAML} framework works with many tasks, also defined as learning episodes, and progresses through two hierarchical levels (or loops): the inner loop, focusing on learning from individual tasks through rapid adaptation to training data, and the outer loop, which aggregates knowledge from numerous tasks to enhance and accelerate the adaptation process.
These levels are formalized in the form of a bilevel optimization problem in \ac{MAML} in the meta-training stage,  which is addressed hereafter.
After meta-training, the `learning to learn' stage, new tasks can be introduced in the meta-testing stage, also known as the `adaptation stage.'
Consider, for $N$ tasks or learning episodes sampled from task distribution $p(T)$, $\mathcal{L}_{meta}$ and $l_{{T}_i}$ be the meta (\ie outer) and $i^{th}$ task-specific (\ie inner) loss functions respectively.
Along the same lines, as formulated in Section~\ref{sec:mtl}, let $\Theta_{meta}$ be the meta parameters, while $\theta_i$ be the parameters for $i^{th}$ task.
For training, the dataset for $N$ tasks can be expressed as $\mathcal{D}= \{(D_{sup_{1}}, D_{query_{1}}),\ldots,(D_{sup_{N}}, D_{query_{N}})\}$, where $D_{sup}$ stands for support set and $D_{query}$ stands for the query set such that $D_{sup_{i}} \cap D_{query_{i}} = \emptyset$.
The bi-level optimization in \ac{MAML}, where one optimization problem involves another as a constraint (\cite{meta_timothy, huisman2021survey, pmlr-v70-finn17a}), can be written as follows :
\begin{equation}\label{eq:meta1}
    \Theta_{meta}^{*} = \argmin_{\Theta_{meta}} \sum_{i = 1}^{N} \mathcal{L}_{meta}(\theta_i^*, \Theta_{meta}, D_{query_{i}}) ~~~~~~~~~~(outer~loop)
\end{equation}
\begin{equation}\label{eq:meta2}
    where,~~ \theta_i^* \equiv \theta_i^*(\Theta_{meta}) = \argmin_{\theta} l_{T_{i}}(\theta, \Theta_{meta}, D_{sup_{i}}) ~~~~~~~~~~(inner~loop)
\end{equation}

Algorithm~\ref{algo:maml} outlines the implementation of the above-discussed meta-learning, particularly the bi-level optimization of MAML. 
So, $\Theta_{meta}$ can be generalized initialization of model parameters; it can also be hyperparameters such as learning rate, regularization parameter, or parameterized loss functions. 
This work aims to concurrently `learning to learn' the structured sparsity regularization parameter along with the multi-task model parameters that offer a suitable initial point for model training in the meta-testing stage.
Further details are provided in Section~\ref{sec:meta-spar}.

\input{tables/meta_algo}

\subsection{Group sparsity}

This section introduces group sparsity, specifically $l_1-l_2$ regularization, as an important building block of the meta-sparsity framework. 
This discussion is not a new contribution; rather, it gives the theoretical context which is necessary for laying the groundwork for the proposed methodology outlined in Section~\ref{sec:meta-spar}.

Sparsity in \ac{DL} can be introduced in two broad ways: as model sparsity, affecting the model structure, or as ephemeral sparsity, impacting individual data examples, as discussed by \cite{review_sparsity}.
As already discussed in Section~\ref{sec:intro}, this work explores model sparsity, especially regularization(penalty)-based model weight sparsification.
In the context of how sparsity is introduced to model parameters, two distinct modes emerge - structured and unstructured sparsity.
Unstructured sparsity (also known as fine-grained sparsity), the more straightforward method, eliminates the least significant weights based on chosen criteria, leading to irregular sparse patterns.
In contrast, structured sparsity (also known as group sparsity) adopts a more holistic strategy by targeting the elimination of whole groups of parameters. 
In this context, a \textit{group} refers to various structural elements within the model, such as channels, filters, neurons, tailored blocks, or attention heads, each representing a distinct set of parameters (\cite{review_sparsity, liu2023lessons}).
Unstructured sparsity can significantly reduce the model size but typically achieves only modest speedups on hardware for dense computations, such as GPUs.
Meanwhile, structured sparsity is designed to complement optimized hardware architectures, leading to improvements in computational efficiency.
This work primarily concentrates on structured sparsity, utilizing penalty-based techniques for model weight sparsification.
Additionally, while our approach centers on structured sparsity, we have also implemented it for unstructured sparsity.
The findings and insights are discussed in Section~\ref{sec:discussion}.

Suppose the weights of a model are divided into G non-overlapping groups of varying sizes, denoted as $\Theta = \{\theta^1,\theta^2 \ldots \theta^G\}$.
$\mathcal{L}(f(\Theta))$ represents the loss function, where $f(\Theta)$ represents the model with parameters $\Theta$.
The objective function for the regularization-based group sparsity (group lasso) given by \cite{groupLasso} can be expressed as -
\begin{equation}\label{eq:group_spar}
\Theta^* = \argmin_\Theta ~~\mathcal{L}(f(\Theta)) + \underbrace{\lambda \sum_{g = 1}^G \sqrt{n^g} ~||\theta^g||_2}_{l_1-l_2 \text{ norm regularization term}: \mathcal{R}} where~~ \theta^g \in \Theta
\end{equation}
where $\mathcal{R}$ is the regularization term that zeros out entire groups of parameters denoted by $\theta^g$. 
Here, $n^g$ is the number of elements in the $g^{th}$ group, such that $n^g > 1~\forall g$, it is multiplied to achieve normalization of the regularization term across groups of varying sizes. 
The $l_2$ norm of the parameter group is represented by the term $||\theta^g||_2 = \sqrt{\sum_j (\theta_j^g)^{2}}$, here $\theta_j^g$ is the parameter at index $j$ for group $g$. 
In Equation~\ref{eq:group_spar}, the regularization parameter is represented by $\lambda$, which controls the degree of sparsity by balancing the trade-off between model complexity and data conformity.
We aim to learn this hyperparameter using MAML in meta-sparsity over multiple tasks. 
Given the regularization term $\mathcal{R}$ combines the $l_1$ norm of $l_2$ norms across groups, it is often referred to as $l_1-l_2$ norm in this work.

To optimize the composite loss function in Equation~\ref{eq:group_spar}, proximal gradient methods can be applied since the regularization term $\mathcal{R}$ is non-differentiable (\cite{parikh2014proximal, bach2012optimization}).
The proximal gradient descent updates the parameters based on the gradient of the differential part of the composite loss function given by
\begin{equation}
    \Theta_{t+1} \leftarrow prox_{\alpha\mathcal{R}}(\Theta_t) = prox_{\alpha\mathcal{R}}(\Theta_t - \alpha \nabla_{\Theta}\mathcal{L}(f(\Theta_t))).
\end{equation}
Here, $prox_{\alpha\mathcal{R}}$ is the proximal operator, and $\alpha$ is the learning rate.
Since $\Theta$ is divided into $G$ disjoint groups, the proximal operator can be written as,
\begin{equation}
    prox_{\alpha\mathcal{R}}(\Theta) = (prox_{\alpha\mathcal{R}}(\theta^1),\ldots,prox_{\alpha\mathcal{R}}(\theta^g),\ldots,prox_{\alpha\mathcal{R}}(\theta^G))
\end{equation}
In the case of group sparsity-inducing regularization, this proximal operator can be computed and has a closed form as discussed by \cite{combettes2005signal, hastie2015statistical, scardapane2017group, deleu2021structured}, given by the equation,
\begin{equation}
      prox_{\alpha\mathcal{R}}(\theta^g) = \begin{cases}\left[1- \dfrac{\alpha \lambda \sqrt{n^g}}{||\theta^g||_2}\right]\theta^g & ;||\theta^g||_2 > \alpha \lambda \sqrt{n^g}\\0 &  ;||\theta^g||_2 \leq \alpha \lambda \sqrt{n^g}\end{cases}
\end{equation}
So, if the $l_2$ norm of a group of parameters is less than (or equal to) a threshold depending on the size of the group and regularization parameter, the entire group of parameters is zeroed out, thereby enforcing group sparsity. 

\subsection{Proposed approach - Meta sparsity} \label{sec:meta-spar}
Now that we have established the concepts of \ac{MTL}, meta-learning, and group sparsity, we will discuss the proposed approach of meta-sparsity or learned sparsity in this section. 
Now, in the context of this work, to precisely answer the three questions related to sparsity-
(i) \textit{what to sparsify?}: the objective is to sparsify the backbone parameters \ie $\theta_b$, also known as the shared layers of the multi-task network.
(ii) \textit{how to sparsify?}: apply group sparsity (\ie Equation~\ref{eq:group_spar}) to the backbone parameters $\theta_b$ only.
The channels of the convolution layer weight matrix are treated as different groups, \ie $\theta_b^l(:,c,:,:)$ which represents one group consisting of the weights of $c^{th}$ channel of the $l^{th}$ layer.
This is symbolically illustrated in Figure~\ref{fig:arch}.
(iii) \textit{when to sparsify?}: this proposed approach is inspired by meta-learning \ie learning to learn sparsity; therefore, the sparsification is done during the meta-training phase in the outer loop.
This approach, in general, can be viewed as a \textit{learning-sparsification} iterative methodology. 
It involves episode-specific learning at the inner loop (Equation~\ref{eq:meta2}) and parameter sparsification at the outer loop (Equation~\ref{eq:meta1}) of \ac{MAML} in an iterative manner.
Therefore, similar to meta-learning, meta-sparsity also has two phases: the meta-learning phase and the meta-testing (adaptation) phase.
An overview of the meta-training and testing stages is given in the table below. 
Adding one more question to the ones above, (iv) \textit{why sparsification?}: besides model compression, the main intention behind applying structured sparsity is efficient feature selection across multiple tasks. 
Furthermore, sparsification promotes generalization, an important aspect when incorporating new yet related tasks into the multi-task framework.

Since \ac{MAML} is an episodic learning approach (\cite{meta_timothy, pmlr-v70-finn17a}), it necessitates several learning episodes, which essentially involve single-task learning. 
In this study, we extend the concept of learning episodes to include multi-task learning episodes (\cite{paper2}), encompassing all possible combinations of the $N$ tasks.
For $N$ tasks \ie $\mathcal{T} = \{T_1, T_2,\ldots, T_N\}$, the learning episodes $\mathcal{E}$ used for meta-training are an ensemble of single-task and multi-task learning episodes given by the power set of $\mathcal{T}$, such that,
\begin{equation}\label{eq:le}
     \mathcal{E} = 2^{\mathcal{T}} \setminus \{\emptyset\} = \{E_1, E_2,..., E_{2^{N}-1}\} = \{T_1, T_2,..,T_1T_N,.., T_1T_2T_N,...\}   
\end{equation}

The training procedure is inspired by \ac{MAML}(\cite{pmlr-v70-finn17a}) but with two notable distinctions.
First, the training process involves a multi-task model being trained on single-task and multi-task learning episodes to utilize sparsity patterns specific to each context;
Second, in addition to the model parameters, the hyperparameter that induces sparsity, denoted as $\lambda$, is also meta-learned, which results in a meta-sparse multi-task model.  
\input{tables/method}

In \ac{CNN}, the weight tensor is organized as a 4D array, with dimensions representing the number of filters, channels(or depth), height, and width.
When considering structured sparsity, the specific structures that can be regularized include filters, channels, filter shapes, or a customized block of parameters.
In this work, we employ channel-wise structured sparsity to the shared backbone of the multi-task architecture. 
Several previous studies, including those by \cite{wen2016learning, deleu2021structured, paper3}, among others, have focused on channel-wise sparsity across a range of applications, mainly because of two reasons.
First, this approach generates smaller dense networks by eliminating redundant channels, effectively compressing the network. 
Furthermore, these optimized networks offer greater hardware benefits, especially for GPUs, as they can be more easily accelerated.
The emphasis on channel-wise sparsity improves both computing efficiency and the suitability of models for deployment on devices with limited resources.

Consider a simple and conventional multi-task architecture with a shared encoder or backbone: a \ac{CNN} and $N$ task-specific networks (a.k.a heads) connected to the backbone. 
In this case the multi-task model parameters $\Theta = \{\theta_b\} \cup \bigcup_{i=1}^{N} \{\theta_i\}$ where $\theta_b$ are the shared parameters and $\theta_i$ are the task-specific parameters for $i^{th}$ task.
Since the group sparsity is applied to the channels of convolution layers of the backbone network, the meta-optimization objective as per Equation~\ref{eq:meta1} and \ref{eq:meta2} can be expressed along with group sparsity (Equation~\ref{eq:group_spar}) as,
\begin{equation}\label{eq:sparmeta1}
    \Theta_{meta}^{*} = \argmin_{\Theta_{meta}} \sum^{N_{b}} \mathcal{L}_{meta}(\Theta^{*}_{E_{i}}, \Theta_{meta}, D_{query_{E_{i}}}) + \lambda \sum_{g =1}^{G} \sqrt{n^g} ||\theta_{b_{E_{i}}}^{g}||_2,~~s.t.~ E_i\in\mathcal{E}~~~(outer~loop)
\end{equation}
\begin{equation}\label{eq:sparmeta2}
    where,~~ \Theta^{*}_{E_{i}} \equiv \Theta^{*}_{E_{i}}(\Theta_{meta}) = \argmin_{\Theta_{E_{i}}} \mathcal{L}_{E_{i}}(\Theta_{E_{i}}, \Theta_{meta}, D_{sup_{E_{i}}}). ~~~~~~~~~~(inner~loop)
\end{equation}
\begin{equation}\label{eq:ch_spar}
    \sum_{g =1}^{G} \sqrt{n^g}~||\theta_{b_{E_{i}}}^{g}||_2 = \sum_{l=1}^{L} \sum_{c=1}^{C_{l}} \sqrt{numel(\theta_{b_{E_{i}}}^{l}(:,c,:,:))}~~||\theta_{b_{E_{i}}}^{l}(:,c,:,:)||_2 ~~~(l_2~ norm~of~a~group). 
\end{equation}

Here, $L$ is the number of convolution layers, and $C_l$ is the number of channels in the $l^{th}$ layer. 
$numel(\cdot)$ represents the number of elements in the vector. 
$N_b$ represents the total number of batches of data in batch-wise training of the model.
$\lambda$ is the learnable regularization parameter. 
$E_i$ stands for the learning episode see Equation~\ref{eq:le}
Here, $\mathcal{L}_{E_{i}}$ is the episode-specific loss if the learning episode has a single task, then $\mathcal{L}_{E_{i}} = l_{T_{j}}$ such that, $T_j \in \mathcal{T}$.
But if it is a multi-task episode, say the $\mathcal{L}_{E_{i}} = \mathcal{F}_c(l_{T_{a}}, l_{T_{b}}, l_{T_{c}})$, where $T_{a}, T_{b}, T_{c} \in \mathcal{T}$ (see Equation~\ref{eq:mtl}).
To avoid the potential divergence of $\lambda$ towards negative infinity, a Softplus function is applied to $\lambda$, which ensures that it remains positive throughout the optimization process. 
The Softplus function, defined as $\text{Softplus}(x) = \frac{1}{\beta} \log(1 + \exp(\beta \cdot x))$, acts as a smooth approximation to the ReLU function and effectively constrains $\lambda$ to positive values. 

The Equation~\ref{eq:sparmeta1} and \ref{eq:sparmeta2} illustrate the process of bi-level optimization. 
The inner loop trains the multi-task model for various task combinations, while the outer loop finds the meta-optimized multi-task network parameters and generates a channel-sparse backbone network.
This process involves meta-learning the channel sparsity and model parameters.
The full algorithm of meta-training is outlined in Algorithm~\ref{algo:meta_sp}.
This algorithm is built on the principles of MTL (Algorithm~\ref{algo:mtl}) and MAML (Algorithm!\ref{algo:maml}). 
It combines MTL's ability to optimize shared and task-specific parameters with MAML's meta-optimization framework, enabling the dynamic learning of both model parameters ($\Theta_{meta}$) and the sparsity-inducing hyperparameter ($\lambda_{meta}$). Note that $\lambda_{meta}$ is simply a representation of $\lambda$ being meta-learned; technically, both are the same. 
\input{tables/algo}

It is worth noticing that the gradient of $\lambda$ in Equation~\ref{eq:sparmeta1} might suggest a consistent decrease, but in practice, the interaction between the sparsity term and model performance prevents $\lambda$ from collapsing to zero.
A very small $\lambda$ would hurt task performance, creating a feedback mechanism that stabilizes $\lambda$.
During meta-training, $\lambda$ gradually induces sparsity while improving performance, and as training progresses, the sparsity level naturally stabilizes (Figure~\ref{fig:l1l2-l1}). 
Additionally, the early stopping mechanism on validation loss ensures that training halts before overfitting, helping maintain a meaningful $\lambda$. 
These factors together prevent trivial solutions and allow the algorithm to effectively leverage sparsity.

Within the meta-training inner loop, the concept of regrowth can also be introduced, which involves reinstating previously pruned (or sparsified) parameters, as highlighted by (\cite{review_sparsity}).
This strategy is employed because regrowth aids in overcoming excessive sparsification, enables the model to adjust to new data patterns, and ensures a balance between achieving sparsity and maintaining performance (\cite{regrow1,regrow2}).
Moreover, the literature suggests that when regrowth is iteratively used along with sparsity (or pruning), it contributes to discovering better parsimonious network architectures.
The extent of parameter regrowth is modulated by a hyper-parameter $r_p$ that governs the probability of regrowth.
Regrowth can be implemented in the inner loop during meta-training in order to enhance episode-specific learning in this work. 
We did not apply regrowth to all trials to avoid influencing the outcome of meta-sparsity. 
However, we included some results for comparison in Section~\ref{sec:discussion} (under the subheading - Regrowing the parameters). 

In meta-testing, we have the meta sparse multi-task model, which is further fine-tuned (as discussed in Figure~\ref{fig:intro_fig}(e)).
A new task may be introduced during this stage to evaluate the generalizability of the meta-model. 
During fine-tuning, the sparsity pattern of the backbone parameters is maintained, and only the task-specific heads of the new tasks are trained. 
This is similar to standard multi-task training (Equation~\ref{eq:mtl}). 
Once we have the fine-tuned multi-task model, its performance is evaluated on the test set. 

In this study, we primarily emphasize meta-learning structured ($l_1-l_2$) sparsity for the shared layers in a multi-task setting.
However, through extensive experiments, we demonstrate that our approach is not limited to structured sparsity alone; it is equally adept at meta-learning other forms of penalty-based sparsity, such as $l_1$ (unstructured) sparsity.
For this work, we limit the focus of our work only to regularization (or penalty) based sparsity-inducing approaches.
However, we are certain that the proposed approach has broader applications and can facilitate learning sparsity in various elements, including weights, activations, or any aspect that can be adjusted via a hyperparameter.
Furthermore, the learning process involves various hyperparameters, such as learning rate, momentum, activation functions, and dropout probability, which can also be optimized through meta-learning.
However, this paper focuses on meta-learning the model parameters and the regularization hyperparameter.
This focused approach will allow us to thoroughly examine the sole influence of learning sparsity on model performance.

\subsection{A theoretical perspective on generalization}
In \ac{ML} and \ac{DL}, generalization refers to the ability of a model to adapt to new, previously unseen data, usually from the same distribution as training data (\cite{Goodfellow-et-al-2016}). 
This work encompasses three key concepts, i.e., \ac{MTL}, meta-learning, and sparsification, particularly group sparsity, and all of these contribute towards generalization in their way, as summarized in Table~\ref{tab:gen}.
This subsection explains the concept of generalization in terms of all the important components of this work that aid in establishing the theoretical foundation for generalization in the proposed meta-sparsity approach.

\begin{table}[h!]
\centering
\caption{Generalization types and their impact w.r.t. meta-learning, MTL, and Sparsification}
\label{tab:gen}
\resizebox{\columnwidth}{!}{%
\begin{tabular}{lll}
\hline
\textbf{Paradigm} & \textbf{\begin{tabular}[c]{@{}l@{}}Type of \\ Generaliza-\\ tion\end{tabular}} & \multicolumn{1}{c}{\textbf{Impact}} \\ \hline
\multirow{3}{*}{\begin{tabular}[c]{@{}l@{}}Meta-\\ Learning\end{tabular}} & \rule{0pt}{3ex}Task-Level & Learns to adapt to new tasks by capturing shared patterns across a distribution of tasks. \\
 & \begin{tabular}[c]{@{}l@{}}Distribution-\\ Level\end{tabular} & \begin{tabular}[c]{@{}l@{}}Models the task distribution, enabling effective transferability to tasks drawn from similar\\ distributions.\end{tabular} \\
 &  &  \\
\multirow{2}{*}{MTL} & Task-Level & Leverages shared representations to enhance performance across related tasks. \\
 & Data-Level & \begin{tabular}[c]{@{}l@{}}Exploits cross-task data regularization to prevent overfitting and improve learning from \\ limited data.\end{tabular} \\
 &  &  \\
\begin{tabular}[c]{@{}l@{}}Sparsi-\\ fication\end{tabular} & \begin{tabular}[c]{@{}l@{}}Feature-\\ Level\end{tabular} & \begin{tabular}[c]{@{}l@{}}Prunes redundant parameters to focus on task-relevant features, reducing overfitting and \\ enhancing generalization to unseen data.
\end{tabular}  \\ \hline
\end{tabular}%
}
\end{table}

Theoretically, the combination of MTL, meta-learning, and sparsification establishes a robust framework for achieving generalization.
For cross-task generalization (task-level and distribution-level), MTL facilitates the learning of task-invariant representations by leveraging shared knowledge across tasks. 
Meta-learning enhances adaptability to new tasks by optimizing model parameters across a distribution of tasks, which acts as a good initialization for a new task. 
Additionally, sparsification focuses on essential shared features, thereby reducing overfitting to specific tasks. 
For within-task generalization (data-level, feature-level), MTL acts as a regularizer by utilizing multi-task data, mitigating overfitting on individual tasks (\cite{caruana1997multitask, crawshaw2020multitask}). 
Meta-learning further ensures efficient adaptation to new, unseen data within a task by identifying generalizable priors (\cite{meta_timothy, huisman2021survey}). 
Meanwhile, sparsification reduces noise and retains only the most critical connections, simplifying the model and enhancing its performance on novel data (\cite{review_sparsity}). 
Together, these methods create an approach to improve generalization across and within tasks.
This motivation drove the adoption of meta-learning to optimize the sparsity-inducing hyperparameter rather than treating it simply as a learnable parameter within the \ac{MTL} setting.
The proposed approach in this work aims to meta-learn the sparsity hyperparameters (as well as the model parameters) across various combinations of multiple tasks, resulting in a sparse meta-learned backbone. 
Theoretically, as per meta-learning principles, backbone should effectively adapt to previously unseen tasks.

Therefore, in the context of this work, generalization mainly refers to the adaptability of the shared backbone in a multi-task network, highlighting its capacity to generalize to previously unseen tasks and data.
The robustness of the backbone lies in its ability to learn task-agnostic features that serve as a strong foundation for introducing new task-specific decoders during testing. 
Theoretically, this ensures that the shared backbone can adapt to previously unseen tasks, enabling efficient transfer of knowledge across tasks and encouraging scalability in multi-task settings.
We demonstrate this empirically through the experimental results (in Section~\ref{sec:discussion}) by showing the capability of a backbone to support new tasks effectively, hence generalizing to new scenarios.

%% file: tables/mtl_algo.tex
\begin{algorithm}
\small
\caption{Multi-Task Learning (MTL)}
\label{algo:mtl}
\begin{algorithmic}
\Require Sample $N$ tasks from a task distribution $p(T)$, say $\mathcal{T} = \{T_1, T_2,...T_N\}$
\Require MTL model with parameters, $\Theta = \{\theta_b\} \cup \bigcup_{i=1}^{N} \{\theta_i\}$ \bluecomment{where $\theta_b =$ shared parameters and $\theta_i =$ task-specific parameters}
\While{not converged}
    \For{$b_s$ = 1 \textbf{to} number of batches in dataset $D$}
        \State Sample a batch of data for each task $T_i$ \bluecomment{training data for all tasks}
        \State Compute task-specific losses, $l_{T_i}$, for each task $T_i$
        \State Compute multi-task loss, $\mathcal{L}_{MTL} = \mathcal{F}_{c}(l_{T_1}, l_{T_2}, ..., l_{T_N})$ \bluecomment{aggregation of task-specific losses}
        \State Compute gradients, $g = \nabla_{\Theta} \mathcal{L}_{MTL}$
        \State Update parameters, $\Theta \gets \Theta - \alpha \cdot g$ \bluecomment{$\alpha$ is the learning rate}
    \EndFor
\EndWhile
\end{algorithmic}
\end{algorithm}

%% file: tables/meta_algo.tex
\begin{algorithm}
\small
\caption{Model-Agnostic Meta-Learning (MAML)}
\label{algo:maml}
\begin{algorithmic}
\Require Sample $N$ tasks from a task distribution $p(\mathcal{T})$, say $\mathcal{T} = \{T_1, T_2, \dots, T_N\}$
\Require Initialize meta-parameters $\Theta_{meta}$ \bluecomment{initial shared model parameters}

\While{not converged}
    \State Initialize gradient accumulator $\mathcal{G} \gets [\hspace{0.4em}]$ \bluecomment{empty list for gradient accumulation}
    \For{each task $T_i \in \mathcal{T}$} \bluecomment{inner loop for task-specific adaptation}
        \State Sample support set $D_{sup_i}$ and query set $D_{query_i}$
        \State Compute task-specific loss $\mathcal{L}_{T_i}$ on $D_{sup_i}$ \bluecomment{use support data for task-specific training}
        \State Update $\Theta_{T_i} \gets \Theta_{meta} - \alpha_{in} \cdot \nabla_{\Theta_{meta}} \mathcal{L}_{T_i}$ \bluecomment{perform one or more gradient steps, $\alpha_{in}$: inner loop learning rate}
        \State Compute meta-loss gradient $G_i \gets \nabla_{\Theta_{T_i}} \mathcal{L}(\Theta_{T_i}, D_{query_i})$ \bluecomment{gradient on query set}
        \State Accumulate gradient $\mathcal{G} \gets \mathcal{G} \cup \{G_i\}$
    \EndFor
    \State Compute aggregated meta-gradient $\mathcal{G}_{meta} = \text{Average}(\mathcal{G})$ \bluecomment{aggregate gradients from all tasks}
    \State Update $\Theta_{meta} \gets \Theta_{meta} - \alpha_{out} \cdot \mathcal{G}_{meta}$ \bluecomment{outer loop meta-update, $\alpha_{out}$: outer loop learning rate}
\EndWhile
\end{algorithmic}
\end{algorithm}

%% file: tables/method.tex
\begin{table}[!h]
\centering
\small
\begin{tabular}{llll}
\cline{1-2} \cline{4-4}
\multicolumn{2}{c}{\textbf{META-TRAINING}}                    &  & \multicolumn{1}{c}{\textbf{META-TESTING}}        \\ \cline{1-2} \cline{4-4} 
\multicolumn{2}{c}{\textbf{Inner loop} \textit{(learning)}}                                &  & \multicolumn{1}{c}{\textbf{Fine-tuning} (\textit{similar to MTL})} \\
Objective:                 & reduce the episode-specific loss &  & \textit{Begin with - Meta sparse multi-task model}                   \\
                           & i.e. $\mathcal{L}_{E_{i}}$       &  & Three possibilities:                             \\
\multicolumn{1}{c}{Train:} & episode-wise training of multi-  &  & (i) Fine-tuning on all the same N tasks          \\
                           & task model with sparse backbone  &  & as meta-training.                                \\
Optimize:                  & multi-task model parameters($\Theta_{E_{i}}$)      &  & (ii) Add new task and fine-tune only             \\
                           & for every learning episode.      &  & on the new $(N+1)^{th}$ task                     \\
\multicolumn{2}{c}{\textbf{Outer loop} \textit{(sparsification)}}  &  & (iii) Add new task and fine-tune on all \\
Objective:                 & reduce the meta-loss             &  & the (N+1) tasks.                                 \\
                           & i.e. $\mathcal{L}_{meta}$        &  & \textit{(maintain sparsity while fine-tuning)}   \\
Train:                     & meta-train the multi-task model  &  & \multicolumn{1}{c}{}                             \\
                           & on all learning episodes.        &  & \multicolumn{1}{c}{\textbf{Testing}}             \\
Optimize:                  & multi-task model parameters($\Theta_{meta}$),     &  & Use an independent test set to evaluate          \\
                           & regularization parameter($\lambda_{meta}$)         &  & the performance of the meta-sparse               \\
                           & (inducing channel-wise sparsity) &  & multi-task model                                            \\
Outcome:                   & Meta sparse multi-task model     &  &                                                  \\ \cline{1-2} \cline{4-4} 
\end{tabular}
\end{table}

%% file: tables/algo.tex
\begin{algorithm}[ht!]
\small
\caption{Meta-sparsity (training)}
\label{algo:meta_sp}
\begin{algorithmic}
\Require Sample N tasks from a task distribution p(T), say $\mathcal{T} = \{T_1, T_2,...T_N\}$
\Require MTL model with parameters, $\Theta = \{\theta_b\} \cup \bigcup_{i=1}^{N} \{\theta_i\}$ \bluecomment{where $\theta_b =$ shared parameters and $\theta_i =$ task-specific parameters} 

\Require Initialize meta parameters $\Theta_{meta}$ and $\lambda_{meta}$\bluecomment{dense model and sparsity hyperparameter}
\Require Single \& multi-task episodes: $\mathcal{E} = \{E_1, E_2,..,E_{2^N-1}\}$
\While{not converged} 
\For{$b_s$ = 1 \textbf{to} no. of batches of $D_{sup}$ and $D_{query}$} \bluecomment{$outer~loop$}
    \State Randomly sample an episode $E_i$ 
    \State $\mathcal{G} \gets [\hspace{0.4em}]$   \bluecomment{Using [\hspace{0.4em}] to denote an empty list}
    \State Initialize $\Theta_{E_{i}} \gets \Theta_{meta}$
    \If{$\Theta_{E_{i}}$ has sparse groups and regrow probability $r_p > 0$} \bluecomment{Regrowing sparse parameters}
        \State - Select groups for regrowth (randomly or some criteria))
        \State - Initialize the selected groups
    \EndIf
    \For{$\kappa$ inner updates} \bluecomment{$inner~loop$}
        \If{$E_i$ contains $n$ tasks such that, $n > 1$}
             \State $\mathcal{L}_{E_{i}} = \mathcal{F}_{c}(l_{T_1}, l_{T_2}, \ldots, l_{T_n})$ \bluecomment{multi-task loss}
        \Else
            \State $\mathcal{L}_{E_{i}} = l_{T_j}$ \bluecomment{single task loss}
        \EndIf

    \State Calculate gradients, $ g= \nabla_{\Theta_{E_{i}}} \mathcal{L}_{E_{i}}(\Theta_{E_{i}}, \Theta_{meta}, D_{sup_{E_{i}}})$
    \State Update $\Theta^{*}_{E_{i}} \gets \Theta_{E_{i}} - \alpha_{in} \hspace{0.3em} g$ \bluecomment{$\alpha_{in}$ is the learning rate for adaptation stage}
    \EndFor
    \State For the query set, calculate $ G_i= \nabla_{\Theta^{*}_{E_{i}}} \mathcal{L}_{meta}(\Theta^{*}_{E_{i}}, \Theta_{meta}, D_{query_{E_{i}}})$
    \State Accumulate gradients, $\mathcal{G} \gets \mathcal{G} \cup \{G_i\}$ \bluecomment{gather gradients for all the learning episodes}
\EndFor
\State Calculate meta gradients, $\mathcal{G}_{meta}$, as the average of all accumulated gradient 
 \State Update $\Theta_{meta}^{*} \gets prox_{\alpha_{out}}(\Theta_{meta} - \alpha_{out}~\mathcal{G}_{meta,\Theta})$   \bluecomment{meta-update for model parameters where $\alpha_{out}$ is the learning rate for meta stage (outer loop)}
 \State Update $\lambda_{\text{meta}}^{*} \gets \text{prox}_{\alpha_{\text{out}}}(\lambda_{\text{meta}} - \alpha_{\text{out}}~\mathcal{G}_{meta,\lambda})$ \bluecomment{meta-update for $\lambda_{meta}$}
\EndWhile
\end{algorithmic}
\end{algorithm}

%% file: sections/4_expsetup_results.tex
In this section, we elaborate on the experimental framework designed to investigate the effectiveness of our proposed meta-sparse multi-task models.
Here, we provide an overview of the datasets chosen for multi-task learning, describe the structure of the multi-task network used, and outline the specific evaluation metrics employed for each task.
Additionally, we outline the types of experiments designed to assess the effectiveness of our proposed methodology.

\input{tables/loss}
In this work, two widely recognized and publicly accessible datasets, the NYU-v2 dataset (\cite{Silberman_ECCV12}) and CelebAMask-HQ dataset (\cite{CelebAMask-HQ}), are used, hereafter referred to as NYU and CelebA, respectively in this paper.
Table~\ref{tab:loss} details the various tasks considered for each dataset, the respective loss functions, and evaluation metrics (see Appendix or supplementary material for details). 
We have adopted the standard loss functions (and metrics) as utilized in previous works such as \cite{sun2020adashare, paper2} and many others, to primarily focus on assessing the impact of sparsity on task performance and avoiding performance enhancements that could arise from the use of complex, custom-designed loss functions.
Overall, in the NYU dataset, all the tasks are pixel-level tasks, segmentation is pixel-level classification, and the rest are pixel-level regression. 
While in the celebA dataset, there is only one pixel-level (classification) task, and the rest are image-level (classification) tasks. 
This set of tasks was chosen explicitly to extensively evaluate the performance of the proposed approach across a wide range of task combinations, proving their effectiveness and adaptability in many contexts.

As already discussed in Section~\ref{sec:method}, a very standard multi-task network architecture is chosen, which has a backbone network \ie the layers shared by all the tasks, and task-specific heads or layers are connected to the output of the backbone, see Figure~\ref{fig:arch}. 
We selected the dilated ResNet-50  (\cite{yu2017dilated}) architecture as our backbone network due to its resilience to sparsity; even when sparsity leads to entire layers having zero parameters, the residual connections within the network ensure continuity by effectively propagating values forward. 
This characteristic maintains the network's structural integrity and facilitates uninterrupted information flow, making it an ideal choice for our experiments.
For the dense prediction tasks, a deeplab-v3 network (\cite{chen2017rethinking}) is employed as a task-specific network, and for the binary classification task, a two-layer fully connected network is used.
In the past, similar network designs and architecture were also used by \cite{kendall2018multi,liebel2018auxiliary, paper2}, and many more.
For a fair comparison of results and to maintain consistency across evaluations, all the experiments in this work use the same architecture, loss functions, metrics, train-validation-test split, and hyperparameters.
Models are trained using NVIDIA A100 Tensor Core GPUs, which have 40 GB of onboard HBM2 VRAM. 
To assess the reliability of the model, we conducted five replications of each experiment using distinct random seeds. 
The findings are presented in terms of the mean and the standard deviation. To ensure reproducibility, the source code can be accessed at: \href{https://github.com/PLACEHOLDER_to_the_git_repo}{https://github.com/PLACEHOLDER TO THE GIT REPOSITORY}

The following are the types of experiments designed and analyzed in this work-
\begin{itemize}
    \item Experiments \textit{without} group sparsity (using the dense model, $\lambda = 0$)
\begin{enumerate}
    \item Single task learning - one model for each task.
    \item Multi-task learning - employ a multi-task model for different task combinations.
    \item Multi-task + meta-learning - learn the multi-task model parameter initialization only during meta-training. 
    \begin{itemize}
        \item These are also referred to as \textit{meta-learning baseline} experiments in the subsequent text. 
    \end{itemize}

\end{enumerate}
\item Experiments \textit{with} group sparsity (applying channel-wise structured sparsity on the backbone layers)
\begin{enumerate}
    \item Single task learning + sparse backbone; Fixed sparsity (\ie $\lambda$ is fixed).
    \item Multi-task learning + sparse backbone; Fixed sparsity (\ie $\lambda$ is fixed).
    \item Multi-task learning + meta-sparse backbone; Learnable sparsity (\ie $\lambda$ is meta-learned). 
    \begin{itemize}
        \item These are the \textit{meta-sparsity} experiments designed to meta-learn both the parameter initialization and hyperparameter ($\lambda$) that induces sparsity.
    \end{itemize}

\end{enumerate}
\end{itemize}
In experiments involving meta-learning, namely the meta-learning baseline (MTL + meta-learning) and meta-sparsity (MTL + meta-learning + learnable sparsity), the performance of tasks under these three settings is evaluated during meta-testing, utilizing the backbone learned during meta-training:
\begin{itemize}
\item[]
    \begin{enumerate}
    \item Meta-testing on the same tasks as meta-training.
    \item Add a new task, and fine-tune only the new task.
    \item Add a new task to the meta-training task and fine-tune all the tasks.
\end{enumerate}
\end{itemize}

To clarify, adding a new task involves integrating a new decoder to the meta-trained (multi-task) backbone for a task that has not been previously introduced \ie during meta-training.
Note that the single experiments without and with fixed sparsity do not use meta-learning for parameter initialization. 
Meta-learning is used in the MTL+ meta-learning (without sparsity) and meta-sparsity experiments. 
This means that the meta-parameter initializations obtained from meta-training are used during meta-testing for these experiments.
Additionally, all the multi-task experiments are performed for 3-4 task combinations. 
Given the computation requirement, not all the $2^N-1$ combinations for N tasks are evaluated.
For the NYU dataset, we examine three multi-task scenarios: (i) the complete set ($T_1, T_2, T_3, T_4$) encompassing all tasks, (ii) a mixed set of classification and regression tasks ($T_1, T_2, T_3$) with one task reserved for meta-testing, and (iii) a subset of solely regression tasks ($T_2, T_3, T_4$), omitting the classification task for meta-testing purposes.
Similarly, for the celebA dataset, the following combinations are considered : (i) all the seven tasks \ie ($T_1-T_7$), (ii) all the binary classification tasks \ie ($T_2-T_7$), and (iii) ($T_1, T_2, T_3, T_7$); other tasks with very similar attributes primarily related to the mouth region of the image were added during meta-testing to study the performance.

%% file: tables/loss.tex
\begin{table}[ht]
\centering
\caption{A table containing the loss functions and evaluation metrics for the various tasks for both the datasets: NYU-v2 and CelebAMask-HQ. In the table a downward arrow ($\downarrow$) represents that a lower value is better while an upward arrow ($\uparrow$)  represents a higher value is better. Also, IoU stands for intersection over union and MAE stands for mean absolute error.}
\label{tab:loss}
\resizebox{\textwidth}{!}{%
\begin{tabular}{lllll}
\hline
Dataset & Tasks & ($T$) & Loss Functions & Evaluation Metric \\ \hline
\multirow{4}{*}{NYU-v2} & Semantic segmentation & $T_1$ & cross-entropy loss & IoU ($\uparrow$) \\
 & Depth estimation & $T_2$ & \begin{tabular}[c]{@{}l@{}}Combination of errors in depth gradient\\ and surface normal(\cite{depth_loss})\end{tabular} & MAE ($\downarrow$) \\
 & Surface Normal estimation & $T_3$ & Inverse cosine similarity & cosine similarity ($\uparrow$) \\
 & Edge detection & $T_4$ & huber loss(\cite{PAUL2022100218})& MAE ($\downarrow$) \\ \hline
 & Semantic segmentation & $T_1$ & cross-entropy loss & IoU ($\uparrow$) \\
\begin{tabular}[c]{@{}l@{}}CelebA \\ Mask-HQ\end{tabular} & \begin{tabular}[c]{@{}l@{}}Binary classification\\ (attributes- male, smile, big lips, \\ high cheekbones, wearing lipstick,\\ bushy eyebrows)\end{tabular} & $T_2 - T_7$ & binary cross-entropy loss & Accuracy ($\uparrow$) \\ \hline
\end{tabular}%
}
\end{table}

%% file: sections/5_discussion.tex
\begin{figure}[ht]
    \centering
    \begin{minipage}{0.43\textwidth}
        \centering
        \includegraphics[width=\linewidth]{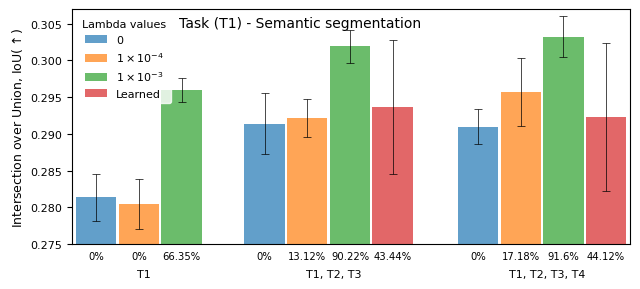}
    \end{minipage}
    \begin{minipage}{0.55\textwidth}
        \centering
        \includegraphics[width=\linewidth]{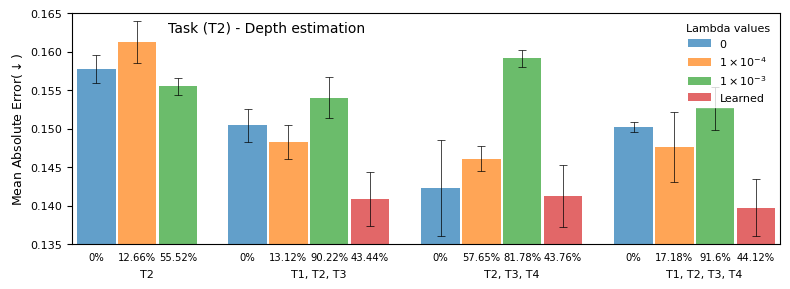}
    \end{minipage}       
    \begin{minipage}{0.55\textwidth}   
        \centering
        \includegraphics[width=\linewidth]{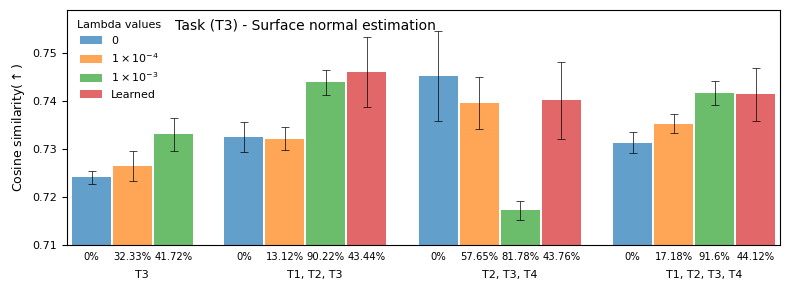}
    \end{minipage}
    \begin{minipage}{0.43\textwidth}   
        \centering
        \includegraphics[width=1.002\linewidth]{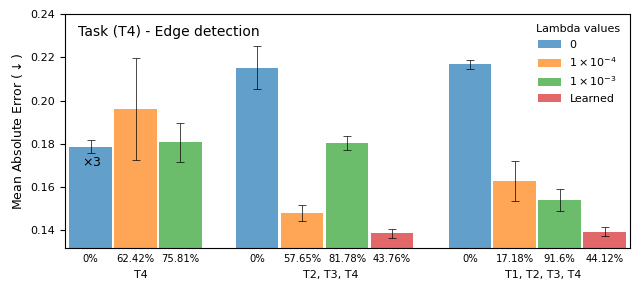}
    \end{minipage}
    \caption{For NYU dataset, task-wise performance comparison of single-task and multi-task no sparse ($\lambda = 0$, in blue), fixed sparsity single and multi-task (for $\lambda = 1\times10^{-4}$ in yellow and $\lambda = 1\times10^{-3}$ in green) and meta sparsity ($\lambda$ = learned in red \ie meta-sparsity) experiments. The vertical axis label is annotated with an upward or downward arrow to indicate whether a higher or lower metric value is preferable. The values below each bar represent the percentage of parameter sparsity in the backbone network. The `x3' on the initial bar for task T4 indicates that the depicted performance metric (MAE) is triple the current value represented by the bar. This notation is employed to simplify the depiction of values that exhibit significant disparities. In these plots, we have selected a narrow range for the y-axis to make it easier to compare the performances. However, this may have enhanced the visual effect of the error bars representing the performance's standard deviation. These results are also presented in tabular form in the supplementary material Table A.1.}
    \label{fig:NYUmain}
\end{figure}

This section primarily answers the following questions: why multi-task over single-task learning?
Is the meta-sparsity approach viable? 
Is the performance of meta-sparsity robust and stable?
Do the observed improvements in meta-sparsity performance compared to the meta-learning baseline result from the effectiveness of meta-learning the sparsity hyperparameter, or are they merely a consequence of the advantages of meta-learning itself?
How do the meta-sparse multi-task models perform compared to the fixed sparsity multi-task models? 
How effectively is the sparse backbone model processing a novel, unseen task?
Therefore, this section aims to provide comprehensive insight into the applicability and advantages of the proposed meta-sparsity approach within \ac{MTL} learning, supported by empirical evidence. 
Detailed results are provided in the Appendix for further reference.

\textit{Note:} For the fixed lambda experiments, we looked at how different tasks and combinations of tasks respond to the same setting of a sparsity parameter, $\lambda$. 
This parameter controls the level or amount of sparsity, which is measured in terms of (\%) parameter sparsity\footnote{In this work, \textit{parameter sparsity} refers to the proportion of the model’s parameters which are zeroed out, represented as a percentage}.
We focused on three specific settings of this parameter: $1\times10^{-3}$, $1\times10^{-4}$, and $1\times10^{-5}$ because these values induced sparsity in most of the single-task and multi-task settings. 
\begin{figure}[ht]
    \centering
    \begin{minipage}{0.42\textwidth}
        \centering
        \includegraphics[width=\linewidth]{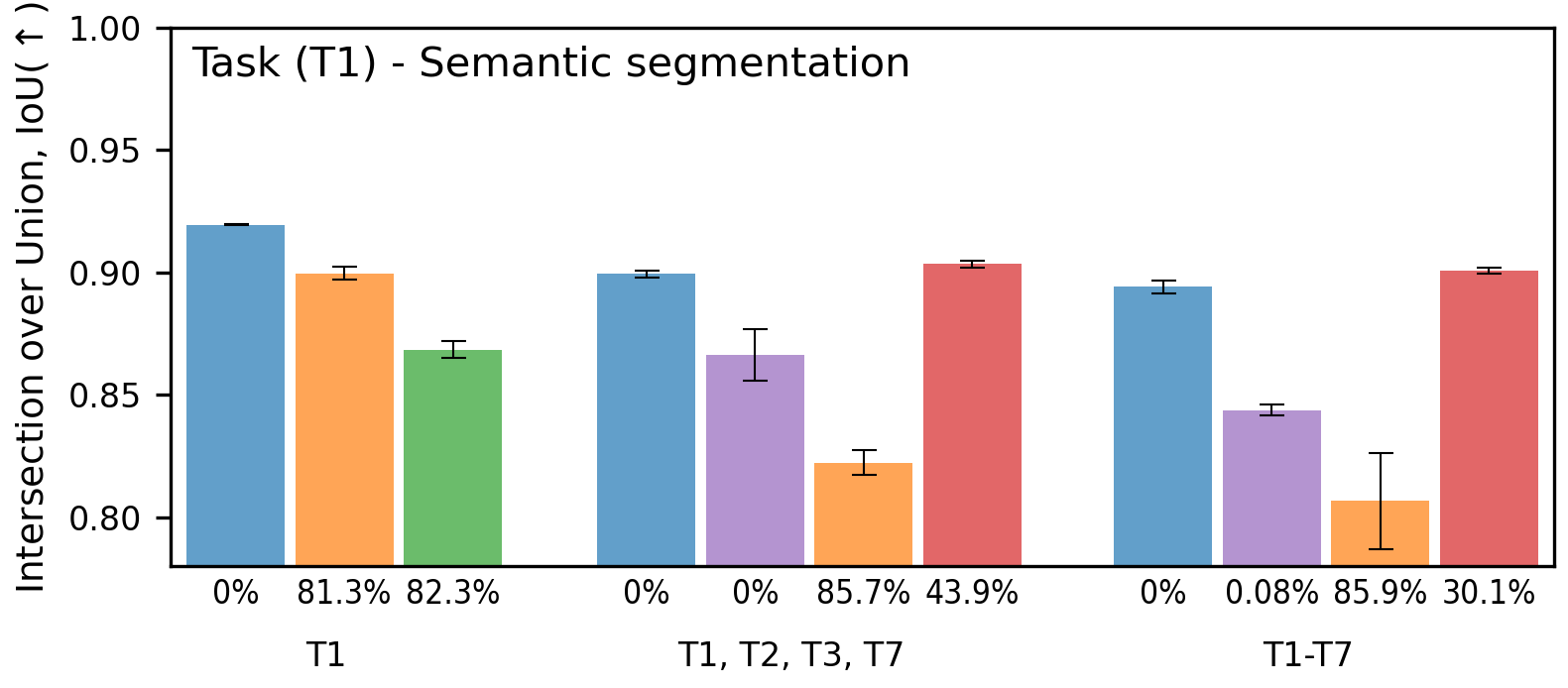}
    \end{minipage}
    \begin{minipage}{0.55\textwidth}
        \centering
        \includegraphics[width=1.02\linewidth]{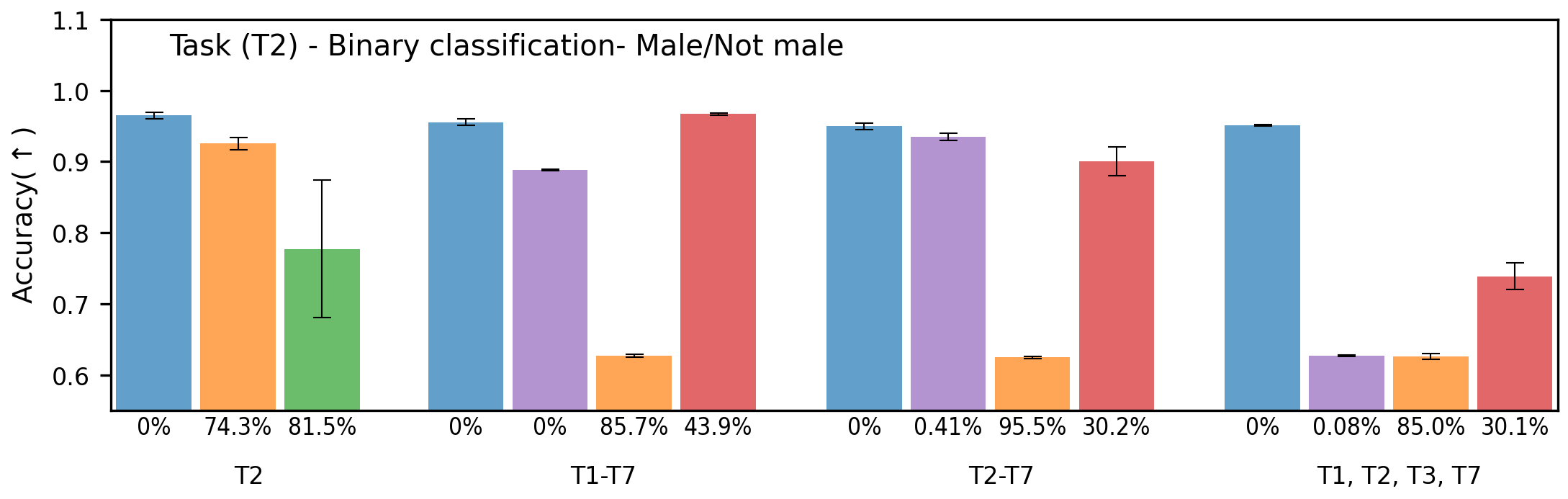}
    \end{minipage}    
    \begin{minipage}{0.55\textwidth}  
        \centering
        \includegraphics[width=\linewidth]{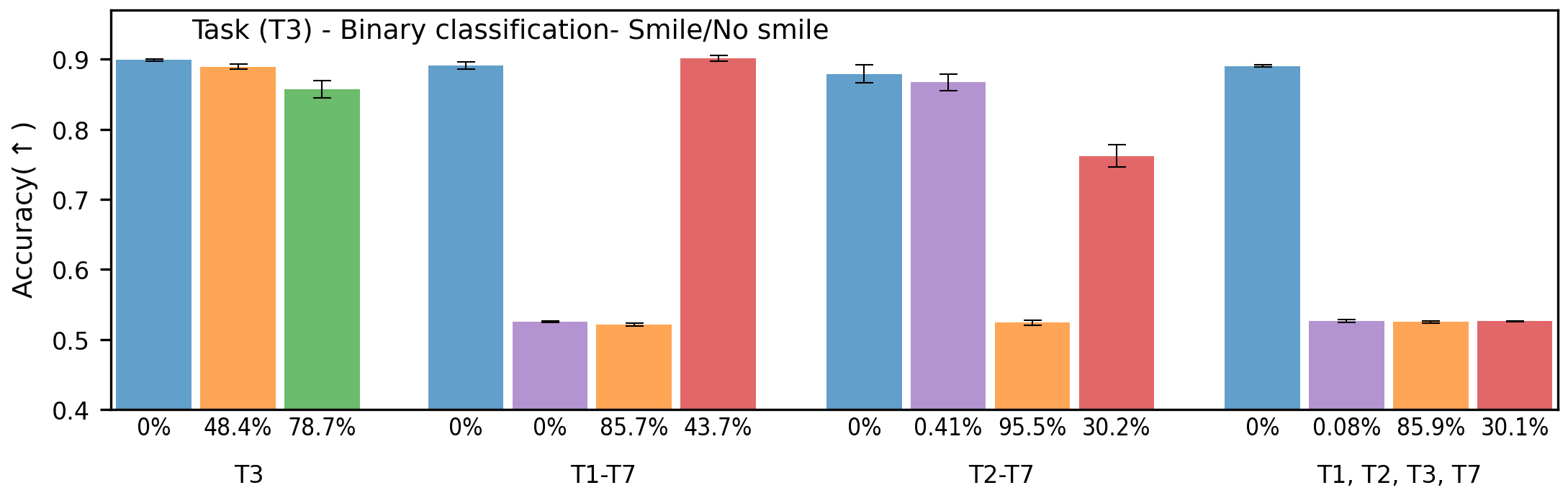}
    \end{minipage}
    \begin{minipage}{0.42\textwidth} 
        \centering
        \includegraphics[width=\linewidth]{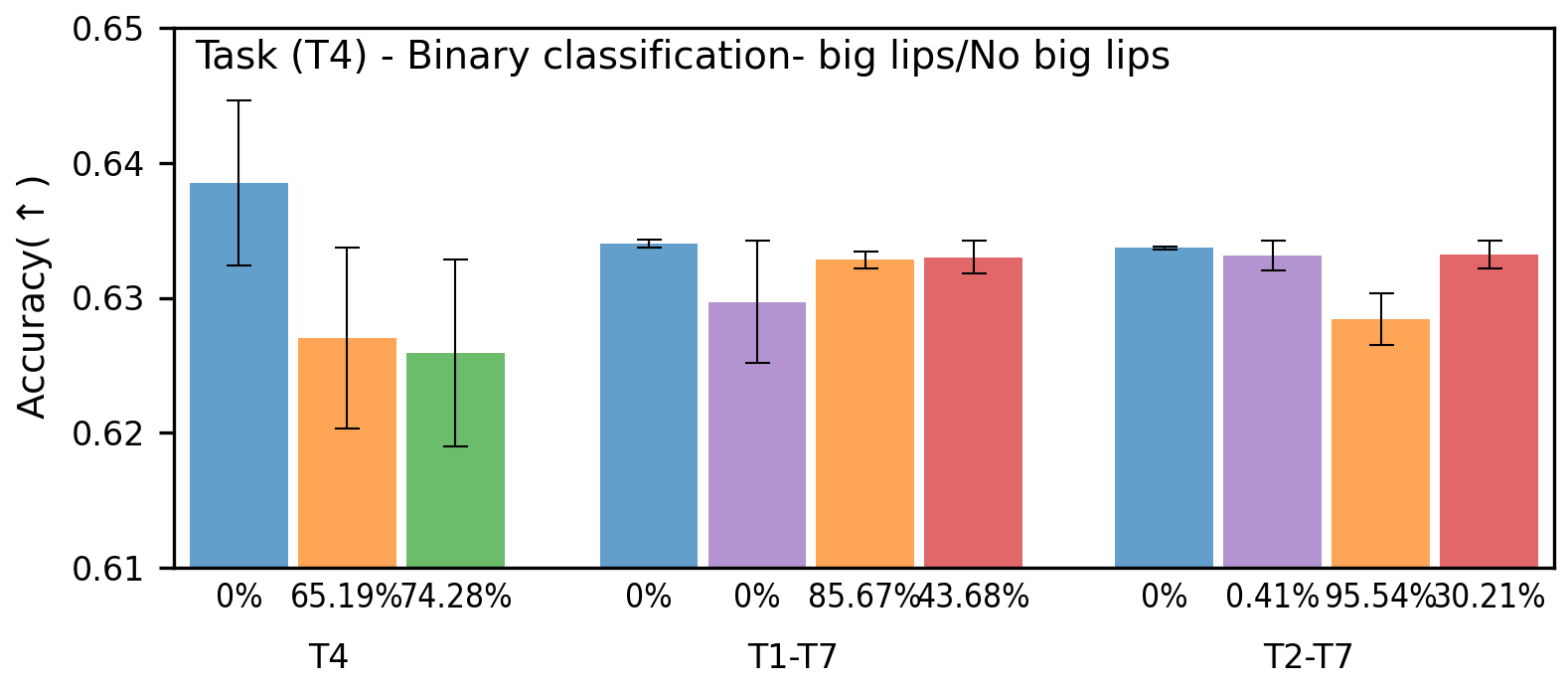}
    \end{minipage}
    \begin{minipage}{0.47\textwidth}  
        \centering
        \includegraphics[width=0.98\linewidth]{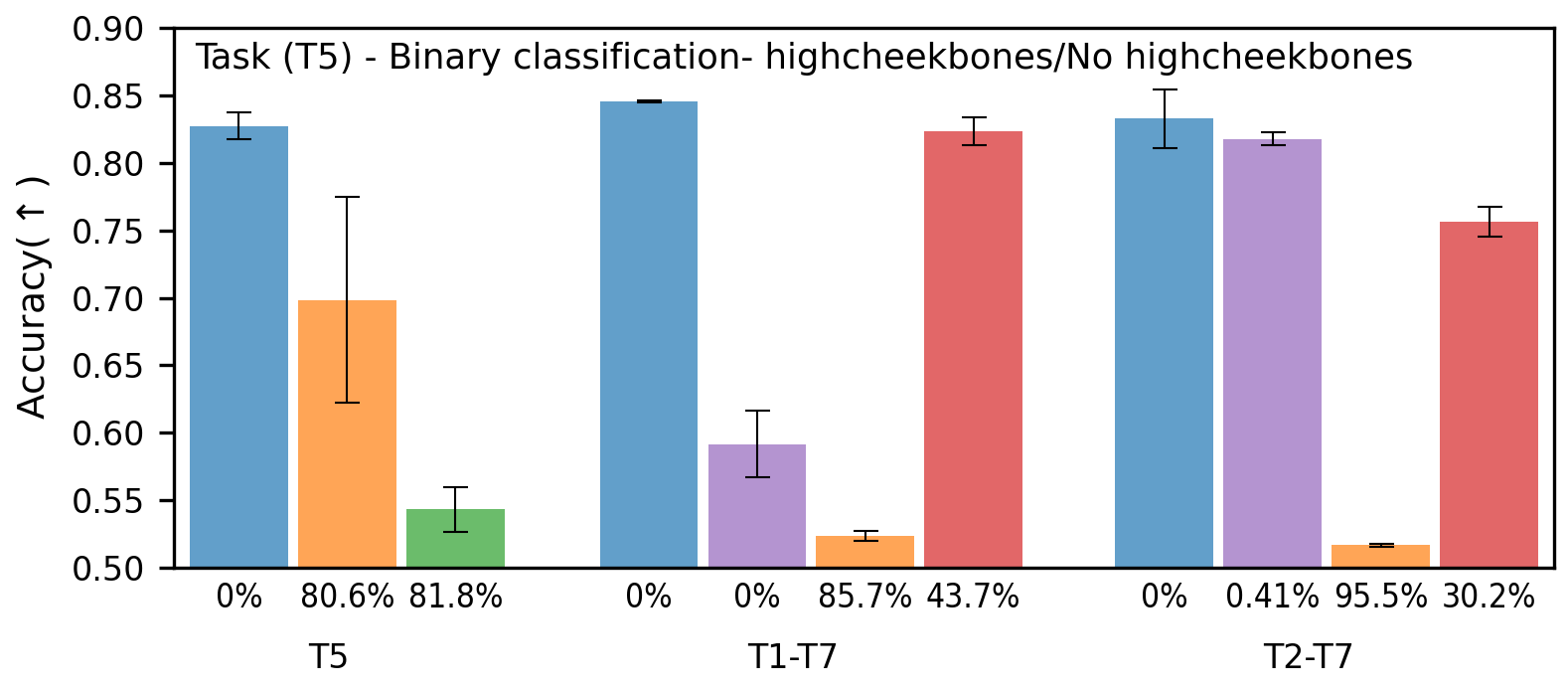}
    \end{minipage}
    \begin{minipage}{0.47\textwidth} 
        \centering
        \includegraphics[width=0.98\linewidth]{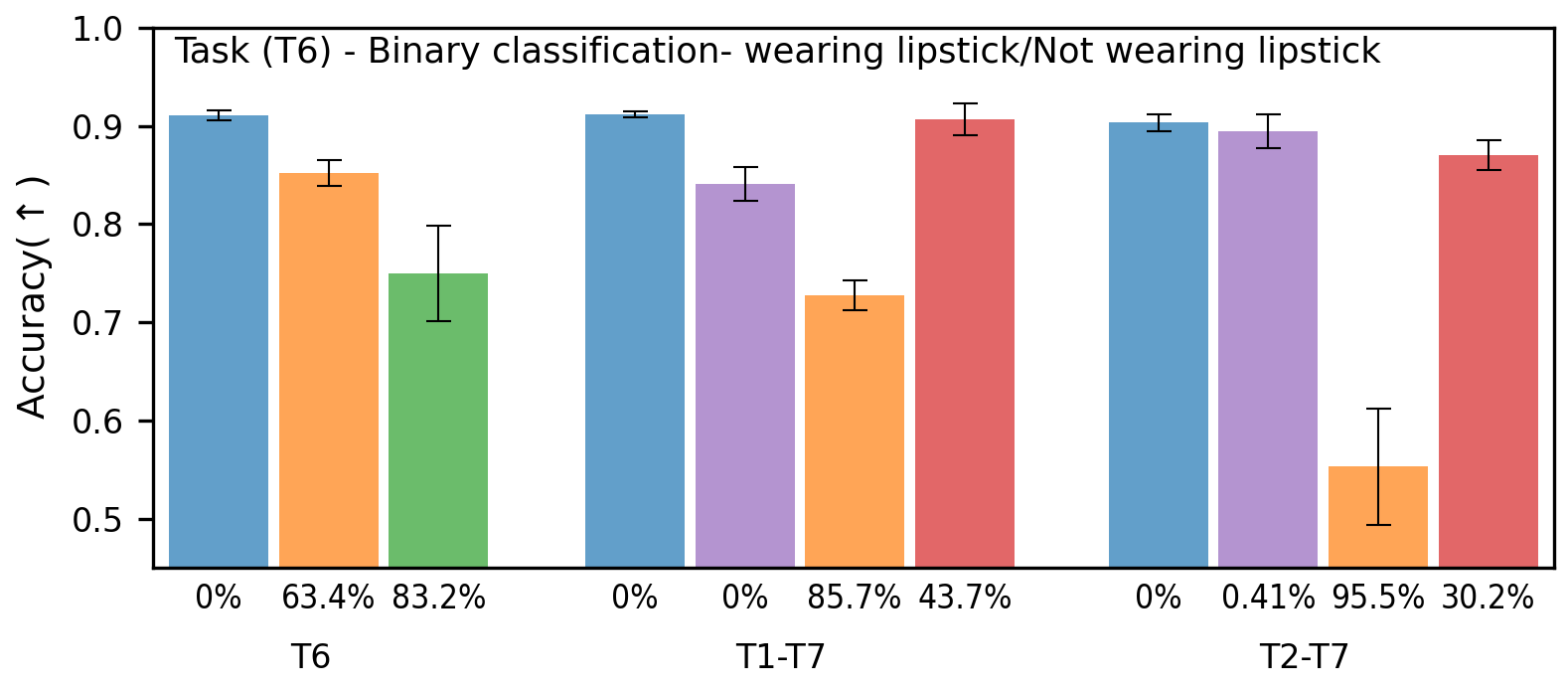}
    \end{minipage}
    \begin{minipage}{0.7\textwidth} 
        \centering
        \includegraphics[width=1\linewidth]{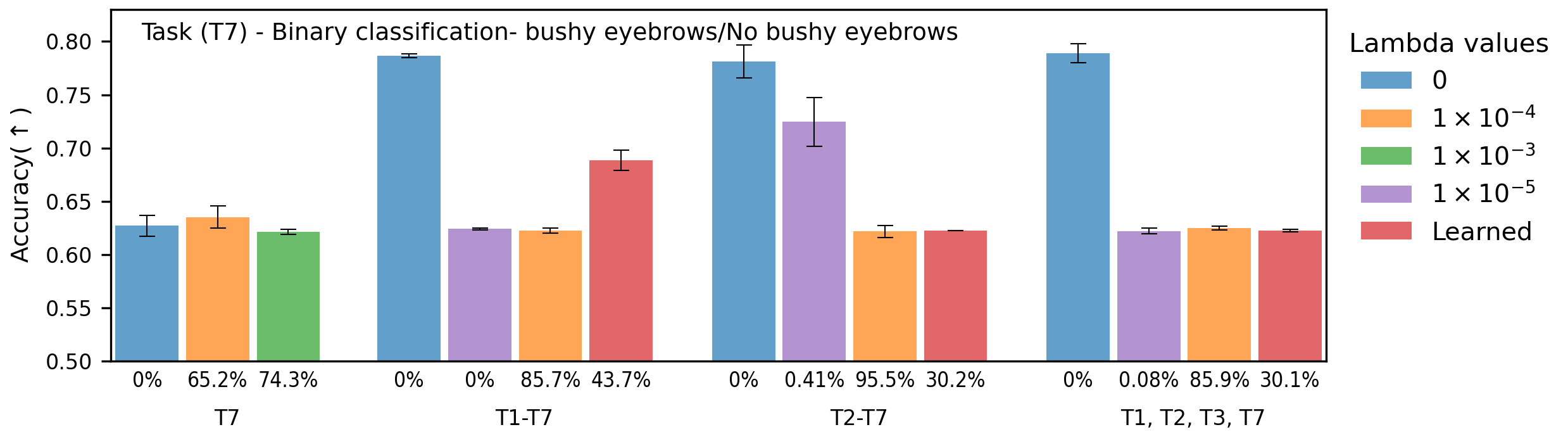}
    \end{minipage}
    \caption{For celebA dataset, task-wise performance comparison of single-task and multi-task no sparse ($\lambda = 0$, in blue), fixed sparsity single and multi-task (for $\lambda = 1\times10^{-5}$ in lavender, $\lambda = 1\times10^{-4}$ in yellow and $\lambda = 1\times10^{-3}$ in green) and meta sparsity (in red) experiments. The vertical axis label is annotated with an upward or downward arrow to indicate whether a higher or lower metric value is preferable. These results are also presented in tabular form in the supplementary material Table A.3.}
    \label{fig:mainceleb}
\end{figure}

\textbf{Assessing the viability of meta-sparsity:} 
Figures~\ref{fig:NYUmain} and \ref{fig:mainceleb} illustrate the task-wise performance for the NYU and celebA datasets, respectively.
For ease of comparison, the bar charts display the results for single-task learning (both with and without fixed sparsity), multi-task learning (again, with and without fixed sparsity), and meta-sparsity multi-task learning (featuring learned sparsity) across all tasks.
For the NYU dataset (Figure~\ref{fig:NYUmain}), for tasks $T_2$ and $T_4$, the meta-sparse multi-task models significantly outperform compared to other model configurations. 
For task $T_3$, these models achieve performance on par with their counterparts, whereas for task $T_1$, they slightly under-perform relative to the single or multi-task fixed sparsity counterparts, while are equivalent or better than not sparse models.
Meta-sparse models surpass their fixed-sparsity counterparts in terms of performance despite having a lower level of parameter sparsity, highlighting a trade-off between performance and sparsity.

For the celebA dataset, meta-sparsity excels for some task combinations.
However, specifically for the task combination $T1, T_2, T_3, T_7$, the performance across all the tasks, except for $T_1$ (\ie segmentation), is undesirable.
A probable reason behind this can be that the task ensemble involves classification tasks aimed at significantly distinct attributes.
This under-performance can primarily be attributed to task interference, where the meta-sparsity approach faces challenges in identifying optimal features for sharing between tasks.
However, the expanded combination of $T_1-T_7$ leads to improved performance for many of these tasks.
This improvement suggests that more related (similar attributes) tasks improve performance through enhanced feature sharing among tasks facilitated by meta-sparsity.
Some exceptions are discussed further in this section. 

\textbf{Regularization Stochasticity and discussion on performance stability:} 
To enhance the visibility and comparability of the performance differences, we chose a narrow y-axis range for the plots in Figure~\ref{fig:NYUmain}, which may have amplified the visual impact of the variability of the error bars representing the standard deviation.
Please refer to Table A.1 in the supplementary material (which forms the basis of Figure~\ref{fig:NYUmain}), which quantitatively demonstrates that the standard deviations of the performance metrics of the tasks are consistently low (in the range of 0.001-0.0101). 
However, it is evident that for many tasks, the variance in the performance of meta-sparsity is slightly greater than the rest (no sparsity and fixed sparsity).
One probable reason can be ‘Regularization Stochasticity’ (\cite{8587027}). 
It can be defined as the variability in the learned sparsity patterns because of the stochastic nature of (mini-batch) gradient-based optimization and the dynamic penalty-based sparsity applied during training. 
This stochasticity can result in different paths of convergence and patterns of sparsity when the same experiment is run multiple times; therefore, there is slight variability in the performance. 
Another factor to consider, that can amplify these stochastic effects, is the random initialization of the strength of the regularization hyperparameter ($\lambda$), which is trainable/learned in meta-sparsity. 
We sample the initial value of $\lambda$ from a uniform distribution between 0.1 and 1. 
While this may seem like a small range, it can still lead to significant variations in the convergence paths and the resulting sparsity patterns.
The aforementioned reasons for performance instability are also applicable to the CelebA dataset. 

\textbf{Stability in the amount (\%) of sparsity with fixed $\lambda$ and in meta-sparsity:} As the preceding subsection addresses performance stability, it is also essential to look at the stability of percentage sparsity over the experiments.
Table A.1 in the supplementary material presents the percentages of group sparsity and parameter sparsity for the NYU-v2 dataset, while task performance is also illustrated in Figure~\ref{fig:NYUmain}.
In both single-task and multi-task experimental settings, under fixed sparsity,  the percentage of group and parameter sparsity exhibits significant variance (high standard deviation) across multiple trials. 
The high variance comes from differences in parameter initialization, causing small differences in the trained parameters (regularization stochasticity as discussed above).
These small changes can disproportionately impact sparsity by leading to the elimination of entire groups or channels of the parameter matrices, resulting in considerable fluctuations in overall sparsity levels. 
In contrast, the meta-sparsity experiments demonstrate substantially lower variance in percentage sparsity, highlighting the robustness of the proposed approach.
This behavior can be observed for both datasets (see Tables A.1 and A.3 in the supplementary material).
The NYU dataset exhibits greater variability in the percentage sparsity of the meta-sparsity experiments compared to the CelebA dataset.
This is because NYU tasks involve dense, pixel-level predictions, which demand diverse and task-specific adjustments to the sparsity patterns across different layers of the shared network to preserve important features. 
In contrast, the classification tasks of the celebA dataset share more uniform feature extraction needs, resulting in comparatively more consistent sparsity levels during meta-training.


\textbf{Comparative performance analysis: }
Before evaluating the proposed meta-sparsity approach, we would like to provide an insight into why \ac{MTL} is preferred over single-task learning.   
Experiments (Figures~\ref{fig:NYUmain} and \ref{fig:mainceleb}) typically show that tasks perform better in a multi-task environment compared to a single-task environment. This observation emphasizes the importance of utilizing multi-task learning episodes.
Additionally, the amount of sparsity varies depending on the specific task (in single-task scenarios) and the combination of tasks (in multi-task scenarios). 
There is no clear trend in sparsity levels that might help to predict how the sparsity of a multi-task model changes in relation to the number of tasks.
It can be inferred that sparsity is affected by the nature and interconnection of the tasks involved.
Further elaboration on task performance in the celebA dataset is provided in the following subsections.

Interestingly, when analyzing the outcomes for the celebA dataset, we noticed that for the lowest value \ie $1\times10^{-5}$, the experiments, both single and multi-task, show no sparsity at all, and the performance is also similar to the without sparsity setting.
While for  $1\times10^{-3}$ and $1\times10^{-4}$, they exhibit very high sparsity (around 85\% - 95\%) in just one or two epochs, but this comes at the cost of not being trained enough, which negatively affects its performance (not for single tasks). 
This is the reason behind the poor performance of all the tasks for $\lambda = 1\times10^{-3}$ and $1\times10^{-4}$, see the Figure~\ref{fig:mainceleb} (the lavender and yellow bars).
In the case of single-task experiments, they perform better since the feature requirement in single-task settings is much simpler (straightforward) than in multi-task settings.
From the above analysis, it is logical to infer that other values of $\lambda$, both lower than $1\times10^{-5}$ and higher than $1\times10^{-3}$, would exhibit similar behaviors.
Specifically, values below $1\times10^{-5}$ may not cause considerable sparsity, potentially resulting in overfit models without performance advantages, while values above $1\times10^{-3}$ may result in sparse models early in training, impeding learning and performance. 
This conjecture highlights the delicate balance needed to optimize sparsity and model performance when selecting $\lambda$.
This fuels the rationale behind the primary aim of this research, which is to learn the optimal sparsity parameter $\lambda$, which in turn is expected to enhance performance.

Also, for the task $T_7$ \ie bushy eyebrows/ no bushy eyebrows, the performance of meta sparsity is not up to the mark for any of the task combinations. 
This shortfall could be attributed to the unique nature of the task. 
Most selected tasks are associated with attributes of the entire image, such as segmentation and gender classification, or specifically relate to the mouth area, including distinctions like smiling/not smiling, having big lips/no big lips, and wearing lipstick/not wearing lipstick; the sparse features dominate towards these tasks. 
In contrast, task $T_7$ is the only task that concentrates on the eye region, specifically eyebrows, which likely requires features different from those needed for the other tasks, which is why its performance in isolation is much better.
Similarly, the semantic segmentation task $T_1$ from the NYU dataset performs satisfactorily in a fixed-sparsity situation but does not reach comparable performance levels under meta-sparsity (although there is a significant standard deviation).
This discrepancy is likely due to the nature of $T_1$ as the sole pixel-level classification task among others that are pixel-level regression tasks, suggesting that sparse meta-parameters might not be well-suited to support $T_1$ effectively.
Overall, it can be summarized that the relatedness of the tasks influences their feature requirements, which in turn dictates both the sparsity pattern and performance outcomes. 
The competitive and cooperative dynamics of \ac{MTL} might cause some tasks to struggle or perform poorly in the ensemble while others may excel. 
It emphasizes the significance of cautiously choosing and balancing tasks in an ensemble to enhance overall performance and sparsity efficiency.

\input{tables/nyu_mtml} 

\textbf{Comparing meta-sparsity and MTL + meta-learning (meta-learning baseline):}  
Tables~\ref{tab:NYU_add} and \ref{tab:celeb_add} offer a comparison between the proposed meta-sparsity approach and a baseline meta-learning method without sparsification in a \ac{MTL} setup.
The proposed meta-sparsity approach performs better than the conventional meta-learning baseline across almost all evaluated tasks for both datasets. 
It is important to observe that the performance metrics of single-task and standard multi-task experiments frequently match closely with the meta-learning baseline. 
The performance of the baseline meta-learning approach may, in certain instances, be marginally inferior to that of standard \ac{MTL} (without sparsity).
This suggests meta-learning gains may not always translate into large improvements across all tasks.
It might be because a meta-learning approach that focuses on optimizing for a shared initialization across tasks may face difficulties in sufficiently capturing the distinct features of individual tasks, particularly when tasks are diverse.
The performance gains in meta-sparsity can be noticed when introducing unseen tasks during the meta-testing stage and fine-tuning only the new tasks. 
Further details on this comparison are provided in the next sub-section, while this section focuses specifically on comparison with the meta-learning baseline.
In the meta-learning baseline experiments involving the addition of unseen tasks, the meta-model adapts to these additional tasks, resulting in performance levels comparable to that observed in single-task and multi-task configurations. 
However, there is no significant improvement in performance, except for a few tasks within the celebA dataset. 

As stated, meta-learning aims to find an optimal parameter initialization that shows good generalization across various tasks. 
In the \ac{MTL} setting, it is worth noticing that parameter initialization alone may not be sufficient for effectively learning various tasks. 
This shared initialization lacks inherent specialization for capturing the distinct and refined features of individual tasks.
This limitation affects its ability to generalize, particularly in the context of unseen tasks.
Overall, the results show that integrating group sparsity into the meta-learning + MTL framework (\ie resulting in meta-sparsity) improves the overall performance and boosts the generalizability of the learned shared representations. 
This is mainly because group sparsity, by reducing redundant parameters, forces the model to focus on features beneficial for both shared learning and individual task performance, thereby enhancing the adaptability of the model.
Hence, the interaction of MTL, meta-learning, and group sparsity  (in meta-sparsity) can generalize better, adapt more efficiently to new tasks, and deliver higher performance compared to the meta-learning baseline that lacks sparsity, thereby validating the hypothesis presented in Section~\ref{sec:method}.

\textbf{Evaluating the efficacy of meta-sparse backbone:}
We study the effectiveness of the sparse backbone by introducing novel, previously unseen tasks during the meta-testing stage, \ie a task different from the ones the model is meta-trained on.
So, both the task and the data during meta-testing are unseen. 
We studied the performance of the sparse shared backbone network under two distinct scenarios:  firstly, when the new task is integrated alongside the tasks from the meta-training phase, and secondly, when the model is fine-tuned exclusively on the new task. 
Note that the level of sparsity achieved during the meta-training stage is maintained during meta-testing by masking the layers that were zeroed out. 
Tables~\ref{tab:celeb_add}~and~\ref{tab:NYU_add} show the performance of the new tasks for the celebA and NYU datasets, respectively. 
For comparison, we also show the performance of the meta-trained tasks during meta-testing. 

For the celebA dataset, when tasks $T_4, T_5, T_6$ are added to the pre-existing set of tasks $T_1, T_2, T_3, T_7$ for meta-testing, the performance across the tasks improves or remains stable, without any significant degradation.
Notably, task $T_3$ (smile classification) shows significant improvement with the inclusion of mouth-related tasks $T_4$ (big lips/no-big lips) and $T_6$ (wearing lipstick/not wearing lipstick) due to more focused learning on mouth features. 
Similarly, tasks $T_2$ (male classification) and $T_7$ (bushy eyebrows classification) benefit from this refined feature extraction, enhancing performance.
Similar observations can be made when semantic segmentation $T_1$ (\ie a pixel-level task) is added to the mix of classification tasks ($T_2 - T_7$).
The segmentation task demands pixel-level, fine-grained image understanding, which enhances the shared feature representation, benefiting the classification tasks with more robust and discriminative features. 
For the NYU dataset (see Table~\ref{tab:NYU_add}), in both the cases, \ie addition of $T_1$ and $T_4$, the performances of the tasks are consistent when compared to the outcomes of other settings.
When these tasks are fine-tuned in a multi-task setting along with other tasks, there is either slight improvement or maintenance of performance levels for the existing tasks without any observed degradation.

These observations highlight the advantages of \ac{MTL}, where simultaneous training on multiple tasks can lead to better overall performance through shared insights and learning dynamics.
It also highlights the performance consistency across tasks, \ie the sparse backbone demonstrates robust behavior even when new tasks are introduced in the mix. 
Occasionally, it is noticed that the addition of new tasks might also enhance the performance of meta-training (old) tasks. 
This suggests that the tasks mutually support and improve learning.
Furthermore, the fact that meta-training tasks maintain stable performance during meta-testing, even with the introduction of new tasks, suggests that the proposed method is robust and may help prevent negative information transfer.

\input{tables/baseline_comp}
\textbf{Comparisons with other sparsity baselines:} 
To give a baseline performance comparison of the proposed meta-sparsity approach with other sparsification methods, Table~\ref{tab:comp_performance} presents a comparison in three levels: (i) Sparsification approaches illustrated in Figure~\ref{fig:intro_fig}, (ii) Sparsity patterns are also referred to as masks in this discussion, and (iii) Parameter initialization before sparsification.
For a fair comparison between the approaches, we forced all the sparsification methods to maintain the same \% parameter sparsity achieved by meta-sparsity. 
From Table~\ref{tab:comp_performance} it is evident that there is no one approach which works the best for all the tasks. 
The lowest magnitude parameter elimination for progressive and iterative sparsification shows promising results for $T_1, T_2, T_3$, but for $T_4$ the performance is suboptimal. 
Zeroing out the parameters randomly also shows some promising performance. 
However, it is to be noted that these performances are at an optimal sparsity budget($\sim$44\%) learned by meta-sparsity. 
As the above table shows, meta-sparsity achieves comparable performance across all tasks. 
The strength of our approach lies in its ability to dynamically (meta) learn sparsity patterns, which leads to an optimal amount of sparsity. 
Other methods require a sparsity budget, thresholds, or a sparsity step in case of iterative and progressive sparsification to regulate the level of sparsity, and very often, it is very tedious to find the correct balance between sparsity percentage and task performance. 
To verify the viability of the learned meta-sparsity patterns, we applied the learned meta-mask across various sparsification approaches.
The meta mask/pattern consistently performed well for almost all tasks and across all approaches, demonstrating that the learned pattern is an optimal sparsity pattern for various tasks in an MTL setting.


\begin{figure}[ht]
    \centering
    \begin{minipage}{0.24\textwidth}
        \centering
        \includegraphics[width=\linewidth]{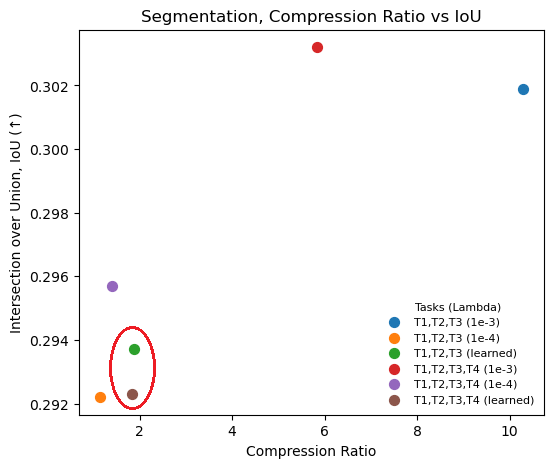}
    \end{minipage}
    \begin{minipage}{0.24\textwidth}
        \centering
        \includegraphics[width=\linewidth]{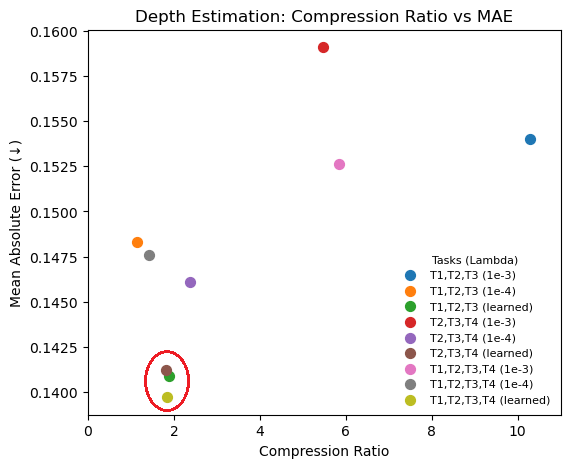}
    \end{minipage}
    \begin{minipage}{0.24\textwidth}
        \centering
        \includegraphics[width=\linewidth]{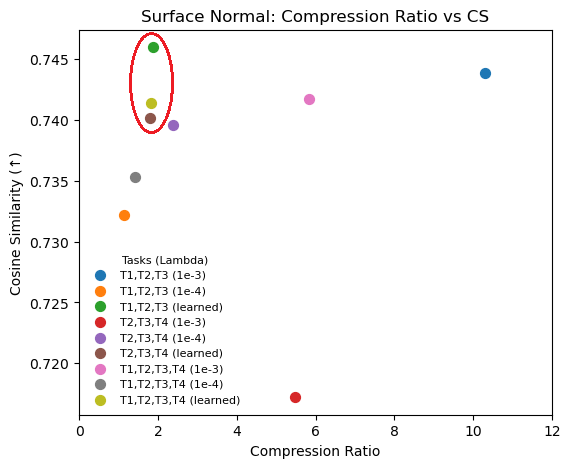}
    \end{minipage}
    \begin{minipage}{0.24\textwidth}
        \centering
        \includegraphics[width=\linewidth]{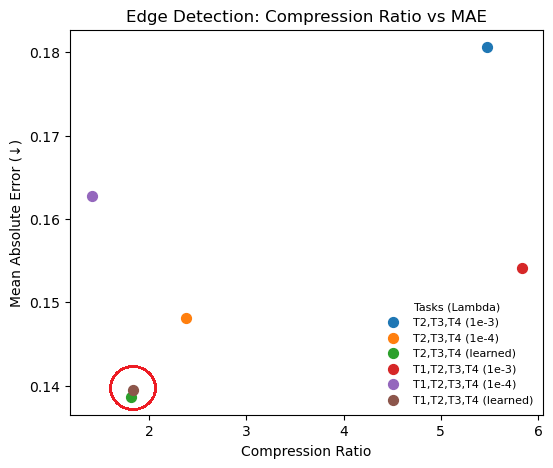}
    \end{minipage}
    
    \vspace{0.5em} 
    
    \begin{minipage}{0.24\textwidth}
        \centering
        \includegraphics[width=\linewidth]{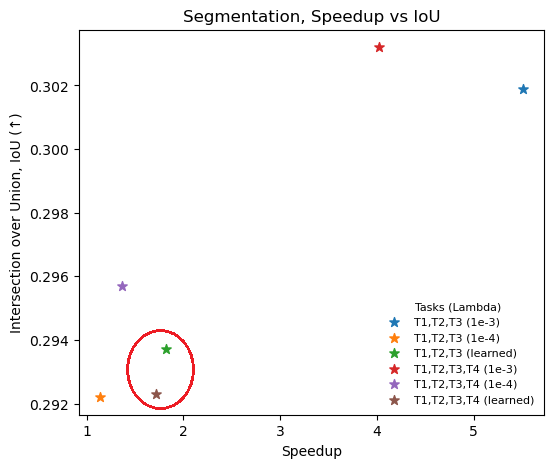}
    \end{minipage}
    \begin{minipage}{0.24\textwidth}
        \centering
        \includegraphics[width=\linewidth]{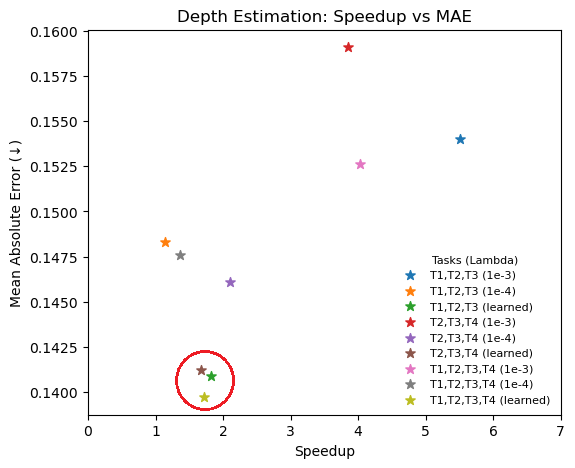}
    \end{minipage}
    \begin{minipage}{0.24\textwidth}
        \centering
        \includegraphics[width=\linewidth]{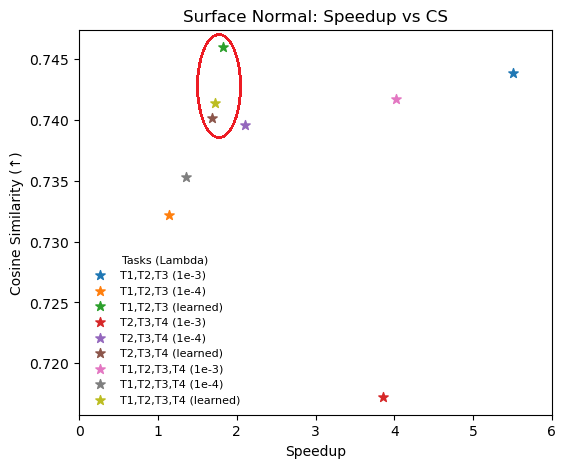}
    \end{minipage}
    \begin{minipage}{0.24\textwidth}
        \centering
        \includegraphics[width=\linewidth]{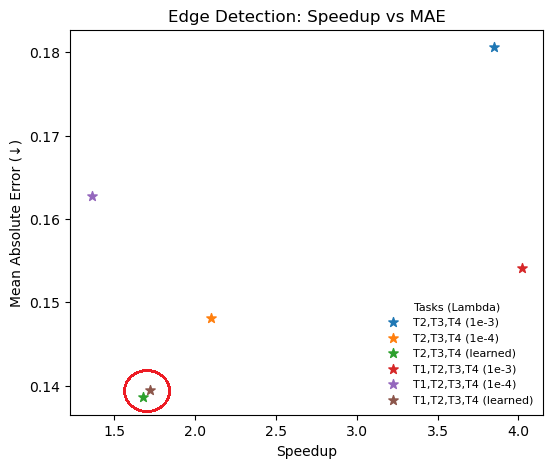}
    \end{minipage}
    
    \caption{This figure presents a comparison of performance against both the compression ratio (top row) and speed-up (bottom row) across all tasks within the NYU dataset, with meta-sparsity experiments distinctly marked by red circles. It is important to note that the upward arrow ($\uparrow$) on the y-axis denotes that a higher value of the metric is preferable, and the downward arrow ($\downarrow$) represents that a lower value of the metric is preferable. The tabulated metrics are presented in Table A.4 in the supplementary material.}
    \label{fig:compNYU}
\end{figure}


\textbf{Other sparsity metrics:}
In the context of sparse models, compression ratio\footnote{$CR = \dfrac{total~parameters}{no.~of~nonzero~parameters}$} (CR) and speed-up\footnote{$Sp = \dfrac{total~FLOPs}{no.~of~nonzero~FLOPs}$; FLOPs = floating point operations} (Sp) are two important metrics that quantify the application of sparsity(\cite{blalock2020state}).
The compression ratio measures the extent of model compression by comparing the size (in a number of parameters) of the original model with the compressed model.  
Speed-up is not a direct measure of the size reduction; it is a result of model compression. 
Due to sparsity, when a model is compressed, it typically requires fewer computational resources \ie FLOPs, which can lead to faster processing times.
Therefore, speed-up is an indirect measure of the operational efficiency of a model achieved due to model compression. 

Figure~\ref{fig:compNYU} shows the task-wise compression ratio and speed-up achieved due to the fixed sparsity and meta sparsity for different task combinations. 
The meta-sparsity metrics are circled in red. 
Except for task $T_1$ (segmentation), these figures illustrate a trade-off between performance and both compression and speed. 
This observation aligns with our earlier discussion on parameter sparsity vs performance in the comparative performance analysis subsection.
For example, in task $T_2$ (depth estimation), models with meta-sparsity achieve a computation speed of approximately $1.6\times$ faster than those of the dense backbone network. 
While fixed-sparsity models may be faster and more compressed than their meta-sparsity counterparts, this advantage often results in a trade-off with performance. 
For the segmentation task, it is apparent that meta-sparsity fails to boost performance and does not meet the anticipated outcomes. 
While the fixed sparsity models achieve notably greater compression, they can increase their speed by fourfold and exceed expected performance levels.
\begin{figure}[h]
    \centering
    \begin{minipage}{0.31\textwidth}
        \centering
        \includegraphics[width=\textwidth]{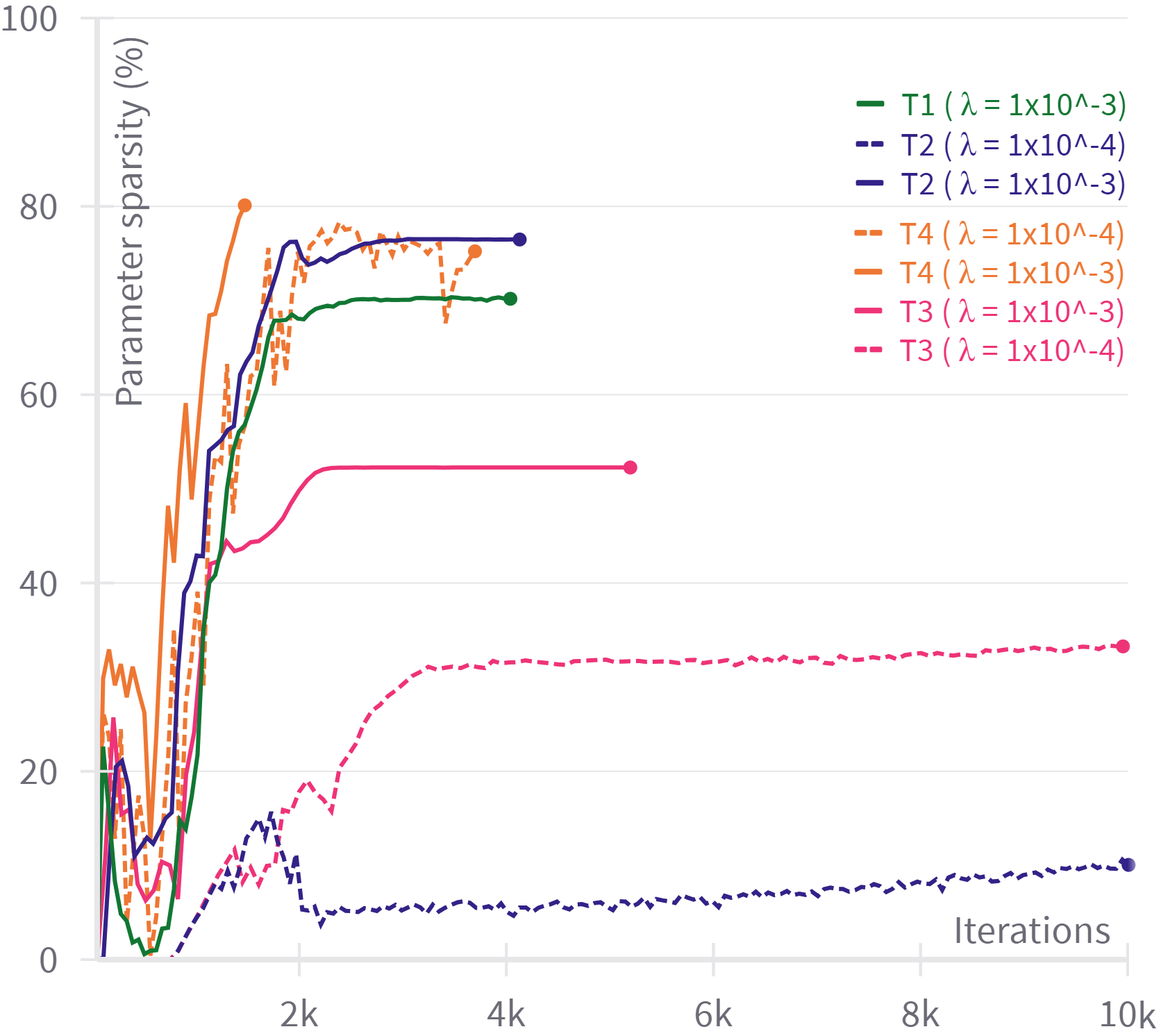} 
        \caption*{(a) single-task fixed sparsity}
        \label{fig:pattern_single}
    \end{minipage}
    \begin{minipage}{0.31\textwidth}
        \centering
        \includegraphics[width=\textwidth]{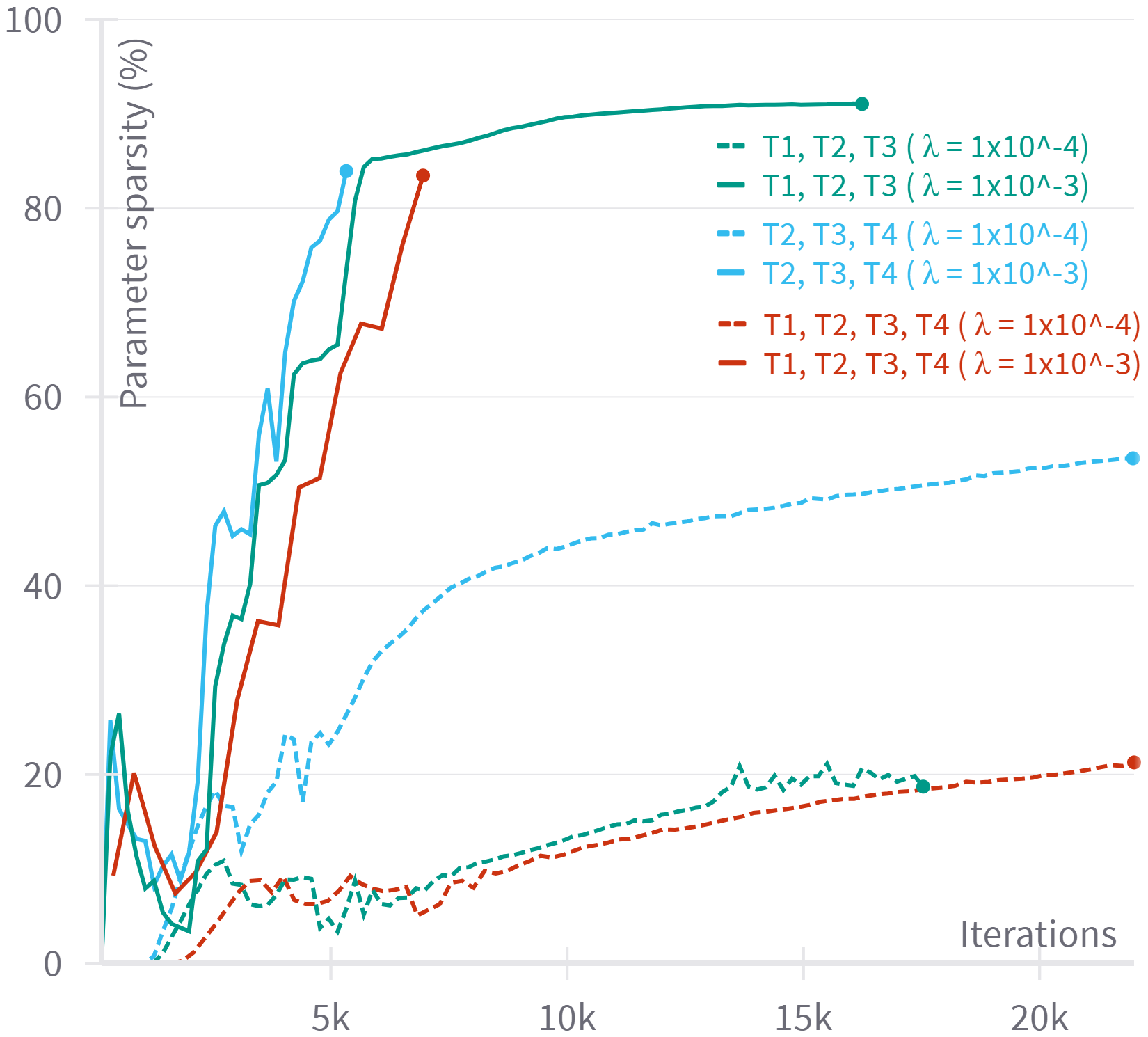} 
        \caption*{(b) multi-task fixed sparsity}
        \label{fig:pattern_multi_fixed}
    \end{minipage}
    \begin{minipage}{0.31\textwidth}
        \centering
        \includegraphics[width=\textwidth]{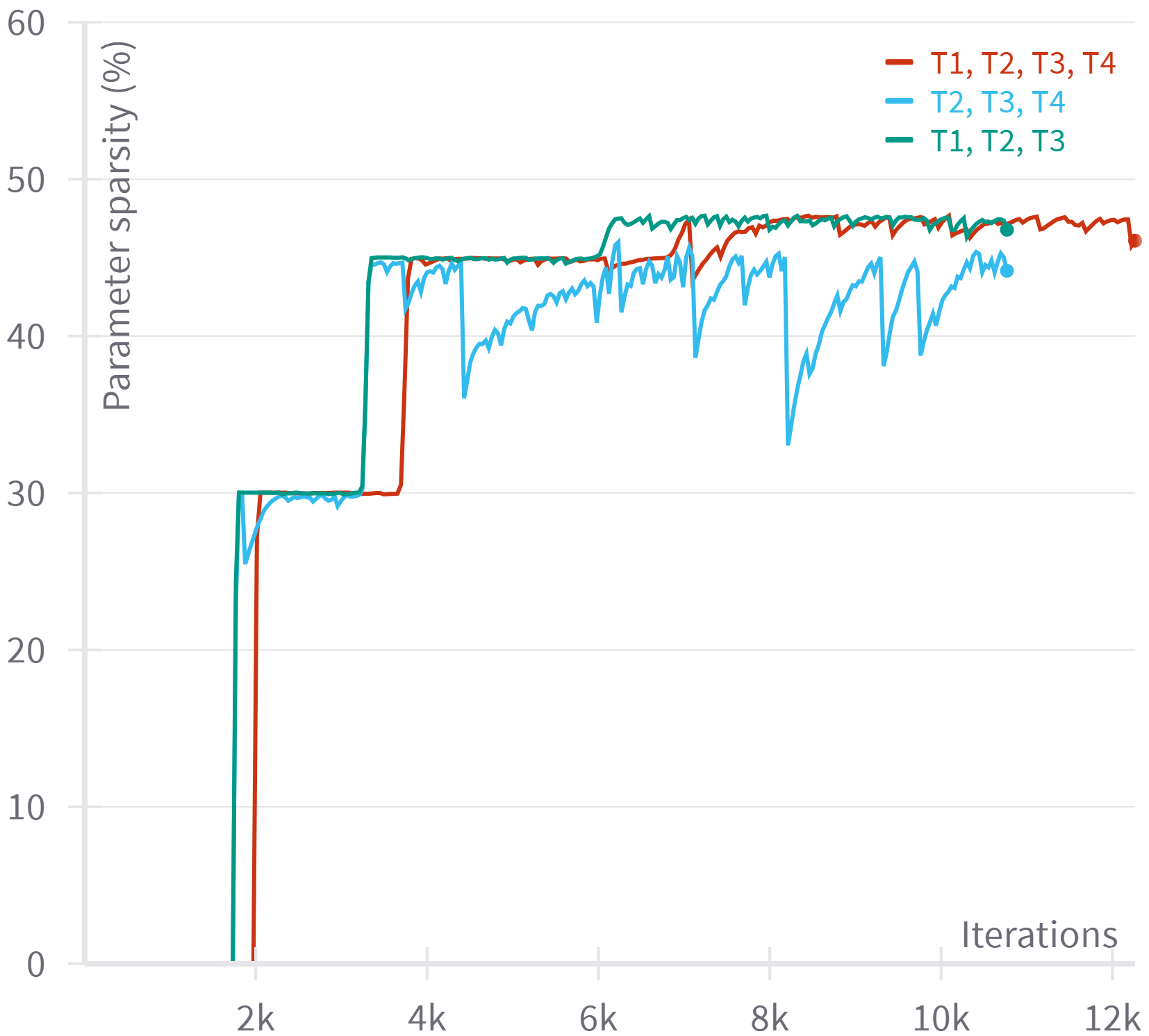} 
        \caption*{(c) multi-task meta sparsity}
        \label{fig:pattern_multi_learn}
    \end{minipage}
    \caption{Parameter sparsity patterns while training for (a) single-task fixed sparsity, (b) multi-task fixed sparsity, and (c) multi-task meta sparsity experiments. In the above plots, T1 represents semantic segmentation, T2 represents depth estimation, T3 represents surface normal estimation, and T4 represents edge detection tasks. In the fixed sparsity plots \ie (a) and (b), the dotted lines represent the experiments where $\lambda$ = $1\times10^{-4}$, while the solid lines represent the sparsity pattern when $\lambda$ = $1\times10^{-3}$.}
    \label{fig:main_pattern}
\end{figure}

\textbf{Sparsity profiles:}
Sparsity profiles illustrate the dynamic changes in the proportion of zero or inactive parameters within a model during the training process, effectively mapping the evolution of model compactness.
Figure~\ref{fig:main_pattern} illustrates the parameter sparsity profiles for the fixed and meta-sparsity experiments during training. 
It is evident from Figure~\ref{fig:main_pattern}(a) and \ref{fig:main_pattern}(b) that sparsity increases sharply early in training and then plateaus, suggesting that sparsity is introduced quickly and maintained throughout the training process.
The experiments with $\lambda=1\times10^{-4}$ attain lower levels of parameter sparsity as compared to the $\lambda = 1\times10^{-3}$; this is obvious since $\lambda$ defines the strength of sparsity.
Moreover, sparsity is achieved more swiftly in single-task setups than in multi-task ones.
This is because \ac{MTL} needs more iterations to determine which features are unnecessary for the ensemble and may be removed.

In Figure~\ref{fig:main_pattern}(c), which shows the sparsity profile for meta-sparsity settings, the sparsity levels appear more stable and less variable over iterations than fixed sparsity settings.
The lines are relatively flat, indicating that once the sparsity level is set, it doesn't change much during the training for a very long time. 
This could suggest that meta-sparsity leads to a more consistent sparsity pattern, resulting in the consistent performance of the sparse model.
A stable sparsity pattern may indicate that the model has learned a general representation that is not overly fitted to the noise or eccentricity of the training data, which can lead to better generalization on unseen data (or tasks).

\textbf{Structured vs unstructured sparsity:} In this work, we have focused on meta-sparsity within the context of a group (channel-wise) or structured sparsity. 
We propose that meta-sparsity is a versatile concept, extendable to various forms of sparsity—be it structured, unstructured, penalty-driven, or any other pruning methods controlled by hyperparameters (denoted by $\lambda$ in this case). 
To support our claim, we extended the use of meta-sparsity to include fine-grained $l_1$ (unstructured) sparsity under various experimental scenarios: single-task fixed sparsity, multi-task fixed sparsity, and multi-task meta-sparsity.

\begin{figure}[h!]
    \centering
    \begin{minipage}{0.31\textwidth}
        \centering
        \includegraphics[width=\textwidth]{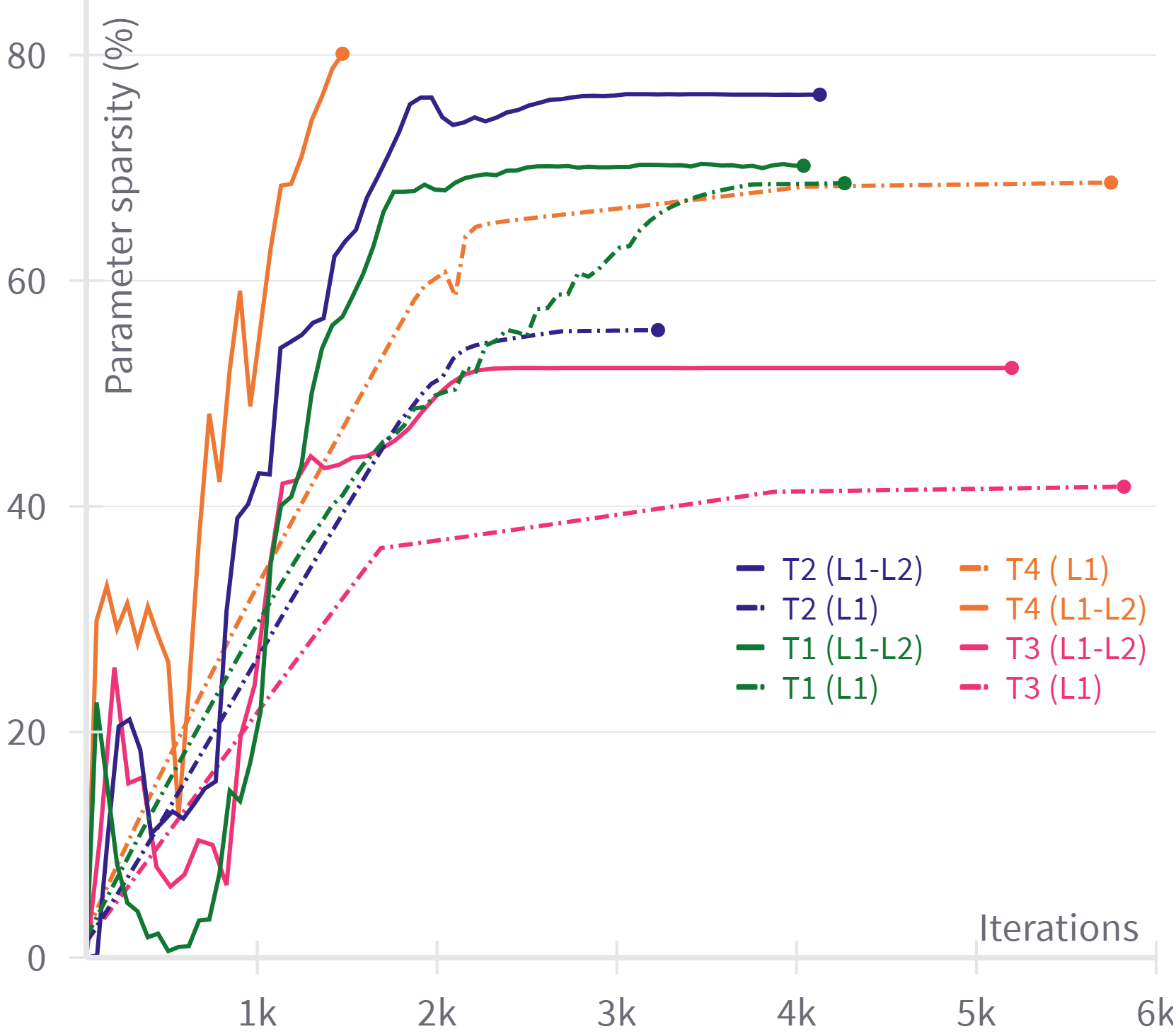} 
        \caption*{(a) single-task fixed sparsity}
        \label{fig:pattern_singletask_l1l2}
    \end{minipage}
    \begin{minipage}{0.31\textwidth}
        \centering
        \includegraphics[width=\textwidth]{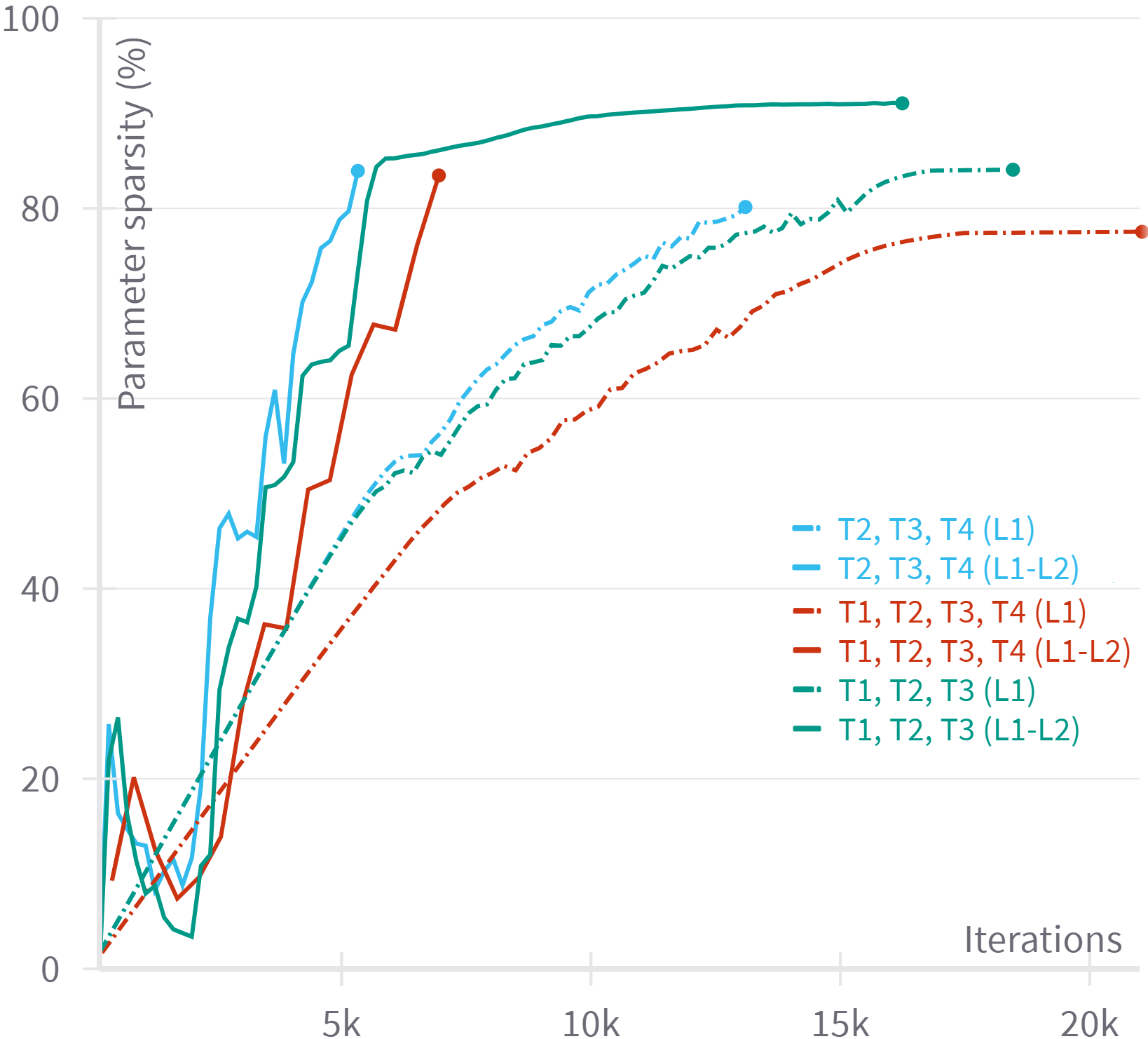} 
        \caption*{(b) multi-task fixed sparsity}
        \label{fig:pattern_fixed_l1l2}
    \end{minipage}
    \begin{minipage}{0.31\textwidth}
        \centering
        \includegraphics[width=\textwidth]{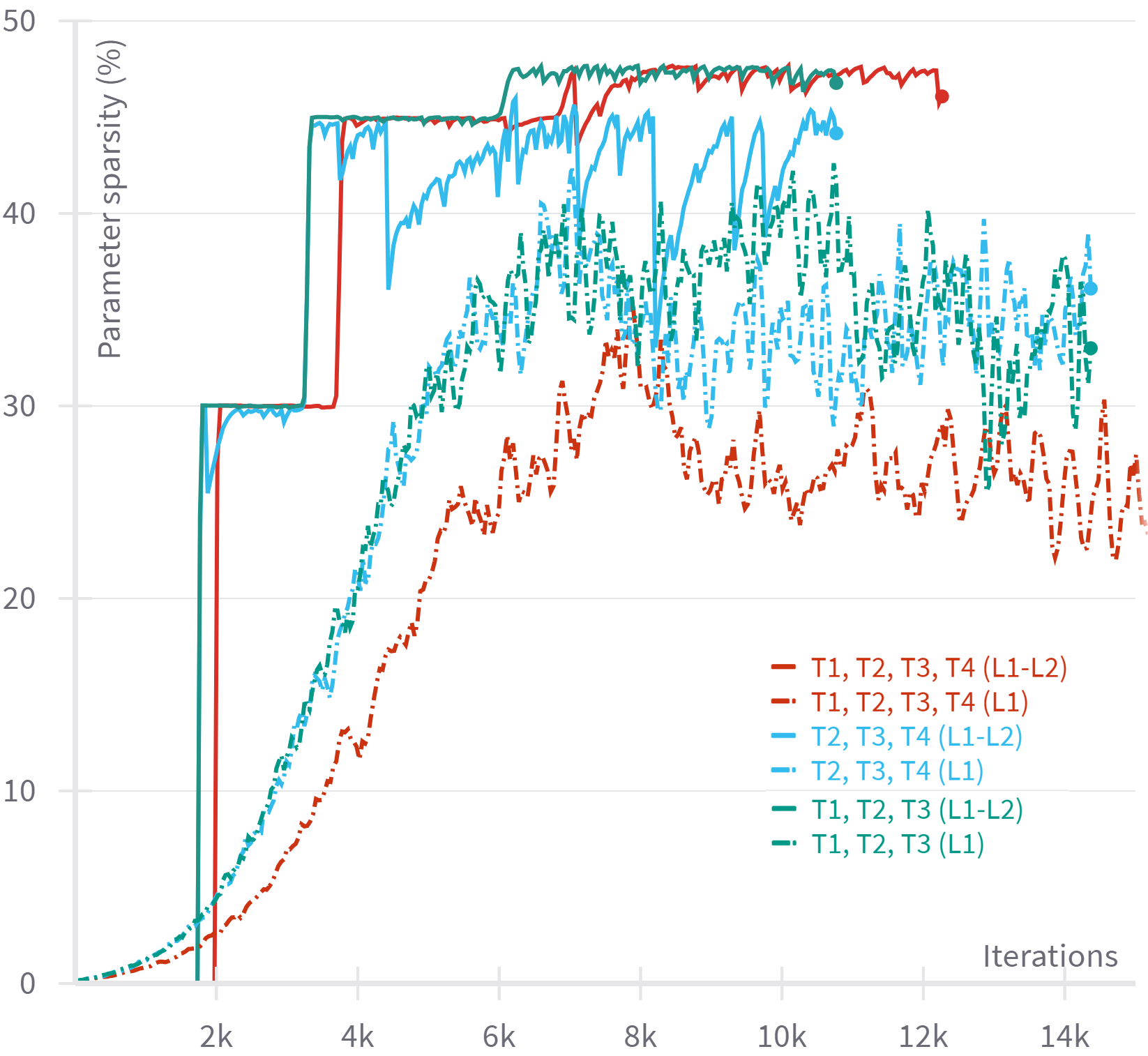} 
        \caption*{(c) multi-task meta sparsity}
        \label{fig:pattern_learn_l1l2}
    \end{minipage}
    \caption{Parameter sparsity patterns for structured ($l_1-l_2$ channel-wise group sparsity, solid lines) and unstructured sparsity (fine-grained $l_1$ sparsity, `\_.' lines) while training for (a) single-task fixed sparsity, (b) multi-task fixed sparsity, and (c) multi-task meta sparsity experiments. In the above plots, T1 represents semantic segmentation, T2 represents depth estimation, T3 represents surface normal estimation, and T4 represents edge detection tasks for the NYU dataset.}
    \label{fig:l1l2-l1}
\end{figure}

Figure~\ref{fig:l1l2-l1} shows the sparsity profile under $l_1-l_2$ and $l_1$ type penalty-based sparsity settings.
Figures~\ref{fig:l1l2-l1}(a) and (b) specifically address fixed sparsity scenarios, where the comparison reveals that models under $l_1$ sparsity achieve sparsity levels comparable to those subjected to group sparsity ($l_1-l_2$). 
The increase of parameter sparsity in models with $l_1$ regularization follows a linear steady pattern.  
In contrast, models with $l_1-l_2$ regularization exhibit initial volatility in parameter sparsity, characterized by quick increases or variations.
This is mainly because parameter groups are collectively deactivated in the structured sparsity approach. 
The performance of the tasks under structured and unstructured sparsity is mostly similar, with the exceptions of a few tasks performing well in a structured setting; see Tables 1 and 2 in the Appendix for a detailed comparison of the task-wise performance. 

In the case of the meta-sparsity setting \ie Figure~\ref{fig:l1l2-l1}(c), it is evident that the structured sparsity gives stable sparsity patterns, while those for unstructured sparsity are highly variable with too many fluctuations. 
The stability could be beneficial in practical applications where consistent performance is crucial.
The variability in unstructured sparsity may reflect a continuous adaptation to the learning task, which could be advantageous in non-stationary environments where the model needs to constantly adjust.

Overall, we show that the proposed approach of meta-sparsity can learn both structured and unstructured sparsity.
However, the choice between these forms of sparsity depends on the specific requirements of the use case at hand.
While unstructured sparsity often leads to greater model compression by zeroing out more parameters, this may come at the cost of network performance (\cite{review_sparsity}).
Structured sparsity, on the other hand, is better suited to the needs of current hardware designs.
By zeroing out entire channels or filters, the sparse network architecture can be implemented more effectively on the current hardware accelerators like GPUs or specialized ASICs.
When combined with hardware acceleration, structured sparsity can significantly reduce computation time.
On the other hand, unstructured sparsity may not result in computational speedups without specific hardware support since the remaining non-zero values are dispersed throughout the matrix, prohibiting the efficient use of vectorized operations.

\textbf{Regrowing the parameters:}
As discussed earlier, regrowing refers to the process of systematically reintroducing the sparsified parameters, which allows the network to recover from over-sparsification. 
While this concept has been outlined in  Section~\ref{sec:method}, it should be noted that the empirical results presented did not involve any regrowth (i.e., $r_p = 0$).
The rationale behind this was to focus our analysis on evaluating the effectiveness of the proposed meta-sparsity approach.
Adding regrowth would require adjusting another hyperparameter, which could affect the clarity of our findings on meta-sparsity.
\input{tables/regrow}

\begin{figure}[h!]
\centering
\begin{subfigure}{.45\textwidth}
  \centering
  \includegraphics[width=0.9\linewidth]{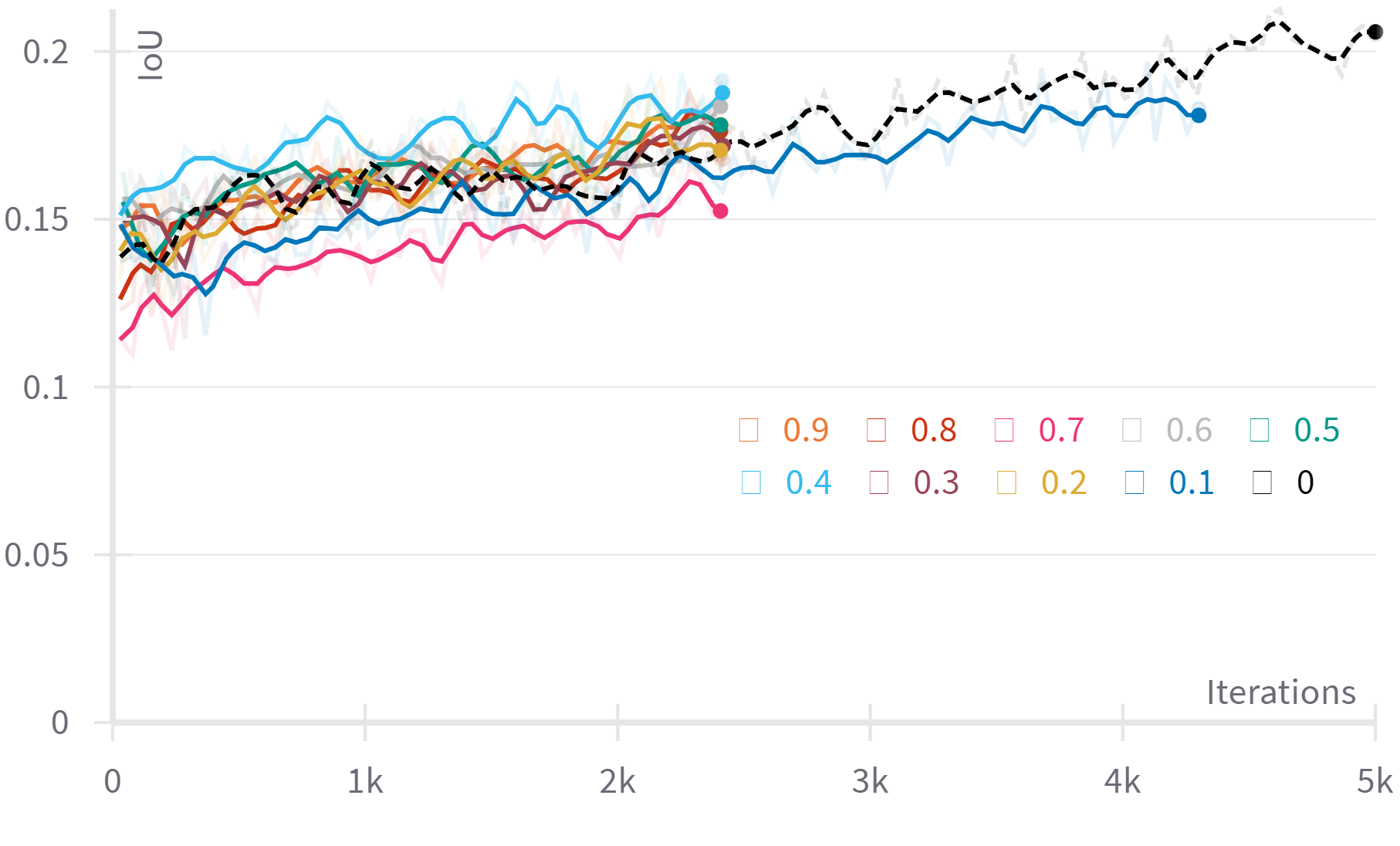}
  \caption{Segmentation}
\end{subfigure}%
\begin{subfigure}{.45\textwidth}
  \centering
  \includegraphics[width=0.9\linewidth]{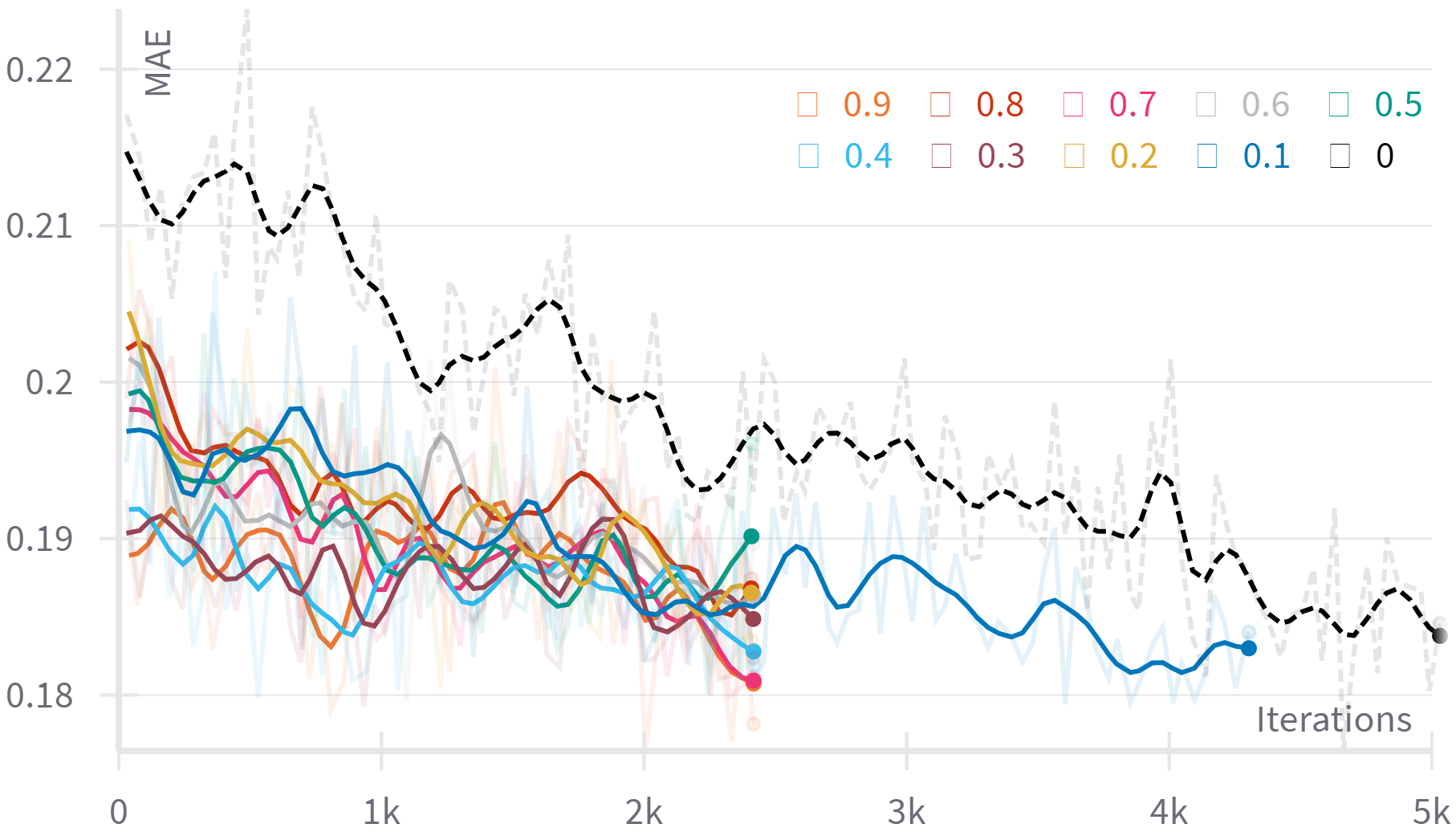}
  \caption{Depth estimation}
\end{subfigure}
\begin{subfigure}{.45\textwidth}
  \centering
  \includegraphics[width=0.9\linewidth]{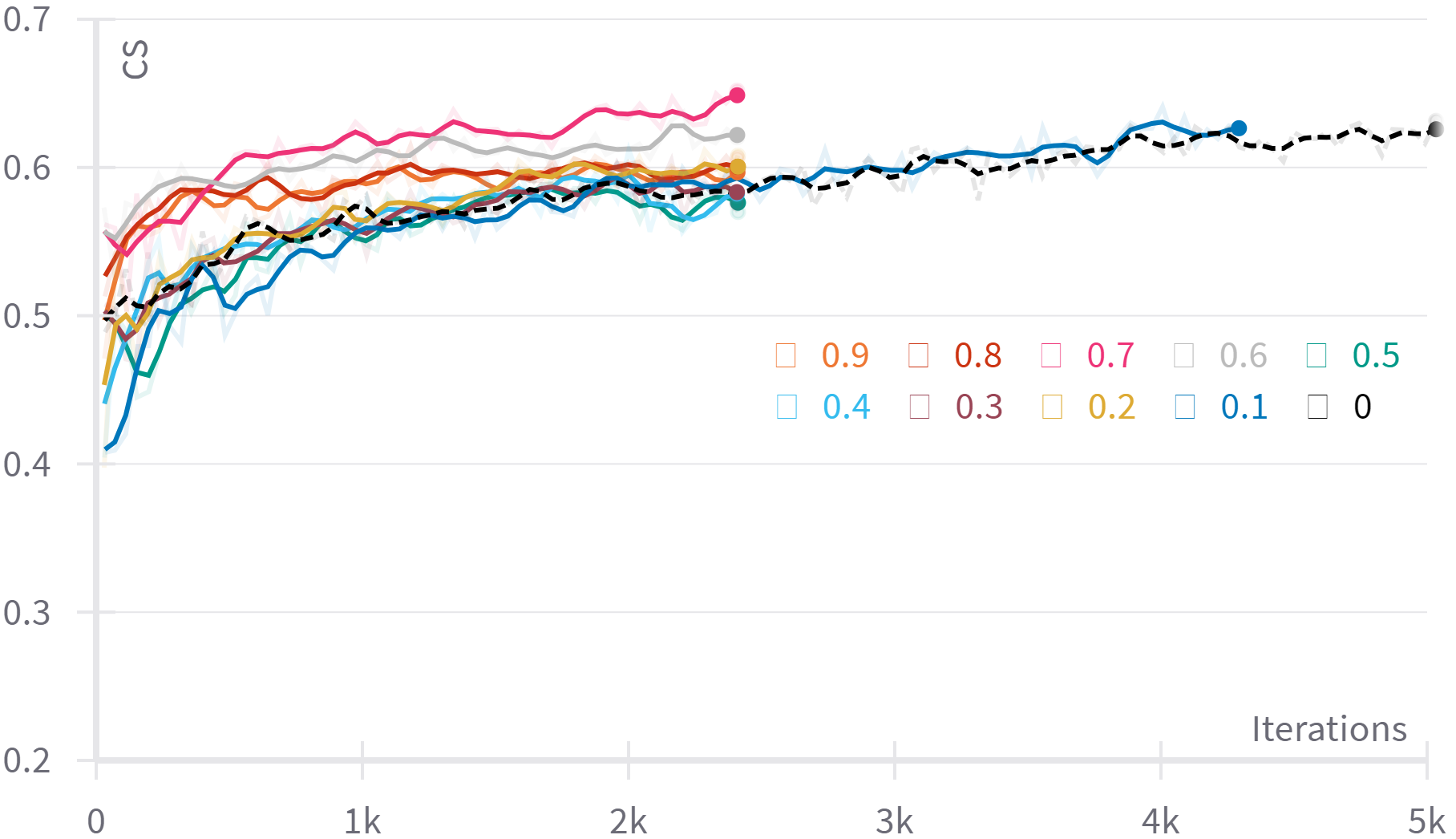}
  \caption{SN estimation}
\end{subfigure}%
\begin{subfigure}{.45\textwidth}
  \centering
  \includegraphics[width=0.9\linewidth]{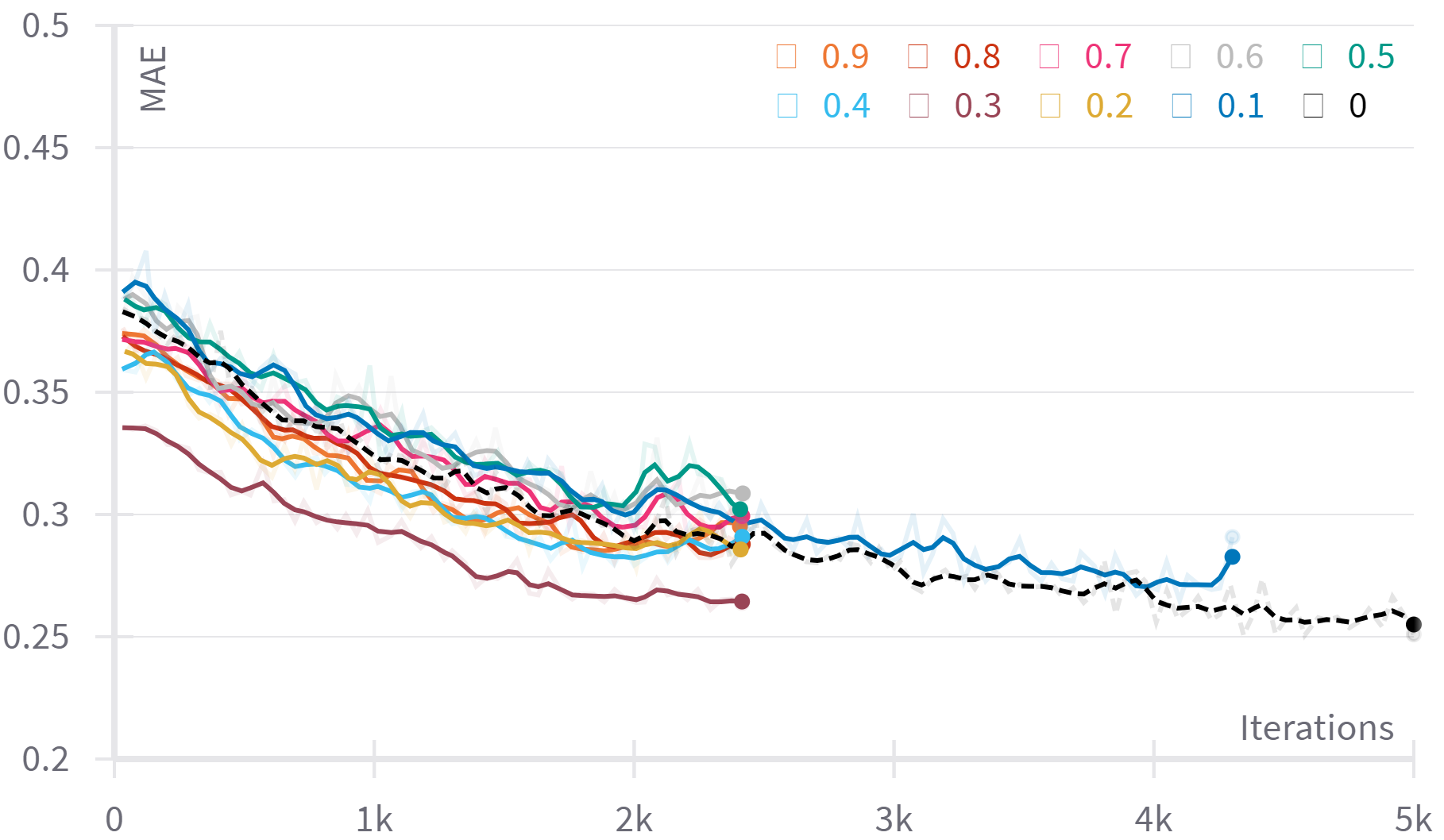}
  \caption{Edge detection}
\end{subfigure}
\caption{Comparison of the no-regrow (for $r_p$ =0, in black dotted lines) vs regrow (for $r_p$ = 0.1, 0.2, 0.3,.., 0.9) on the performance of the tasks during meta-training.}
\label{fig:rp}
\end{figure}

Figure~\ref{fig:rp} illustrates the subtle impact of the regrowth rate on the meta-training phase of the different tasks. 
It is evident from the plots that the presence of regrowth leads to faster convergence, which indicates that reintroducing the parameters helps the model quickly find the optimal parameters.
Although there are not very significant performance gains during meta-training. 
The test performance of some of these models that incorporated regrowth is shown in Table~\ref{tab:regrow}, along with the \% parameter sparsity.
This table compares the no-regrow setting to the regrow settings with $r_p = 0.2, 0.4, 0.6, 0.8$.
Analysis of the table reveals minimal performance improvements observed with different regrowth rates, except for the segmentation task. 
However, the model with regrowth is relatively less sparse than the one without regrowth. 
Figure~\ref{fig:rp} and Table~\ref{tab:regrow} suggest that an optimal regrowth probability may exist, which balances the trade-offs between model compression and learning efficacy.





%% file: tables/nyu_mtml.tex

\begin{table}[h]
\centering
\caption{Performance comparison between the proposed meta-sparsity approach (with meta-learnable $\lambda$) and a conventional meta-learning baseline that only learns parameter initialization in an MTL setting, evaluated on the NYU dataset. The baseline incorporates both multi-task and meta-learning approaches while omitting sparsification. This table compares the performance of the tasks for NYU dataset when a new unseen task is added during the meta-testing stage. The new tasks are highlighted in blue, and the best (mean value) metrics for each task are in bold.}
\label{tab:NYU_add}
\resizebox{0.95\columnwidth}{!}{%
\begin{tabular}{ccccccc}
\hline
{\textbf{Meta-}} & \rule{0pt}{3ex}\textbf{Meta-} & \textbf{Segmentation} & \textbf{Depth est.} & \textbf{SN est.} & \textbf{Edge det.} & \textbf{Parameter} \\
 \textbf{training}& \textbf{testing} & \textbf{$T_1$} & \textbf{$T_2$} & \textbf{$T_3$} & \textbf{$T_4$} & \textbf{sparsity} \\
 &  & IoU($\uparrow$) & MAE($\downarrow$) & CS($\uparrow$) & MAE($\downarrow$) & \textbf{(\%)} \\ \hline
\multicolumn{7}{c}{\rule{0pt}{3ex} \textbf{  MTL + meta-learning}   (without any sparsity in the shared backbone, $\lambda$ = 0) \rule[-1.5ex]{0pt}{0pt}} \\ \hline
\rule{0pt}{3ex} $T_1$,$T_2$,$T_3$ & $T_1$,$T_2$,$T_3$ & \multicolumn{1}{l}{0.2750 ± 0.0109} & \multicolumn{1}{l}{0.1449 ± 0.0006} & \multicolumn{1}{l}{0.7263 ± 0.0009} &  & 0 \\
$T_2$,$T_3$,$T_4$ & $T_2$,$T_3$,$T_4$ &  & \multicolumn{1}{l}{0.1563 ± 0.0048} & \multicolumn{1}{l}{0.7278 ± 0.0018} & \multicolumn{1}{l}{0.1414 ± 0.0037} & 0 \\
$T_1$,$T_2$,$T_3$,$T_4$ & $T_1$,$T_2$,$T_3$,$T_4$ & \multicolumn{1}{l}{0.2769 ± 0.0039} & \multicolumn{1}{l}{0.1476 ± 0.0050} & \multicolumn{1}{l}{0.7235 ± 0.0017} & \multicolumn{1}{l}{0.1441 ± 0.0030} & 0 \\ 
\multicolumn{7}{c}{\rule{0pt}{3ex} \textcolor{blue}{Adding unseen tasks during meta-testing in \textbf{MTL+meta-learning}} \rule[-2ex]{0pt}{0pt}} \\ 
$T_1$,$T_2$,$T_3$ & $T_1$,$T_2$,$T_3$,\textcolor{blue}{+$T_4$} & 0.2796 $\pm$ 0.0053 & 0.1467 $\pm$ 0.0033 & 0.7257 $\pm$ 0.0050 & \textbf{0.1380 $\pm$ 0.0009} &0 \\
$T_1$,$T_2$,$T_3$ & finetuning only \textcolor{blue}{$T_4$} & \multicolumn{1}{c}{-} & \multicolumn{1}{c}{-} & \multicolumn{1}{c}{-} & 0.1541 $\pm$ 0.0028 & 0 \\
 &  &  &  &  &  &  \\
$T_2$,$T_3$,$T_4$ & $T_2$,$T_3$,$T_4$,\textcolor{blue}{+$T_1$} & 0.2704 $\pm$ 0.0064 & 0.1497 $\pm$ 0.0059 & 0.7237 $\pm$ 0.0048 & 0.1428 $\pm$ 0.0024 & 0 \\
$T_2$,$T_3$,$T_4$ & finetuning only \textcolor{blue}{$T_1$} & 0.2619 $\pm$ 0.0038 & \multicolumn{1}{c}{-} & \multicolumn{1}{c}{-} & \multicolumn{1}{c}{-} & 0 \\
\hline
\multicolumn{7}{c}{\rule{0pt}{3ex} \textbf{Meta-Sparsity} (MTL+ meta-learning + group sparsity, $\lambda$ = meta-learnable ) \rule[-1.5ex]{0pt}{0pt}} \\ \hline
\rule{0pt}{3ex} $T_1$,$T_2$,$T_3$ & $T_1$,$T_2$,$T_3$ & 0.2937 ± 0.0091 & 0.1409 ± 0.0035 & \textbf{0.7460 ± 0.0073} &  & 43.4391 ± 0.1925 \\
$T_2$,$T_3$,$T_4$ & $T_2$,$T_3$,$T_4$ &  & 0.1412 ± 0.0040 & 0.7402 ± 0.0080 & 0.1386 ± 0.0019 & 43.7577 ± 0.1886 \\
$T_1$,$T_2$,$T_3$,$T_4$ & $T_1$,$T_2$,$T_3$,$T_4$  & 0.2923 ± 0.0101 & 0.1397 ± 0.0037 & 0.7414 ± 0.0055 & 0.1395 ± 0.0021 & 44.1159 ± 0.3696 \\ 
\multicolumn{7}{c}{\rule{0pt}{3ex} \textcolor{blue}{Adding unseen tasks during meta-testing in \textbf{Meta-Sparsity}} \rule[-2ex]{0pt}{0pt}} \\ 
$T_1$,$T_2$,$T_3$ & $T_1$,$T_2$,$T_3$,\textcolor{blue}{+$T_4$} & \textbf{0.3003 $\pm$ 0.0067} & 0.1389 $\pm$ 0.0030 & 0.7438 $\pm$ 0.0005 & 0.1404 $\pm$ 0.0015 & 43.4391 $\pm$ 0.1925 \\
$T_1$,$T_2$,$T_3$ & finetuning only \textcolor{blue}{$T_4$} & \multicolumn{1}{c}{-} & \multicolumn{1}{c}{-} & \multicolumn{1}{c}{-} & 0.1431 $\pm$ 0.0020 & 43.4391 $\pm$ 0.1925 \\
 &  &  &  &  &  &  \\
$T_2$,$T_3$,$T_4$ & $T_2$,$T_3$,$T_4$,\textcolor{blue}{+$T_1$} & 0.2950 $\pm$ 0.0069 & \textbf{0.1378 $\pm$ 0.0020} & 0.7444 $\pm$ 0.0017 & 0.1386 $\pm$ 0.0003 & 43.7577 $\pm$ 0.1886 \\
$T_2$,$T_3$,$T_4$ & finetuning only \textcolor{blue}{$T_1$} & 0.2828 $\pm$ 0.0392 & \multicolumn{1}{c}{-} & \multicolumn{1}{c}{-} & \multicolumn{1}{c}{-} & 43.7577 $\pm$ 0.1886 \\

\hline
\end{tabular}%
}
\end{table}


\begin{table}[ht]
\centering
\caption{Performance comparison between the proposed meta-sparsity approach (with meta-learnable $\lambda$) and a conventional meta-learning baseline that only learns parameter initialization in an MTL setting, evaluated on the celebA dataset. The baseline incorporates both multi-task and meta-learning approaches while omitting sparsification. This table compares the performance of the tasks for celebA dataset when a new unseen task is added during the meta-testing stage. The new tasks are highlighted in blue, and the best (mean value) metrics for each task are in bold. }
\label{tab:celeb_add}
\resizebox{\columnwidth}{!}{%
\begin{tabular}{cccccccccc}
\hline
\textbf{Meta-} & \textbf{Meta-} & \textbf{Segmen-} & \textbf{male} & \textbf{smile} & \textbf{biglips} & \textbf{highcheek} & \textbf{wearing} & \textbf{bushy} & \textbf{Parameter} \\
\textbf{training} & \textbf{testing} & \textbf{tation} & \textbf{} & \textbf{} & \textbf{} & \textbf{bones} & \textbf{lipstick} & \textbf{eyebrows} & \textbf{sparsity} \\
 &  & \textbf{$T_1$} & \textbf{$T_2$} & \textbf{$T_3$} & \textbf{$T_4$} & \textbf{$T_5$} & \textbf{$T_6$} & \textbf{$T_7$} & \textbf{} \\
 &  & IoU($\uparrow$) & Accuracy($\uparrow$) & Accuracy($\uparrow$) & Accuracy($\uparrow$) & Accuracy($\uparrow$) & Accuracy($\uparrow$) & Accuracy($\uparrow$) & (\%) \\ \hline
\multicolumn{10}{c}{\rule{0pt}{3ex}  \textbf{MTL + meta-learning }  (without any sparsity in the shared backbone, $\lambda$ = 0)\rule[-1.5ex]{0pt}{0pt}} \\ \hline
\rule{0pt}{3ex} $T_1$-$T_7$ & $T_1$-$T_7$ & \multicolumn{1}{l}{0.8922±0.0067} & \multicolumn{1}{l}{0.8467±0.0108} & \multicolumn{1}{l}{0.6446±0.0021} & \multicolumn{1}{l}{0.6010±0.0013} & \multicolumn{1}{l}{0.6272±0.0018} & \multicolumn{1}{l}{0.9138±0.0020} & \multicolumn{1}{l}{0.5993±0.0510} & 0 \\
$T_2$-$T_7$ & $T_2$-$T_7$ & \multicolumn{1}{l}{} & \multicolumn{1}{l}{0.6297±0.0100} & \multicolumn{1}{l}{0.5071±0.0022} & \multicolumn{1}{l}{0.5993±0.0012} & \multicolumn{1}{l}{0.5221±0.0010} & \multicolumn{1}{l}{0.7560±0.0078} & \multicolumn{1}{l}{0.6053±0.0480} & 0 \\
$T_1$,$T_2$,$T_3$,$T_7$ & $T_1$,$T_2$,$T_3$,$T_7$ & \multicolumn{1}{l}{0.8942±0.0010} & \multicolumn{1}{l}{0.6297±0.0100} & \multicolumn{1}{l}{0.5261±0.0030} & \multicolumn{1}{l}{} & \multicolumn{1}{l}{} & \multicolumn{1}{l}{} & \multicolumn{1}{l}{0.6103±0.0501} & 0 \\ 
\multicolumn{10}{c}{\rule{0pt}{3ex} \textcolor{blue}{Adding unseen tasks during meta-testing in \textbf{MTL+meta-learning}} \rule[-2ex]{0pt}{0pt}} \\ 
$T_2$-$T_7$ & $T_2$-$T_7$, \textcolor{blue}{+$T_1$} & 0.8731$\pm$0.0159 & 0.9479$\pm$0.0136 & 0.5261$\pm$0.0024 & 0.6140$\pm$0.0014 & 0.5905$\pm$0.0118 & 0.9127$\pm$0.0080 & 0.6223$\pm$0.0173 & 0 \\
\multirow{2}{*}{$T_2$-$T_7$} & finetune & \multirow{2}{*}{0.8913$\pm$0.0015} & \multirow{2}{*}{-} & \multirow{2}{*}{-} & \multirow{2}{*}{-} & \multirow{2}{*}{-} & \multirow{2}{*}{-} & \multirow{2}{*}{-} & \multirow{2}{*}{0} \\
 & only \textcolor{blue}{$T_1$} &  &  &  &  &  &  &  &  \\
\multirow{2}{*}{$T_1$,$T_2$,$T_3$,$T_7$} & $T_1$,$T_2$,$T_3$,$T_7$, & \multirow{2}{*}{0.8944$\pm$0.0022} & \multirow{2}{*}{0.9494$\pm$0.0257} & \multirow{2}{*}{0.5261$\pm$0.0016} & \multirow{2}{*}{\textbf{0.6340$\pm$0.0031}} & \multirow{2}{*}{0.5221$\pm$0.0095} & \multirow{2}{*}{\textbf{0.9205$\pm$0.0098}} & \multirow{2}{*}{0.6223$\pm$0.0105} & \multirow{2}{*}{0} \\
 & \textcolor{blue}{+ $T_4$,$T_5$,$T_6$} &  &  &  &  &  &  &  &  \\
\multirow{2}{*}{$T_1$,$T_2$,$T_3$,$T_7$} & finetune only & \multirow{2}{*}{-} & \multirow{2}{*}{-} & \multirow{2}{*}{-} & \multirow{2}{*}{0.6290$\pm$0.0021} & \multirow{2}{*}{0.5221$\pm$0.0015} & \multirow{2}{*}{0.7620$\pm$0.0306} & \multirow{2}{*}{-} & \multirow{2}{*}{0} \\
 & \textcolor{blue}{$T_4$,$T_5$,$T_6$} &  &  &  &  &  &  &  &  \\
 &  &  &  &  &  &  &  &  &  \\

\hline
\multicolumn{10}{c}{\rule{0pt}{3ex}  \textbf{Meta-Sparsity} (MTL+   meta-learning + group sparsity)\rule[-1.5ex]{0pt}{0pt}} \\ \hline
\rule{0pt}{3ex} $T_1$-$T_7$ & $T_1$-$T_7$& 0.9007±0.0012 & \textbf{0.9669±0.0014} & \textbf{0.9014±0.0043} & 0.6330±0.1680 & 0.8235±0.0104 & 0.9068±0.0165 & 0.6885±0.0936 & 43.6893±0.3593 \\
$T_2$-$T_7$ & $T_2$-$T_7$  &  & 0.9004±0.0203 & 0.7618±0.1602 & 0.6330±0.0720 & 0.7562±0.1109 & 0.8703±0.0154 & 0.6223±0.0002 & 30.2145±0.0016 \\
$T_1$,$T_2$,$T_3$,$T_7$ & $T_1$,$T_2$,$T_3$,$T_7$  & \textbf{0.9034±0.0013} & 0.7386±0.0188 & 0.5261±0.0010 &  &  &  & 0.6223±0.0010 & 30.0187±0.0011 \\ 
\multicolumn{10}{c}{\rule{0pt}{3ex} \textcolor{blue}{Adding unseen tasks during meta-testing in \textbf{Meta-Sparsity}} \rule[-2ex]{0pt}{0pt}} \\ 
$T_2$-$T_7$ & $T_2$-$T_7$, \textcolor{blue}{+$T_1$} & 0.8985$\pm$0.0027 & 0.9615$\pm$0.0051 & 0.8977$\pm$0.0072 & 0.6330$\pm$0.0091 & \textbf{0.8278$\pm$0.0057} & 0.9102$\pm$0.0144 & \textbf{0.7529$\pm$0.0199} & 30.2145$\pm$0.0016 \\
\multirow{2}{*}{$T_2$-$T_7$} & finetune & \multirow{2}{*}{0.9033$\pm$0.0019} & \multirow{2}{*}{-} & \multirow{2}{*}{-} & \multirow{2}{*}{-} & \multirow{2}{*}{-} & \multirow{2}{*}{-} & \multirow{2}{*}{-} & \multirow{2}{*}{30.2145$\pm$0.0016} \\
 & only \textcolor{blue}{$T_1$} &  &  &  &  &  &  &  &  \\
\multirow{2}{*}{$T_1$,$T_2$,$T_3$,$T_7$} & $T_1$,$T_2$,$T_3$,$T_7$, & \multirow{2}{*}{0.9019$\pm$0.0005} & \multirow{2}{*}{0.9559$\pm$0.0191} & \multirow{2}{*}{0.8995$\pm$0.0116} & \multirow{2}{*}{0.6330$\pm$0.0130} & \multirow{2}{*}{0.8206$\pm$0.0066} & \multirow{2}{*}{0.9073$\pm$0.0070} & \multirow{2}{*}{0.7250$\pm$0.0890} & \multirow{2}{*}{30.0187$\pm$0.0011} \\
 & \textcolor{blue}{+ $T_4$,$T_5$,$T_6$} &  &  &  &  &  &  &  &  \\
\multirow{2}{*}{$T_1$,$T_2$,$T_3$,$T_7$} & finetune only & \multirow{2}{*}{-} & \multirow{2}{*}{-} & \multirow{2}{*}{-} & \multirow{2}{*}{0.6330$\pm$0.0011} & \multirow{2}{*}{0.7373$\pm$0.0117} & \multirow{2}{*}{0.8812$\pm$0.0096} & \multirow{2}{*}{-} & \multirow{2}{*}{30.0187$\pm$0.0011} \\
 & \textcolor{blue}{$T_4$,$T_5$,$T_6$} &  &  &  &  &  &  &  &  \\
 &  &  &  &  &  &  &  &  &  \\

\hline
\end{tabular}%
}
\end{table}

%% file: tables/baseline_comp.tex
\begin{table}[]
\centering
\caption{Performance analysis of the sparsification approaches presented in Figure~\ref{fig:intro_fig}, for the NYU-v2 datset.}
\label{tab:comp_performance}
\resizebox{0.98\columnwidth}{!}{%
\begin{tabular}{lllcccc}
\hline
\textbf{Sparsification} & \textbf{Sparsity} & \textbf{parameter} & \textbf{Segmentation} & \textbf{Depth est.} & \textbf{SN est.} & \textbf{Edge det.} \\
\textbf{approaches} & \textbf{patterns} & \textbf{init. before} & \textbf{$T_1$} & \textbf{$T_2$} & \textbf{$T_3$} & \textbf{$T_4$} \\
\textbf{} & \textbf{} & \textbf{sparsification} & \textbf{IoU(↑)} & \textbf{MAE(↓)} & \textbf{CS(↑)} & \textbf{MAE(↓)} \\ \hline
\multirow{3}{*}{One-shot} & Mask-1, magnitude & MTL dense & 0.2345 $\pm$ 0.0008 & 0.1777 $\pm$ 0.0097 & 0.6601 $\pm$ 0.0250 & 0.3069 $\pm$ 0.0325 \\
 & Mask-3, random & MTL dense & 0.3185 $\pm$ 0.0010 & 0.1337 $\pm$ 0.0018 & 0.7412 $\pm$ 0.0027 & \textbf{0.1318 $\pm$ 0.0026} \\
 & Meta mask & MTL dense & 0.2308 $\pm$ 0.0106 & 0.1682 $\pm$ 0.0065 & 0.6897 $\pm$ 0.0062 & 0.3036 $\pm$ 0.0214 \\ \hline
\multirow{2}{*}{Iterative} & Mask-2, iterative magnitude & MTL dense & 0.3082 $\pm$ 0.0094 & 0.1367 $\pm$ 0.0014 & 0.7346 $\pm$ 0.0043 & 0.1353 $\pm$ 0.0008 \\
 & Meta mask & MTL dense & 0.3008 $\pm$ 0.0080 & 0.1373 $\pm$ 0.0007 & 0.7314 $\pm$ 0.0027 & 0.1356 $\pm$ 0.0008 \\ \hline
\multirow{3}{*}{Progressive} & Mask-2, iterative magnitude & random & 0.3855 $\pm$ 0.0139 & 0.1491 $\pm$ 0.0209 & 0.7800 $\pm$ 0.0102 & 0.1957 $\pm$ 0.0161 \\
 & Mask-3, random & random & 0.3740 $\pm$ 0.0155 & 0.1361 $\pm$ 0.0040 & 0.7772 $\pm$ 0.0079 & 0.2176 $\pm$ 0.0228 \\
 & Meta mask & random & \textbf{0.4018 $\pm$ 0.0082} & 0.1391 $\pm$ 0.0007 & \textbf{0.7877 $\pm$ 0.0060} & 0.1978 $\pm$ 0.0017 \\ \hline
\multirow{3}{*}{Sparse training} & Mask-1 , magnitude & random & 0.2918 $\pm$ 0.0112 & 0.1488 $\pm$ 0.0049 & 0.7202 $\pm$ 0.0058 & 0.2289 $\pm$ 0.0035 \\
 & Mask-3, random & random & 0.2959 $\pm$ 0.0077 & 0.1530 $\pm$ 0.0009 & 0.7180 $\pm$ 0.0033 & 0.2389 $\pm$ 0.0105 \\
 & Meta mask & random & 0.3906 $\pm$ 0.0087 & \textbf{0.1282 $\pm$ 0.0050} & 0.7827 $\pm$ 0.0022 & 0.2001 $\pm$ 0.0102 \\ \hline
Meta-sparsity & Meta-mask/pattern & random & 0.2923 $\pm$ 0.0101 & 0.1397 $\pm$ 0.0037 & 0.7414 $\pm$ 0.0055 & 0.1395 $\pm$ 0.0021 \\ \hline
 &  &  & \multicolumn{1}{l}{} & \multicolumn{1}{l}{} & \multicolumn{1}{l}{} & \multicolumn{1}{l}{} \\
\multicolumn{1}{r}{\textit{Note:}} & \multicolumn{6}{l}{Mask-1, one-shot sparsification by eliminating the lowest magnitude weights(\cite{PhysRevA.39.6600}).} \\
 & \multicolumn{6}{l}{Mask-2, iterative magnitude sparsification by eliminating the lowest magnitude weights in steps.} \\
 & \multicolumn{6}{l}{Mask-3, random sparsification.} \\
 & \multicolumn{6}{l}{Meta-mask is the optimal sparsity pattern/mask learned by the meta-sparsity experiment.} \\
 & \multicolumn{6}{l}{The \% parameter sparsity is kept constant across all the experiments, equal to the meta-sparsity achieved, i.e.,$\sim$44.11\%.}
 \vspace{-2em}
\end{tabular}%
}
\end{table}

%% file: tables/regrow.tex
\begin{table}[h!]
\centering
\caption{Average performance for the four task combination \ie $(T_1, T_2, T_3, T_4)$ of the NYU dataset under the meta-sparsity setting for various values of the regrowth parameter, $r_p$. The regrow parameter, say $r_p = x$, signifies that there is a $x\%$ chance that any sparsified channel will be regrown or reintroduced. In this work, we use Xavier initialization (\cite{pmlr-v9-glorot10a}) to set the new values of the regrown filters. $ r_p = 0$ represents the case when none of the parameters is re-grown during meta-training.}
\label{tab:regrow}
\begin{tabular}{cccccc}
\hline

Regrow & Segmentation & Depth est. & SN est. & Edge det. & Parameter \\
prob. & $T_1$ & $T_2$ & $T_3$ & $T_4$ & sparsity \\ 
$r_p$ & IoU($\uparrow$) & MAE($\downarrow$) & CS($\uparrow$) & MAE($\downarrow$) & (\%) \\ \hline
&&&&&\\
0 (no regrow) & 0.2923 & 0.1397 & 0.7414 & 0.1395 & 44.1159 \\
0.2 & 0.3177 & 0.1339 & 0.7514 & 0.1398 & 30.1157 \\
0.4 & 0.3100 & 0.1409 & 0.7437 & 0.1405 & 30.0186 \\
0.6 & 0.3137 & 0.1356 & 0.7500 & 0.1382 & 29.9991 \\
0.8 & 0.3170 & 0.1351 & 0.7472 & 0.1409 & 31.0982 \\ \hline
\end{tabular}%
\end{table}

%% file: sections/7_conclusion.tex
This study demonstrates that model sparsity can be a learnable attribute rather than a feature determined by heuristic hyperparameters.
Our proposed framework for learned sparsity, also referred to as \textit{meta-sparsity}, shows that models may be (meta-)trained to naturally adopt sparse structures, eliminating the need for manual tuning of sparsity levels.
Our outcomes show that this strategy can produce models that are not only efficient and compact but also perform well on a variety of tasks.
Although the segmentation task for the NYU dataset, in particular, poses a challenge due to its inherent complexity, such as dense pixel maps of around 40 segmentation classes, the meta-sparsity framework has proved robust. 
This study especially focused on the \ac{MAML} framework within the context of meta-learning.  
However, it is acknowledged that various advanced extensions of MAML or similar gradient-based meta-learning algorithms can potentially improve performance and are viable alternatives for implementation in this work.
The aim of this study extends beyond merely enhancing the performance of the tasks; it proposes a concept that could lead to the development of parsimonious models.

Theoretically, the proposed meta-sparsity approach is versatile and can be applied to any number and types of tasks, as well as various types of sparsity. 
Based on the experiments, we have identified a few scenarios where the method may prove to be highly effective-
\begin{itemize}
    \item \textit{Similar or closely related tasks:}  The definition of similar tasks is very subjective; however,  in the context of this work, they can consider them as tasks that may require some similar set of features. Our experiments indicate that meta-sparsity performs best when the tasks are similar or closely related, like depth estimation, surface normal estimation, segmentation, and edge detection for the NYU-v2 dataset. In such cases, the approach is better able to identify and leverage optimal shared sparse patterns that benefit all tasks involved. When tasks are highly diverse, it may be more challenging to find an optimal shared sparse pattern that effectively supports all the tasks. In such cases, the variability in the task feature requirement can make it difficult to achieve good performance for all the tasks. The performance of the task combination $T_1,T_2,T_3,T_7$ for celebA is one such example. 
    \item \textit{Inclusion of pixel-level tasks:} Adding a pixel-level task to the task mix can enhance the performance of all tasks, as observed in our experiments with the CelebA dataset ($T_1-T_7$ vs $T_2-T_7$, here $T_1$ is the only pixel level task). Pixel-level tasks require more granular features, which may also benefit image-level tasks by providing richer feature representations.
\end{itemize}

Our approach represents a step towards developing \textit{black box} sparsity, \ie allowing models to learn an optimal sparsity pattern. 
This approach is not limited to a specific model, task, or type of sparsity.
It is strengthened by the concept of meta-learning, which learns the sparsity pattern across a range of tasks and allows easy integration of new tasks to the sparse models. 
\ac{MTL} facilitates the joint training of diverse tasks within a single model.
When combined with sparsity, MTL not only helps with model compression but also enhances the task's performance by effective feature sharing between the tasks.

In continuation of this work, various directions could be pursued.
A few of them are listed below:

\textit{Structured sparsity in diverse architectures:} Although structural sparsity has demonstrated potential in networks with residual connections like ResNet architectures, its applicability to other networks remains unclear.  
Future research might examine how different types of structured or block sparsity could be customized for different network architectures.

\textit{Task transference in sparse models:} Exploring task transference in sparse multi-task models is a promising direction for further research. 
Investigating the sparse features that promote better sharing across tasks could provide new insights into multi-task networks and the impact of sparsity on cooperative learning.

\textit{Hardware efficiency of sparse models: }
It is also necessary to investigate how sparse models align with present and upcoming hardware capabilities.
The aim is to translate model compression due to sparsity into tangible computational efficiency and speed improvements on different hardware platforms.

%% file: sections/8_appendix.tex
\subsection{Datasets and tasks}\label{app:dataset}
In this work, two widely recognized and publicly accessible datasets: the NYU-v2 dataset (\cite{Silberman_ECCV12}) and the CelebAMask-HQ dataset (\cite{CelebAMask-HQ}), are used.
The NYU dataset is composed of high-resolution indoor scene images, along with dense segmentation and depth maps.
This work focuses on four dense prediction tasks that are semantic segmentation($T_1$), depth estimation($T_2$), surface normal estimation($T_3$), and edge detection($T_4$); for task definitions.
For this work, 1450 RGB images with the corresponding labels are used, and the train-validation-test data split is the same as \cite{sun2020adashare}. 
The CelebAMask-HQ dataset is a high-quality facial attribute dataset created for semantic segmentation($T_1$). 
It contains over 30,000 photos of celebrities with precisely annotated masks and covers 19 facial attributes, out of which only six are used in this work that are: male, smile, big lips, bushy eyebrows, high-cheekbones, and wearing lipstick (\ie $T_2 - T_7$).
These attributes are used for binary classification tasks. 

The task definitions are as follows- 

\begin{itemize}[leftmargin=*]
    \item[]\textbf{Semantic segmentation:} Semantic segmentation is the process of breaking down an image into many segments or areas to assign each segment to a given class or category. 
    It is a pixel-wise classification task wherein each pixel associated with a given label is classified.
    This task is vital in computer vision because it allows machines to grasp and interpret images at the pixel level, which is critical for applications like self-driving cars, medical imaging analysis, and landscape classification.   
For the NYU dataset, we have considered 39 segmentation classes, while for the celebA dataset, there are 3 classes (skin, hair, and background).

\item[]\textbf{Depth Estimation:} Distance from a pixel to the optical center of the camera is used to calculate Euclidean depth.
Estimating depth is a crucial process for locating obstacles and points of interest in a scene.
This is a pixel-wise regression task.
Depth estimation provides crucial information about the structure of the environment, allowing for the reconstruction of a 3D scene from one or more 2D images.

\item[]\textbf{Surface normal estimation:}
It involves the task of predicting the surface orientation of the objects.
The shapes of three-dimensional objects can be approximated using the surface normals in computer vision.
In robot navigation, finding the environment's orientation is often useful for predicting the type of movement, such as smooth, inclined, or step.
This is also a pixel-wise regression task that outputs continuous values that represent the geometric orientation of each point on an object's surface, typically expressed as vectors normal (perpendicular) to the surfaces.

\item[]\textbf{Edge detection:}
Edge detection is a method of image processing that identifies points in a digital image that have abrupt transitions between adjacent pixels.
In most cases, it is done to lessen the quantity of data while still maintaining the structural properties of the objects that are depicted in the image.
Edge detection has typically been a classification task, differentiating pixels as edge or non-edge, but in this paper, we approach it as a pixel-wise regression task. This adaptation makes the task more easily learnable by a deep neural network, allowing for a more sophisticated understanding and detection of edges by treating intensity gradients and orientations as continuous variables rather than discrete classes.

\item[]\textbf{Image classification:} These are binary classification tasks that categorize images into one of two predefined attributes (\ie attribute present or non-present). 
 This task is foundational for many applications that require a straightforward decision-making process, such as determining whether an image has a particular attribute (e.g., male/no male, smile/no smile, big lips/no big lips, bushy eyebrows/no bushy eyebrows, high-cheekbones/no high-cheekbones, and wearing lipstick/not wearing lipstick).

\end{itemize}


\subsection{Experimental details}


As discussed in Section 4 of the main manuscript, we conducted two broad types of experiments. 
The first set of experiments does not involve sparsity or meta-learning. 
These are standard single-task and multi-task learning experiments, where the model is initialized with random weights and trained for single or multiple tasks.
The second set of experiments incorporates sparsity ($l_1$ or $l_1$-$l_2$). 
In these experiments, the sparsity hyperparameter (i.e., the value of the $\lambda$ ) is either fixed, with models initialized using random weights before training, or learnable, as in the meta-sparsity experiments.
For the fixed sparsity experiments, we use the trained parameters of the best single-task or multi-task model for inference.
In the case of meta-sparsity experiments, $\lambda$ is learnable. 
These experiments involve meta-training with random weight initialization, followed by meta-testing using the meta-weights obtained during meta-training. 
Thus, only experiments with learnable $\lambda$ involve meta-learning.

For reproducibility, the hyperparameter values used in the experiments are as follows: 

\begin{itemize}
    \item \textbf{Batch Specifications:}
    \begin{itemize}
        \item Train batch size: 16
        \item Validation batch size: 16
        \item Test batch size: 16
    \end{itemize}

    \item \textbf{Hyperparameters:}
    \begin{itemize}
        \item Input shape: 256
        \item Epochs: 500
        \item Number of workers: 8
        \item Early stopping patience: 15 
        \item Sparsity patience: 30 (number of epochs after which training stops if sparsity does not increase)
    \end{itemize}

    \item \textbf{Task Decoders Optimizer Parameters:}
    \begin{itemize}
        \item Optimizer: adam 
        \item Optimizer parameters:
        \begin{itemize}
            \item Learning rate: 0.0001
            \item Betas: [0.9, 0.999]
            \item Weight decay: 0.01

        \end{itemize}
    \end{itemize}

    \item \textbf{Backbone Optimizer Parameters:}
    \begin{itemize}
        \item Backbone optimizer: adam 
        \item Backbone optimizer parameters:
        \begin{itemize}
            \item Learning rate: 0.00001
            \item Betas: [0.9, 0.999]
            \item Weight decay: 0.1
            \item Penalty: $l_1$-$l_2$ (alternatives: $l_1$-$l_2$, $l_1$)
            \item Lambda: 0.001 (used for fixed sparsity, for meta-sparsity it is learnable)
        \end{itemize}
    \end{itemize}
\end{itemize}

More details about the experiments and hyper-parameters can be found in the github repository. 

\subsection{Performance analysis}\label{app:performance}
For all the results and analysis in Section 5 of the main paper, the detailed results are tabulated below. 
For the NYU-v2 dataset, we have produced the tables for structured channel-wise $l_1-l_2$ sparsity (without re-growth), unstructured $l_1$ sparsity (without re-growth) in Tables~\ref{tab:main_results_NYU} and \ref{tab:NYU-l1}, respectively. 
While for the celebA dataset, the results for the structured channel-wise $l_1-l_2$ sparsity are in Table~\ref{tab:app_celebA}.

\input{tables/nyu_table_1}
\input{tables/nyu_l1}
\input{tables/celeb_results}

\subsection{Compression ration and Speed up}
Compression ratio (CR) is the degree of model compression by comparing the total number of parameters in the original model to the number of nonzero parameters in the sparse model. 
Speed-up (Sp), on the other hand, measures the improvement in computational efficiency by comparing the total FLOPs (floating-point operations) of the dense model to the sparse model.
Together, these metrics capture both the size reduction and the operational efficiency achieved through sparsity.
Table~\ref{tab:CR_SP} gives a comparison of compression ratio and speed-up metrics along with the tasks performances. 

\begin{table}[ht]
\centering
\caption{Comparison of performance, compression ratio (CR), and speed-up (Sp) for various tasks and task combinations with fixed $\lambda$ values and learned sparsity (meta-sparsity experiments). Metrics demonstrate the effectiveness of the proposed approach in balancing task performance and parameter efficiency. In the table, IoU stands for Intersection over Union, MAE is the Mean Absolute Error, and CS represents the Cosine Similarity. }
\label{tab:CR_SP}
\begin{tabular}{lccccccc}
\hline
Tasks  & $\lambda$ & CR & Sp & Seg. & Depth Est. & SN est. & Edge Det. \\
Combination&  &  &  & IoU($\uparrow$) & MAE($\downarrow$) & CS($\uparrow$) & MAE($\downarrow$) \\ \hline
T1,T2,T3 & $1 \times 10^{-3}$ & 10.29 & 5.51 & 0.3019 & 0.154 & 0.7439 & - \\
T1,T2,T3 & $1 \times 10^{-4}$ & 1.15 & 1.14 & 0.2922 & 0.1483 & 0.7322 & -\\
T1,T2,T3 & learned & 1.88 & 1.82 & 0.2937 & 0.1409 & 0.746 & - \\
T2,T3,T4 & $1 \times 10^{-3}$ & 5.47 & 3.85 & - & 0.1591 & 0.7172 & 0.1806 \\
T2,T3,T4 & $1 \times 10^{-4}$ & 2.38 & 2.1 & - & 0.1461 & 0.7396 & 0.1481 \\
T2,T3,T4 & learned & 1.81 & 1.68 & - & 0.1412 & 0.7402 & 0.1386 \\
T1,T2,T3,T4 & $1 \times 10^{-3}$ & 5.83 & 4.02 & 0.3032 & 0.1526 & 0.7417 & 0.1541 \\
T1,T2,T3,T4 & $1 \times 10^{-4}$ & 1.41 & 1.36 & 0.2957 & 0.1476 & 0.7353 & 0.1628 \\
T1,T2,T3,T4 & learned & 1.83 & 1.72 & 0.2923 & 0.1397 & 0.7414 & 0.1395 \\ \hline
\end{tabular}
\end{table}

%% file: tables/nyu_table_1.tex
\begin{table}[ht]
\centering
\caption{Results for single task and multi-task experiments on the NYU dataset. These include the no-sparsity, fixed-sparsity, and meta-sparsity experiments for \textbf{structured channel-wise $l_1-l_2$ type sparsity}. The results are shown in the form of mean $\pm$ standard deviation across five trials. The percentage of group sparsity and parameter sparsity exhibits a significant standard deviation across various trials during the training of single-task and multi-task models, utilizing a fixed lambda value for group sparsity. This variation arises because even small changes in parameter values during training can disproportionately impact sparsity by leading to the elimination of entire groups or channels, resulting in significant fluctuations in the overall sparsity levels. However, in meta-sparsity experiments, the variance in the percentage sparsity is minimal.}
\label{tab:main_results_NYU}
\resizebox{\textwidth}{!}{%
\begin{tabular}{cccccccc}
\hline
\multirow{3}{*}{\textbf{Experiments}} & \textbf{Sparsity} & \textbf{Segmentation} & \textbf{Depth est.} & \textbf{SN est.} & \textbf{Edge det.} & \textbf{Group} & \textbf{Parameter} \\
 & \textbf{parameter} & \textbf{$T_1$} & \textbf{$T_2$} & \textbf{$T_3$} & \textbf{$T_4$} & \textbf{sparsity} & \textbf{sparsity} \\
 & \textbf{$\lambda$} & IoU($\uparrow$) & MAE($\downarrow$) & CS($\uparrow$) & MAE($\downarrow$) & \textbf{(\%)} & \textbf{(\%)} \\ \hline
\multicolumn{1}{l}{} & \multicolumn{1}{l}{} & \multicolumn{1}{l}{} & \multicolumn{1}{l}{} & \multicolumn{1}{l}{} & \multicolumn{1}{l}{} & \multicolumn{1}{l}{} & \multicolumn{1}{l}{} \\
 & 0 & 0.2814 $\pm$ 0.0032 &- & - & - & - & -  \\
$T_1$ & 1.00E-04 & 0.2805 $\pm$ 0.0034 & - & - & - & 0.0000 $\pm$ 0.0000 & 0.0000 $\pm$ 0.0000 \\
 & 1.00E-03 & 0.2960 $\pm$ 0.0016 & - & - & - & 37.1098 $\pm$ 13.5190 & 66.3542 $\pm$ 11.0763 \\
 &  &  &  &  &  &  &  \\
 & 0 & - & 0.1577 $\pm$ 0.0018 &-  & - & - & - \\
$T_2$ & 1.00E-04 & - & 0.1612 $\pm$ 0.0027 & - & - & 2.5453 $\pm$ 0.9063 & 12.6610 $\pm$ 4.4319 \\
 & 1.00E-03 & - & 0.1555 $\pm$ 0.0011 & - & - & 29.2667 $\pm$ 22.2775 & 55.5215 $\pm$ 16.7659 \\
 &  &  &  &  &  &  &  \\
 & 0 & - & - & 0.7241 $\pm$ 0.0013 & - & - & -\\
$T_3$ & 1.00E-04 & - & - & 0.7265 $\pm$ 0.0032 & - & 7.4063 $\pm$ 0.8764 & 32.3315 $\pm$ 3.2081 \\
 & 1.00E-03 & - & - & 0.7331 $\pm$ 0.0034 & - & 20.9979 $\pm$ 12.2303 & 41.7222 $\pm$ 17.3127 \\
 &  &  &  &  &  &  &  \\
 & 0 & - & - & - & 0.5363 $\pm$ 0.0029 & - & - \\
$T_4$ & 1.00E-04 & - & - & - & 0.1962 $\pm$ 0.0235 & 39.0266 $\pm$ 22.5407 & 62.4191 $\pm$ 14.7042 \\
 & 1.00E-03 & - & - & - & 0.1807 $\pm$ 0.0090 & 52.2271 $\pm$ 25.1950 & 75.8073 $\pm$ 26.2364 \\
\multicolumn{6}{c}{} &  &  \\
 & 0 & 0.2914 $\pm$ 0.0041 & 0.1504 $\pm$ 0.0021 & 0.7325 $\pm$ 0.0031 &  & -& - \\
$T_1$, $T_2$, $T_3$ & 1.00E-04 & 0.2922 $\pm$ 0.0026 & 0.1483 $\pm$ 0.0022 & 0.7322 $\pm$ 0.0024 & - & 3.1158 $\pm$ 0.7348 & 13.1160 $\pm$ 3.6505 \\
 & 1.00E-03 & 0.3019 $\pm$ 0.0023 & 0.1540 $\pm$ 0.0027 & 0.7439 $\pm$ 0.0026 & - & 71.4829 $\pm$ 2.1812 & 90.2227 $\pm$ 0.4807 \\
 &  &  &  &  &  &  &  \\
 & 0 & - & 0.1423 $\pm$ 0.0062 & 0.7452 $\pm$ 0.0094 & 0.2152 $\pm$ 0.0098 & - & -\\
$T_2$, $T_3$, $T_4$ & 1.00E-04 & - & 0.1461 $\pm$ 0.0016 & 0.7396 $\pm$ 0.0054 & 0.1481 $\pm$ 0.0035 & 15.4064 $\pm$ 2.1178 & 57.6458 $\pm$ 4.0533 \\
 & 1.00E-03 & - & 0.1591 $\pm$ 0.0011 & 0.7172 $\pm$ 0.0019 & 0.1806 $\pm$ 0.0032 & 56.5270 $\pm$ 5.5767 & 81.7792 $\pm$ 1.4452 \\
 &  &  &  &  &  &  &  \\
 & 0 & 0.2910 $\pm$ 0.0024 & 0.1502 $\pm$ 0.0006 & 0.7313 $\pm$ 0.0022 & 0.2168 $\pm$ 0.0020 & - & - \\
$T_1$, $T_2$, $T_3$, $T_4$ & 1.00E-04 & 0.2957 $\pm$ 0.0046 & 0.1476 $\pm$ 0.0046 & 0.7353 $\pm$ 0.0020 & 0.1628 $\pm$ 0.0091 & 4.0918 $\pm$ 1.4320 & 17.1764 $\pm$ 6.1138 \\
 & 1.00E-03 & 0.3032 $\pm$ 0.0028 & 0.1526 $\pm$ 0.0028 & 0.7417 $\pm$ 0.0025 & 0.1541 $\pm$ 0.0051 & 72.4676 $\pm$ 2.1369 & 91.5959 $\pm$ 1.4343 \\\hline
 &  &  &  &  &  &  &  \\ 
$T_1$, $T_2$, $T_3$ & \multirow{3}{*}{\begin{tabular}[c]{@{}c@{}}meta \\ sparsity\end{tabular}} & 0.2937 $\pm$ 0.0091 & 0.1409 $\pm$ 0.0035 & 0.7460 $\pm$ 0.0073 & - & 12.7950 $\pm$ 0.0685 & 43.4391 $\pm$ 0.1925 \\
$T_2$, $T_3$, $T_4$ &  & - & 0.1412 $\pm$ 0.0040 & 0.7402 $\pm$ 0.0080 & 0.1386 $\pm$ 0.0019 & 12.9337 $\pm$ 0.0600 & 43.7577 $\pm$ 0.1886 \\
$T_1$, $T_2$, $T_3$, $T_4$ &  & 0.2923 $\pm$ 0.0101 & 0.1397 $\pm$ 0.0037 & 0.7414 $\pm$ 0.0055 & 0.1395 $\pm$ 0.0021 & 13.0806 $\pm$ 0.1508 & 44.1159 $\pm$ 0.3696 \\ \hline
\end{tabular}%
}
\end{table}

%% file: tables/nyu_l1.tex
\begin{table}[ht]
\centering
\caption{Results for single task and multi-task experiments on the NYU dataset. These include the no-sparsity, fixed-sparsity, and meta-sparsity experiments for\textbf{ unstructured $l_1$ type sparsity}. The results are shown in the form of mean $\pm$ standard deviation across five trials. The column for group sparsity (\%) is not added in this table, as $l_1$ sparsity does not induce such sparsity patterns wherein en entire group is zero. }
\label{tab:NYU-l1}
\resizebox{0.9\textwidth}{!}{%
\begin{tabular}{ccccccc}
\hline
\multirow{3}{*}{\textbf{Experiments}} & \textbf{Sparsity} & \textbf{Segmentation} & \textbf{Depth est.} & \textbf{SN est.} & \textbf{Edge det.} & \textbf{Parameter} \\
 & \textbf{parameter} & \textbf{$T_1$} & \textbf{$T_2$} & \textbf{$T_3$} & \textbf{$T_4$} & \textbf{sparsity} \\
 & $\lambda $ & IoU($\uparrow$) & MAE($\downarrow$) & CS($\uparrow$) & MAE($\downarrow$) & \textbf{(\%)} \\ \hline
 &  &  &  &  &  &  \\
 & 0 & 0.2814 $\pm$ 0.0032 & - & - & - & - \\
$T_1$ & 1.00E-04 & 0.2815 $\pm$ 0.0030 & - & - & - & 27.9499 $\pm$ 2.4901 \\
 & 1.00E-03 & 0.2847 $\pm$ 0.0062 & - & - & - & 62.1179 $\pm$ 2.5710 \\
 &  &  &  &  &  &  \\
 & 0 & - & 0.1577 $\pm$ 0.0018 & - & - & - \\
$T_2$ & 1.00E-04 & - & 0.1597 $\pm$ 0.0007 & - & - & 32.7109 $\pm$ 3.7194 \\
 & 1.00E-03 & - & 0.1602 $\pm$ 0.0023 & - & - & 60.9005 $\pm$ 5.0870 \\
 &  &  &  &  &  &  \\
 &  &  &  &  &  &  \\
 & 0 & - & - & 0.7241 $\pm$ 0.0013 & - & - \\
$T_3$ & 1.00E-04 & - & - & 0.7200 $\pm$ 0.0043 & - & 42.6871 $\pm$ 5.6393 \\
 & 1.00E-03 & - & - & 0.7058 $\pm$ 0.0070 & - & 46.7091 $\pm$ 9.5540 \\
 &  &  &  &  &  &  \\
 & 0 & - & - & - & 0.5363 $\pm$ 0.0029 & - \\
$T_4$ & 1.00E-04 & - & - & - & 0.1947 $\pm$ 0.0125 & 56.2187 $\pm$ 11.2639 \\
 & 1.00E-03 & - & - & - & 0.2123 $\pm$ 0.0233 & 64.0356 $\pm$ 8.9001 \\
 &  &  &  &  &  &  \\
 & 0 & 0.2914 $\pm$ 0.0041 & 0.1504 $\pm$ 0.0021 & 0.7325 $\pm$ 0.0031 & - & - \\
$T_1$, $T_2$, $T_3$ & 1.00E-04 & 0.2922 $\pm$ 0.0022 & 0.1511 $\pm$ 0.0014 & 0.7339 $\pm$ 0.0043 & - & 48.6015 $\pm$ 3.6821 \\
 & 1.00E-03 & 0.3025 $\pm$ 0.0101 & 0.1482 $\pm$ 0.0028 & 0.7504 $\pm$ 0.0055 & - & 77.7464 $\pm$ 3.6155 \\
 &  &  &  &  &  &  \\
 & 0 & - & 0.1423 $\pm$ 0.0062 & 0.7452 $\pm$ 0.0094 & 0.2152 $\pm$ 0.0098 & - \\
$T_2$, $T_3$, $T_4$ & 1.00E-04 & - & 0.1497 $\pm$ 0.0030 & 0.7375 $\pm$ 0.0030 & 0.2156 $\pm$ 0.0057 & 68.4097 $\pm$ 8.6707 \\
 & 1.00E-03 & - & 0.1574 $\pm$ 0.0031 & 0.7208 $\pm$ 0.0024 & 0.2374 $\pm$ 0.0019 & 79.3508 $\pm$ 1.7206 \\
 &  &  &  &  &  &  \\
 & 0 & 0.2910 $\pm$ 0.0024 & 0.1502 $\pm$ 0.0006 & 0.7313 $\pm$ 0.0022 & 0.2168 $\pm$ 0.0020 & - \\
$T_1$, $T_2$, $T_3$, $T_4$ & 1.00E-04 & 0.2948 $\pm$ 0.0007 & 0.1510 $\pm$ 0.0017 & 0.7364 $\pm$ 0.0015 & 0.2212 $\pm$ 0.0039 & 46.0315 $\pm$ 1.8424 \\
 & 1.00E-03 & 0.2992 $\pm$ 0.0041 & 0.1510 $\pm$ 0.0013 & 0.7458 $\pm$ 0.0016 & 0.2318 $\pm$ 0.0042 & 76.4574 $\pm$ 1.9152 \\
 &  &  &  &  &  &  \\ \hline
$T_1$, $T_2$, $T_3$ &  & 0.2675$\pm$0.0064 & 0.1440$\pm$0.0056 & 0.7310$\pm$0.0034 & - & 40.4412$\pm$0.1967 \\
$T_2$, $T_3$, $T_4$ & meta sparsity & - & 0.1407$\pm$0.0020 & 0.7313$\pm$0.0072 & 0.1439$\pm$0.0002 & 40.5969$\pm$0.1064 \\
$T_1$, $T_2$, $T_3$, $T_4$ &  & 0.2692$\pm$0.0017 & 0.1385$\pm$0.0010 & 0.7307$\pm$0.0033 & 0.1445$\pm$0.0011 & 30.2017$\pm$0.0846 \\ \hline
\end{tabular}%
}
\end{table}

%% file: tables/celeb_results.tex

\begin{landscape}
\begin{table}[ht]
\centering

\caption{Results for single task and multi-task experiments on the CelebA dataset. These include the no-sparsity, fixed-sparsity, and meta-sparsity experiments. The results are shown in the form of mean $\pm$ standard deviation across five trials. here $T_1-T_7$ represents all tasks from $T_1$ to $T_7$, similarly for $T_2-T_7$ it considers tasks ranging from $T_2$ to $T_7$. Also, $\lambda = 0$ represents experiments without sparsity.}
\label{tab:app_celebA}
\resizebox{1.2\textwidth}{!}{%
\begin{tabular}{cccccccccc}
\hline
\textbf{Experiments} & \textbf{Sparsity} & \textbf{Segmentation} & \textbf{male/female} & \textbf{smile/no smile} & \textbf{biglips/no biglips} & \textbf{highcheekbones/no} & \textbf{wearing lipstick/} & \textbf{bushy eyebrows/ no} & \textbf{Parameter} \\
\textbf{} & \textbf{parameter} & \textbf{} & \textbf{} & \textbf{} & \textbf{} & \textbf{highcheekbones} & \textbf{not wearing lipstick} & \textbf{bushy eyebrows} & \textbf{sparsity} \\
 & $\lambda$ & $T_1$ & $T_2$ & $T_3$ & $T_4$ & $T_5$ & $T_6$ & $T_7$ &  \\
 &  & IoU($\uparrow$) & Accuracy($\uparrow$) & Accuracy($\uparrow$) & Accuracy($\uparrow$) & Accuracy($\uparrow$) & Accuracy($\uparrow$) & Accuracy($\uparrow$) & (\%) \\ \hline
 &  &  &  &  &  &  &  &  &  \\
 & 0 & 0.9195 $\pm$ 0.0001 & - & - & - & - & - & - & - \\
$T_1$ & 1.00E-04 & 0.8996 $\pm$ 0.0026 & - & - & - & - & - & - & 81.2797 $\pm$ 0.6728 \\
 & 1.00E-03 & 0.8685 $\pm$ 0.0033 & - & - & - & - & - & - & 82.2677 $\pm$ 0.7133 \\
 &  &  &  &  &  &  &  &  &  \\
 & 0 & - & 0.9648 $\pm$ 0.0044 & - & - & - & - & - &  -\\
$T_2$ & 1.00E-04 & - & 0.9255 $\pm$ 0.0088 & - & - & - & - & - & 74.3208 $\pm$ 4.8320 \\
 & 1.00E-03 & - & 0.7773 $\pm$ 0.0971 & - & - & - & - & - & 81.5482 $\pm$ 1.1930 \\
 &  &  &  &  &  &  &  &  &  \\
 & 0 & - & - & 0.8987 $\pm$ 0.0016 & - & - & - & - & - \\
$T_3$ & 1.00E-04 & - & - & 0.8891 $\pm$ 0.0034 & - & - & - & - & 48.3989 $\pm$ 0.8911 \\
 & 1.00E-03 & - & - & 0.8569 $\pm$ 0.0127 & - & - & - & - & 78.7108 $\pm$ 2.3941 \\
 &  &  &  &  &  &  &  &  &  \\
 & 0 & - & - & - & 0.6358 $\pm$ 0.0061 &  &  &  & - \\
$T_4$ & 1.00E-04 & - & - & - & 0.6270 $\pm$ 0.0067 & - & - & - & 67.6497 $\pm$ 5.9694 \\
 & 1.00E-03 & - & - & - & 0.6259 $\pm$ 0.0069 & - & - & - & 71.6123 $\pm$ 0.4355 \\
 &  &  &  &  &  &  &  &  &  \\
 & 0 & - & - & - & - & 0.8273 $\pm$ 0.0101 & - & - & - \\
$T_5$ & 1.00E-04 & - & - & - & - & 0.6984 $\pm$ 0.0764 & - & - & 80.6301 $\pm$ 0.0829 \\
 & 1.00E-03 & - & - & - & - & 0.5433 $\pm$ 0.0167 & - & - & 81.8055 $\pm$ 1.8168 \\
 &  &  &  &  &  &  &  &  &  \\
 & 0 & - & - & - & - & - & 0.9106 $\pm$ 0.0049 & - & - \\
$T_6$ & 1.00E-04& - & - & - & - & - & 0.8522 $\pm$ 0.0129 & - & 63.3811 $\pm$ 6.2434 \\
 & 1.00E-03 & - & - & - & - & - & 0.7502 $\pm$ 0.0486 & - & 83.2237 $\pm$ 3.1300 \\
 &  &  &  &  &  &  &  &  &  \\
 & 0 & - & - & - & - & - & - & 0.6270 $\pm$ 0.0100 & - \\
$T_7$ & 1.00E-04 & - & - & - & - & - & - & 0.6351 $\pm$ 0.0104 & 65.1942 $\pm$ 5.8890 \\
 & 1.00E-03 & - & - & - & - & - & - & 0.6212 $\pm$ 0.0023 & 74.2825 $\pm$ 6.2813 \\
 &  &  &  &  &  &  &  &  &  \\
 & 0 & 0.8941 $\pm$ 0.0027 & 0.9554 $\pm$ 0.0055 & 0.8909 $\pm$ 0.0047 & 0.6340 $\pm$ 0.0003 & 0.8454 $\pm$ 0.0010 & 0.9122 $\pm$ 0.0031 & 0.7867 $\pm$ 0.0016 & - \\
$T_1$-$T_7$ & 1.00E-05 & 0.8437 $\pm$ 0.0022 & 0.8882 $\pm$ 0.0041 & 0.5253 $\pm$ 0.0011 & 0.6297 $\pm$ 0.0045 & 0.5916 $\pm$ 0.0247 & 0.8412 $\pm$ 0.0172 & 0.6241 $\pm$ 0.0008 & 0.0000 $\pm$ 0.0000 \\
 & 1.00E-04 & 0.8066 $\pm$ 0.0196 & 0.6268 $\pm$ 0.0011 & 0.5215 $\pm$ 0.0020 & 0.6328 $\pm$ 0.0006 & 0.5235 $\pm$ 0.0034 & 0.7274 $\pm$ 0.0152 & 0.6225 $\pm$ 0.0021 & 85.6721 $\pm$ 1.7966 \\
 &  &  &  &  &  &  &  &  &  \\
 & 0 & - & 0.9496 $\pm$ 0.0048 & 0.8790 $\pm$ 0.0128 & 0.6337 $\pm$ 0.0000 & 0.8327 $\pm$ 0.0217 & 0.9034 $\pm$ 0.0084 & 0.7814 $\pm$ 0.0154 & - \\
$T_2$-$T_7$ & 1.00E-05 & - & 0.9350 $\pm$ 0.0054 & 0.8669 $\pm$ 0.0121 & 0.6331 $\pm$ 0.0011 & 0.8177 $\pm$ 0.0047 & 0.8947 $\pm$ 0.0170 & 0.7247 $\pm$ 0.0229 & 0.4095 $\pm$ 0.0884 \\
 & 1.00E-04 & - & 0.6248 $\pm$ 0.0014 & 0.5240 $\pm$ 0.0037 & 0.6284 $\pm$ 0.0019 & 0.5166 $\pm$ 0.0014 & 0.5529 $\pm$ 0.0595 & 0.6216 $\pm$ 0.0058 & 95.5407 $\pm$ 1.3195 \\
 &  &  &  &  &  &  &  &  &  \\
 & 0 & 0.8995 $\pm$ 0.0014 & 0.9513 $\pm$ 0.0010 & 0.8901 $\pm$ 0.0017 & - & - & - & 0.7890 $\pm$ 0.0087 & - \\
$T_1$,$T_2$,$T_3$,$T_7$ & 1.00E-05 & 0.8663 $\pm$ 0.0104 & 0.6270 $\pm$ 0.0010 & 0.5260 $\pm$ 0.0021 & - & - & - & 0.6221 $\pm$ 0.0027 & 0.0811 $\pm$ 0.0659 \\
 & 1.00E-04 & 0.8223 $\pm$ 0.0050 & 0.6261 $\pm$ 0.0043 & 0.5249 $\pm$ 0.0019 & - & - & - & 0.6246 $\pm$ 0.0018 & 85.9002 $\pm$ 2.0208 \\
 &  &  &  &  &  &  &  &  &  \\
$T_1$-$T_7$ &  & 0.9007$\pm$0.0012 & 0.9669$\pm$0.0014 & 0.9014$\pm$0.0043 & 0.6330$\pm$0.1680 & 0.8235$\pm$0.0104 & 0.9068$\pm$0.0165 & 0.6885$\pm$0.0936 & 43.6893$\pm$0.3593 \\
$T_2$-$T_7$ & meta-sparsity & - & 0.9004$\pm$0.0203 & 0.7618$\pm$0.1602 & 0.6330$\pm$0.0720 & 0.7562$\pm$0.1109 & 0.8703$\pm$0.0154 & 0.6223$\pm$0.0000 & 30.2145$\pm$0.0016 \\
$T_1$,$T_2$,$T_3$,$T_7$ &  & 0.9034$\pm$0.0013 & 0.7386$\pm$0.0188 & 0.5261$\pm$0.0000 & - & - & - & 0.6223$\pm$0.0000 & 30.0187$\pm$0.0011 \\ \hline
\end{tabular}%
}
\end{table}
\end{landscape}

%% file: main.bbl
\begin{thebibliography}{100}
\providecommand{\natexlab}[1]{#1}
\providecommand{\url}[1]{\texttt{#1}}
\expandafter\ifx\csname urlstyle\endcsname\relax
  \providecommand{\doi}[1]{doi: #1}\else
  \providecommand{\doi}{doi: \begingroup \urlstyle{rm}\Url}\fi

\bibitem[Andrychowicz et~al.(2016)Andrychowicz, Denil, Gomez, Hoffman, Pfau, Schaul, Shillingford, and De~Freitas]{andrychowicz2016learning}
Marcin Andrychowicz, Misha Denil, Sergio Gomez, Matthew~W Hoffman, David Pfau, Tom Schaul, Brendan Shillingford, and Nando De~Freitas.
\newblock Learning to learn by gradient descent by gradient descent.
\newblock \emph{Advances in neural information processing systems}, 29, 2016.

\bibitem[Argyriou et~al.(2006)Argyriou, Evgeniou, and Pontil]{argyriou2006multi}
Andreas Argyriou, Theodoros Evgeniou, and Massimiliano Pontil.
\newblock Multi-task feature learning.
\newblock \emph{Advances in neural information processing systems}, 19, 2006.

\bibitem[Bach et~al.(2012)Bach, Jenatton, Mairal, Obozinski, et~al.]{bach2012optimization}
Francis Bach, Rodolphe Jenatton, Julien Mairal, Guillaume Obozinski, et~al.
\newblock Optimization with sparsity-inducing penalties.
\newblock \emph{Foundations and Trends{\textregistered} in Machine Learning}, 4\penalty0 (1):\penalty0 1--106, 2012.

\bibitem[Baik et~al.(2020)Baik, Choi, Choi, Kim, and Lee]{baik2020meta}
Sungyong Baik, Myungsub Choi, Janghoon Choi, Heewon Kim, and Kyoung~Mu Lee.
\newblock Meta-learning with adaptive hyperparameters.
\newblock \emph{Advances in neural information processing systems}, 33:\penalty0 20755--20765, 2020.

\bibitem[Baxter(1998)]{baxter1998theoretical}
Jonathan Baxter.
\newblock Theoretical models of learning to learn.
\newblock In \emph{Learning to learn}, pp.\  71--94. Springer, 1998.

\bibitem[Bechtle et~al.(2021)Bechtle, Molchanov, Chebotar, Grefenstette, Righetti, Sukhatme, and Meier]{bechtle2021meta}
Sarah Bechtle, Artem Molchanov, Yevgen Chebotar, Edward Grefenstette, Ludovic Righetti, Gaurav Sukhatme, and Franziska Meier.
\newblock Meta learning via learned loss.
\newblock In \emph{2020 25th International Conference on Pattern Recognition (ICPR)}, pp.\  4161--4168. IEEE, 2021.

\bibitem[Bengio et~al.(1991)Bengio, Bengio, and Cloutier]{155621}
Y.~Bengio, S.~Bengio, and J.~Cloutier.
\newblock Learning a synaptic learning rule.
\newblock In \emph{IJCNN-91-Seattle International Joint Conference on Neural Networks}, volume~ii, pp.\  969 vol.2--, 1991.
\newblock \doi{10.1109/IJCNN.1991.155621}.

\bibitem[Blalock et~al.(2020)Blalock, Gonzalez~Ortiz, Frankle, and Guttag]{blalock2020state}
Davis Blalock, Jose~Javier Gonzalez~Ortiz, Jonathan Frankle, and John Guttag.
\newblock What is the state of neural network pruning?
\newblock \emph{Proceedings of machine learning and systems}, 2:\penalty0 129--146, 2020.

\bibitem[Bohdal et~al.(2021)Bohdal, Yang, and Hospedales]{NEURIPS2021_bac49b87}
Ondrej Bohdal, Yongxin Yang, and Timothy Hospedales.
\newblock Evograd: Efficient gradient-based meta-learning and hyperparameter optimization.
\newblock In M.~Ranzato, A.~Beygelzimer, Y.~Dauphin, P.S. Liang, and J.~Wortman Vaughan (eds.), \emph{Advances in Neural Information Processing Systems}, volume~34, pp.\  22234--22246. Curran Associates, Inc., 2021.

\bibitem[Boyd et~al.(2011)Boyd, Parikh, Chu, Peleato, Eckstein, et~al.]{boyd2011distributed}
Stephen Boyd, Neal Parikh, Eric Chu, Borja Peleato, Jonathan Eckstein, et~al.
\newblock Distributed optimization and statistical learning via the alternating direction method of multipliers.
\newblock \emph{Foundations and Trends{\textregistered} in Machine learning}, 3\penalty0 (1):\penalty0 1--122, 2011.

\bibitem[Carreira-Perpinan \& Idelbayev(2018)Carreira-Perpinan and Idelbayev]{8578988}
Miguel~A. Carreira-Perpinan and Yerlan Idelbayev.
\newblock "learning-compression" algorithms for neural net pruning.
\newblock In \emph{2018 IEEE/CVF Conference on Computer Vision and Pattern Recognition}, pp.\  8532--8541, 2018.
\newblock \doi{10.1109/CVPR.2018.00890}.

\bibitem[Caruana(1997)]{caruana1997multitask}
Rich Caruana.
\newblock Multitask learning.
\newblock \emph{Machine learning}, 28:\penalty0 41--75, 1997.

\bibitem[Chen et~al.(2009)Chen, Tang, Liu, and Ye]{10.1145/1553374.1553392}
Jianhui Chen, Lei Tang, Jun Liu, and Jieping Ye.
\newblock A convex formulation for learning shared structures from multiple tasks.
\newblock In \emph{Proceedings of the 26th Annual International Conference on Machine Learning}, ICML '09, pp.\  137–144, New York, NY, USA, 2009. Association for Computing Machinery.
\newblock ISBN 9781605585161.
\newblock \doi{10.1145/1553374.1553392}.

\bibitem[Chen et~al.(2017)Chen, Papandreou, Schroff, and Adam]{chen2017rethinking}
Liang-Chieh Chen, George Papandreou, Florian Schroff, and Hartwig Adam.
\newblock Rethinking atrous convolution for semantic image segmentation, 2017.
\newblock arXiv:1706.05587.

\bibitem[Chen et~al.(2023)Chen, Liu, Li, Jiang, Ding, and Zhou]{meta_LR}
Yixiong Chen, Li~Liu, Jingxian Li, Hua Jiang, Chris Ding, and Zongwei Zhou.
\newblock Metalr: Meta-tuning of learning rates for transfer learning in medical imaging.
\newblock In Hayit Greenspan, Anant Madabhushi, Parvin Mousavi, Septimiu Salcudean, James Duncan, Tanveer Syeda-Mahmood, and Russell Taylor (eds.), \emph{Medical Image Computing and Computer Assisted Intervention -- MICCAI 2023}, pp.\  706--716, Cham, 2023. Springer Nature Switzerland.
\newblock ISBN 978-3-031-43907-0.

\bibitem[Combettes \& Wajs(2005)Combettes and Wajs]{combettes2005signal}
Patrick~L Combettes and Val{\'e}rie~R Wajs.
\newblock Signal recovery by proximal forward-backward splitting.
\newblock \emph{Multiscale modeling \& simulation}, 4\penalty0 (4):\penalty0 1168--1200, 2005.

\bibitem[Crawshaw(2020)]{crawshaw2020multitask}
Michael Crawshaw.
\newblock Multi-task learning with deep neural networks: A survey.
\newblock \emph{arXiv preprint arXiv:2009.09796}, 2020.

\bibitem[Deleu \& Bengio(2021)Deleu and Bengio]{deleu2021structured}
Tristan Deleu and Yoshua Bengio.
\newblock Structured sparsity inducing adaptive optimizers for deep learning.
\newblock \emph{arXiv preprint arXiv:2102.03869}, 2021.

\bibitem[Deng et~al.(2020)Deng, Li, Han, Shi, and Xie]{9043731}
Lei Deng, Guoqi Li, Song Han, Luping Shi, and Yuan Xie.
\newblock Model compression and hardware acceleration for neural networks: A comprehensive survey.
\newblock \emph{Proceedings of the IEEE}, 108\penalty0 (4):\penalty0 485--532, 2020.
\newblock \doi{10.1109/JPROC.2020.2976475}.

\bibitem[Desai \& Shrivastava(2023)Desai and Shrivastava]{desai2023defense}
Aditya Desai and Anshumali Shrivastava.
\newblock In defense of parameter sharing for model-compression.
\newblock \emph{arXiv preprint arXiv:2310.11611}, 2023.

\bibitem[Dettmers \& Zettlemoyer(2019)Dettmers and Zettlemoyer]{dettmers2019sparse}
Tim Dettmers and Luke Zettlemoyer.
\newblock Sparse networks from scratch: Faster training without losing performance.
\newblock \emph{arXiv preprint arXiv:1907.04840}, 2019.

\bibitem[Ding et~al.(2022)Ding, Wu, Huang, Tang, Yang, Wei, Zhuang, and Tian]{Ding_2022_CVPR}
Yadong Ding, Yu~Wu, Chengyue Huang, Siliang Tang, Yi~Yang, Longhui Wei, Yueting Zhuang, and Qi~Tian.
\newblock Learning to learn by jointly optimizing neural architecture and weights.
\newblock In \emph{Proceedings of the IEEE/CVF Conference on Computer Vision and Pattern Recognition (CVPR)}, pp.\  129--138, June 2022.

\bibitem[Elsken et~al.(2019)Elsken, Metzen, and Hutter]{elsken2019neural}
Thomas Elsken, Jan~Hendrik Metzen, and Frank Hutter.
\newblock Neural architecture search: A survey.
\newblock \emph{The Journal of Machine Learning Research}, 20\penalty0 (1):\penalty0 1997--2017, 2019.

\bibitem[Elsken et~al.(2020)Elsken, Staffler, Metzen, and Hutter]{elsken2020meta}
Thomas Elsken, Benedikt Staffler, Jan~Hendrik Metzen, and Frank Hutter.
\newblock Meta-learning of neural architectures for few-shot learning.
\newblock In \emph{Proceedings of the IEEE/CVF conference on computer vision and pattern recognition}, pp.\  12365--12375, 2020.

\bibitem[Evci et~al.(2020)Evci, Gale, Menick, Castro, and Elsen]{evci2020rigging}
Utku Evci, Trevor Gale, Jacob Menick, Pablo~Samuel Castro, and Erich Elsen.
\newblock Rigging the lottery: Making all tickets winners.
\newblock In \emph{International Conference on Machine Learning}, pp.\  2943--2952. PMLR, 2020.

\bibitem[Finn et~al.(2017)Finn, Abbeel, and Levine]{pmlr-v70-finn17a}
Chelsea Finn, Pieter Abbeel, and Sergey Levine.
\newblock Model-agnostic meta-learning for fast adaptation of deep networks.
\newblock In Doina Precup and Yee~Whye Teh (eds.), \emph{Proceedings of the 34th International Conference on Machine Learning}, volume~70 of \emph{Proceedings of Machine Learning Research}, pp.\  1126--1135. PMLR, 06--11 Aug 2017.

\bibitem[Franceschi et~al.(2018)Franceschi, Frasconi, Salzo, Grazzi, and Pontil]{franceschi2018bilevel}
Luca Franceschi, Paolo Frasconi, Saverio Salzo, Riccardo Grazzi, and Massimiliano Pontil.
\newblock Bilevel programming for hyperparameter optimization and meta-learning.
\newblock In \emph{International conference on machine learning}, pp.\  1568--1577. PMLR, 2018.

\bibitem[Frankle \& Carbin(2019)Frankle and Carbin]{frankle2018the}
Jonathan Frankle and Michael Carbin.
\newblock The lottery ticket hypothesis: Finding sparse, trainable neural networks.
\newblock In \emph{International Conference on Learning Representations}, 2019.

\bibitem[Gale et~al.(2019)Gale, Elsen, and Hooker]{gale2019state}
Trevor Gale, Erich Elsen, and Sara Hooker.
\newblock The state of sparsity in deep neural networks.
\newblock \emph{arXiv preprint arXiv:1902.09574}, 2019.

\bibitem[Gao et~al.(2021)Gao, Gouk, and Hospedales]{gao2021searching}
Boyan Gao, Henry Gouk, and Timothy~M Hospedales.
\newblock Searching for robustness: Loss learning for noisy classification tasks.
\newblock In \emph{Proceedings of the IEEE/CVF International Conference on Computer Vision}, pp.\  6670--6679, 2021.

\bibitem[Gao et~al.(2022{\natexlab{a}})Gao, Gouk, Lee, and Hospedales]{gao2022meta}
Boyan Gao, Henry Gouk, Hae~Beom Lee, and Timothy~M Hospedales.
\newblock Meta mirror descent: Optimiser learning for fast convergence.
\newblock \emph{arXiv preprint arXiv:2203.02711}, 2022{\natexlab{a}}.

\bibitem[Gao et~al.(2022{\natexlab{b}})Gao, Gouk, Yang, and Hospedales]{gao2022loss}
Boyan Gao, Henry Gouk, Yongxin Yang, and Timothy Hospedales.
\newblock Loss function learning for domain generalization by implicit gradient.
\newblock In \emph{International Conference on Machine Learning}, pp.\  7002--7016. PMLR, 2022{\natexlab{b}}.

\bibitem[Glorot \& Bengio(2010)Glorot and Bengio]{pmlr-v9-glorot10a}
Xavier Glorot and Yoshua Bengio.
\newblock Understanding the difficulty of training deep feedforward neural networks.
\newblock In Yee~Whye Teh and Mike Titterington (eds.), \emph{Proceedings of the Thirteenth International Conference on Artificial Intelligence and Statistics}, volume~9 of \emph{Proceedings of Machine Learning Research}, pp.\  249--256, Chia Laguna Resort, Sardinia, Italy, 13--15 May 2010. PMLR.

\bibitem[Gon{\c{c}}alves et~al.(2016)Gon{\c{c}}alves, Zuben, and Banerjee]{JMLR:v17:15-215}
Andr{{\'e}}~R. Gon{\c{c}}alves, Fernando J.~Von Zuben, and Arindam Banerjee.
\newblock Multi-task sparse structure learning with gaussian copula models.
\newblock \emph{Journal of Machine Learning Research}, 17\penalty0 (33):\penalty0 1--30, 2016.

\bibitem[Goodfellow et~al.(2016)Goodfellow, Bengio, and Courville]{Goodfellow-et-al-2016}
Ian Goodfellow, Yoshua Bengio, and Aaron Courville.
\newblock \emph{Deep Learning}.
\newblock MIT Press, 2016.
\newblock URL \url{http://www.deeplearningbook.org}.
\newblock Chapter 5: Machine Learning Basics.

\bibitem[Han et~al.(2015)Han, Pool, Tran, and Dally]{han2015learning}
Song Han, Jeff Pool, John Tran, and William Dally.
\newblock Learning both weights and connections for efficient neural network.
\newblock \emph{Advances in neural information processing systems}, 28, 2015.

\bibitem[Harrison et~al.(2022)Harrison, Metz, and Sohl-Dickstein]{harrison2022closer}
James Harrison, Luke Metz, and Jascha Sohl-Dickstein.
\newblock A closer look at learned optimization: Stability, robustness, and inductive biases.
\newblock \emph{Advances in Neural Information Processing Systems}, 35:\penalty0 3758--3773, 2022.

\bibitem[Hastie et~al.(2015)Hastie, Tibshirani, and Wainwright]{hastie2015statistical}
Trevor Hastie, Robert Tibshirani, and Martin Wainwright.
\newblock \emph{Statistical learning with sparsity: the lasso and generalizations}.
\newblock CRC press, 2015.

\bibitem[Hinton et~al.(2015)Hinton, Vinyals, and Dean]{hinton2015distilling}
Geoffrey Hinton, Oriol Vinyals, and Jeff Dean.
\newblock Distilling the knowledge in a neural network.
\newblock \emph{arXiv preprint arXiv:1503.02531}, 2015.

\bibitem[Hoefler et~al.(2021)Hoefler, Alistarh, Ben-Nun, Dryden, and Peste]{review_sparsity}
Torsten Hoefler, Dan Alistarh, Tal Ben-Nun, Nikoli Dryden, and Alexandra Peste.
\newblock Sparsity in deep learning: Pruning and growth for efficient inference and training in neural networks.
\newblock \emph{J. Mach. Learn. Res.}, 22\penalty0 (1), jan 2021.
\newblock ISSN 1532-4435.

\bibitem[Hospedales et~al.(2022)Hospedales, Antoniou, Micaelli, and Storkey]{meta_timothy}
T.~Hospedales, A.~Antoniou, P.~Micaelli, and A.~Storkey.
\newblock Meta-learning in neural networks: A survey.
\newblock \emph{IEEE Transactions on Pattern Analysis \&amp; Machine Intelligence}, 44\penalty0 (09):\penalty0 5149--5169, sep 2022.
\newblock ISSN 1939-3539.
\newblock \doi{10.1109/TPAMI.2021.3079209}.

\bibitem[Hu et~al.(2019)Hu, Ozay, Zhang, and Okatani]{depth_loss}
Junjie Hu, Mete Ozay, Yan Zhang, and Takayuki Okatani.
\newblock Revisiting single image depth estimation: Toward higher resolution maps with accurate object boundaries.
\newblock In \emph{2019 IEEE Winter Conference on Applications of Computer Vision (WACV)}, 2019.

\bibitem[Hubens(2020)]{hubens2020pruning}
Nathan Hubens.
\newblock Neural network pruning.
\newblock \emph{nathanhubens.github.io}, 2020.

\bibitem[Huisman et~al.(2021)Huisman, Van~Rijn, and Plaat]{huisman2021survey}
Mike Huisman, Jan~N Van~Rijn, and Aske Plaat.
\newblock A survey of deep meta-learning.
\newblock \emph{Artificial Intelligence Review}, 54\penalty0 (6):\penalty0 4483--4541, 2021.

\bibitem[Janowsky(1989)]{PhysRevA.39.6600}
Steven~A. Janowsky.
\newblock Pruning versus clipping in neural networks.
\newblock \emph{Phys. Rev. A}, 39:\penalty0 6600--6603, Jun 1989.
\newblock \doi{10.1103/PhysRevA.39.6600}.
\newblock URL \url{https://link.aps.org/doi/10.1103/PhysRevA.39.6600}.

\bibitem[Kendall et~al.(2018)Kendall, Gal, and Cipolla]{kendall2018multi}
Alex Kendall, Yarin Gal, and Roberto Cipolla.
\newblock Multi-task learning using uncertainty to weigh losses for scene geometry and semantics.
\newblock In \emph{Proceedings of the IEEE conference on computer vision and pattern recognition}, pp.\  7482--7491, 2018.

\bibitem[Kshirsagar et~al.(2017)Kshirsagar, Yang, and Lozano]{kshirsagar2017learning}
Meghana Kshirsagar, Eunho Yang, and Aur{\'e}lie~C Lozano.
\newblock Learning task structure via sparsity grouped multitask learning.
\newblock \emph{arXiv preprint arXiv:1705.04886}, 2017.

\bibitem[Kusupati et~al.(2020)Kusupati, Ramanujan, Somani, Wortsman, Jain, Kakade, and Farhadi]{pmlr-v119-kusupati20a}
Aditya Kusupati, Vivek Ramanujan, Raghav Somani, Mitchell Wortsman, Prateek Jain, Sham Kakade, and Ali Farhadi.
\newblock Soft threshold weight reparameterization for learnable sparsity.
\newblock In Hal~Daumé III and Aarti Singh (eds.), \emph{Proceedings of the 37th International Conference on Machine Learning}, volume 119 of \emph{Proceedings of Machine Learning Research}, pp.\  5544--5555. PMLR, 13--18 Jul 2020.

\bibitem[Lee et~al.(2020)Lee, Liu, Wu, and Luo]{CelebAMask-HQ}
Cheng-Han Lee, Ziwei Liu, Lingyun Wu, and Ping Luo.
\newblock Maskgan: Towards diverse and interactive facial image manipulation.
\newblock In \emph{IEEE Conference on Computer Vision and Pattern Recognition (CVPR)}, 2020.

\bibitem[Lee et~al.(2021)Lee, Park, Mo, Ahn, and Shin]{lee2021layeradaptive}
Jaeho Lee, Sejun Park, Sangwoo Mo, Sungsoo Ahn, and Jinwoo Shin.
\newblock Layer-adaptive sparsity for the magnitude-based pruning.
\newblock In \emph{International Conference on Learning Representations}, 2021.

\bibitem[Li \& Malik(2017)Li and Malik]{li2017learning}
Ke~Li and Jitendra Malik.
\newblock Learning to optimize neural nets.
\newblock \emph{arXiv preprint arXiv:1703.00441}, 2017.

\bibitem[Li et~al.(2017)Li, Zhou, Chen, and Li]{li2017meta}
Zhenguo Li, Fengwei Zhou, Fei Chen, and Hang Li.
\newblock Meta-sgd: Learning to learn quickly for few-shot learning.
\newblock \emph{arXiv preprint arXiv:1707.09835}, 2017.

\bibitem[Lian et~al.(2019)Lian, Zheng, Xu, Lu, Lin, Zhao, Huang, and Gao]{lian2019towards}
Dongze Lian, Yin Zheng, Yintao Xu, Yanxiong Lu, Leyu Lin, Peilin Zhao, Junzhou Huang, and Shenghua Gao.
\newblock Towards fast adaptation of neural architectures with meta learning.
\newblock In \emph{International Conference on Learning Representations}, 2019.

\bibitem[Liebel \& K{\"o}rner(2018)Liebel and K{\"o}rner]{liebel2018auxiliary}
Lukas Liebel and Marco K{\"o}rner.
\newblock Auxiliary tasks in multi-task learning.
\newblock \emph{arXiv preprint arXiv:1805.06334}, 2018.

\bibitem[Liu \& Wang(2023)Liu and Wang]{liu2023lessons}
Shiwei Liu and Zhangyang Wang.
\newblock Ten lessons we have learned in the new "sparseland": A short handbook for sparse neural network researchers, 2023.

\bibitem[Liu et~al.(2017)Liu, Li, Shen, Huang, Yan, and Zhang]{liu2017learning}
Zhuang Liu, Jianguo Li, Zhiqiang Shen, Gao Huang, Shoumeng Yan, and Changshui Zhang.
\newblock Learning efficient convolutional networks through network slimming.
\newblock In \emph{Proceedings of the IEEE international conference on computer vision}, pp.\  2736--2744, 2017.

\bibitem[Liu et~al.(2018)Liu, Sun, Zhou, Huang, and Darrell]{liu2018rethinking}
Zhuang Liu, Mingjie Sun, Tinghui Zhou, Gao Huang, and Trevor Darrell.
\newblock Rethinking the value of network pruning.
\newblock \emph{arXiv preprint arXiv:1810.05270}, 2018.

\bibitem[Lv et~al.(2017)Lv, Jiang, and Li]{lv2017learning}
Kaifeng Lv, Shunhua Jiang, and Jian Li.
\newblock Learning gradient descent: Better generalization and longer horizons.
\newblock In \emph{International Conference on Machine Learning}, pp.\  2247--2255. PMLR, 2017.

\bibitem[Meng et~al.(2020)Meng, Cheng, Li, Luo, Guo, Lu, and Sun]{meng2020pruning}
Fanxu Meng, Hao Cheng, Ke~Li, Huixiang Luo, Xiaowei Guo, Guangming Lu, and Xing Sun.
\newblock Pruning filter in filter.
\newblock \emph{Advances in Neural Information Processing Systems}, 33:\penalty0 17629--17640, 2020.

\bibitem[Metz et~al.(2020)Metz, Maheswaranathan, Freeman, Poole, and Sohl-Dickstein]{metz2020tasks}
Luke Metz, Niru Maheswaranathan, C~Daniel Freeman, Ben Poole, and Jascha Sohl-Dickstein.
\newblock Tasks, stability, architecture, and compute: Training more effective learned optimizers, and using them to train themselves.
\newblock \emph{arXiv preprint arXiv:2009.11243}, 2020.

\bibitem[Metz et~al.(2022)Metz, Harrison, Freeman, Merchant, Beyer, Bradbury, Agrawal, Poole, Mordatch, Roberts, et~al.]{metz2022velo}
Luke Metz, James Harrison, C~Daniel Freeman, Amil Merchant, Lucas Beyer, James Bradbury, Naman Agrawal, Ben Poole, Igor Mordatch, Adam Roberts, et~al.
\newblock Velo: Training versatile learned optimizers by scaling up.
\newblock \emph{arXiv preprint arXiv:2211.09760}, 2022.

\bibitem[Nathan~Silberman \& Fergus(2012)Nathan~Silberman and Fergus]{Silberman_ECCV12}
Pushmeet~Kohli Nathan~Silberman, Derek~Hoiem and Rob Fergus.
\newblock Indoor segmentation and support inference from rgbd images.
\newblock In \emph{ECCV}, 2012.

\bibitem[Nichol \& Schulman(2018)Nichol and Schulman]{nichol2018reptile}
Alex Nichol and John Schulman.
\newblock Reptile: a scalable metalearning algorithm.
\newblock \emph{arXiv preprint arXiv:1803.02999}, 2\penalty0 (3):\penalty0 4, 2018.

\bibitem[Nichol et~al.(2018)Nichol, Achiam, and Schulman]{nichol2018first}
Alex Nichol, Joshua Achiam, and John Schulman.
\newblock On first-order meta-learning algorithms.
\newblock \emph{arXiv preprint arXiv:1803.02999}, 2018.

\bibitem[Obozinski et~al.(2010)Obozinski, Taskar, and Jordan]{obozinski2010joint}
Guillaume Obozinski, Ben Taskar, and Michael~I Jordan.
\newblock Joint covariate selection and joint subspace selection for multiple classification problems.
\newblock \emph{Statistics and Computing}, 20:\penalty0 231--252, 2010.

\bibitem[Parikh et~al.(2014)Parikh, Boyd, et~al.]{parikh2014proximal}
Neal Parikh, Stephen Boyd, et~al.
\newblock Proximal algorithms.
\newblock \emph{Foundations and trends{\textregistered} in Optimization}, 1\penalty0 (3):\penalty0 127--239, 2014.

\bibitem[Paul et~al.(2022)Paul, Jhamb, Mishra, and Kumar]{PAUL2022100218}
Sandip Paul, Bhuvan Jhamb, Deepak Mishra, and M.~Senthil Kumar.
\newblock Edge loss functions for deep-learning depth-map.
\newblock \emph{Machine Learning with Applications}, 7:\penalty0 100218, 2022.
\newblock ISSN 2666-8270.
\newblock \doi{https://doi.org/10.1016/j.mlwa.2021.100218}.

\bibitem[Raghu et~al.(2019)Raghu, Raghu, Bengio, and Vinyals]{raghu2019rapid}
Aniruddh Raghu, Maithra Raghu, Samy Bengio, and Oriol Vinyals.
\newblock Rapid learning or feature reuse? towards understanding the effectiveness of maml.
\newblock \emph{arXiv preprint arXiv:1909.09157}, 2019.

\bibitem[Raymond et~al.(2023{\natexlab{a}})Raymond, Chen, and Xue]{raymond2023learning}
Christian Raymond, Qi~Chen, and Bing Xue.
\newblock Learning symbolic model-agnostic loss functions via meta-learning.
\newblock \emph{IEEE Transactions on Pattern Analysis and Machine Intelligence}, 2023{\natexlab{a}}.

\bibitem[Raymond et~al.(2023{\natexlab{b}})Raymond, Chen, Xue, and Zhang]{raymond2023online}
Christian Raymond, Qi~Chen, Bing Xue, and Mengjie Zhang.
\newblock Online loss function learning.
\newblock \emph{arXiv preprint arXiv:2301.13247}, 2023{\natexlab{b}}.

\bibitem[Ren et~al.(2021)Ren, Xiao, Chang, Huang, Li, Chen, and Wang]{ren2021comprehensive}
Pengzhen Ren, Yun Xiao, Xiaojun Chang, Po-Yao Huang, Zhihui Li, Xiaojiang Chen, and Xin Wang.
\newblock A comprehensive survey of neural architecture search: Challenges and solutions.
\newblock \emph{ACM Computing Surveys (CSUR)}, 54\penalty0 (4):\penalty0 1--34, 2021.

\bibitem[Rong et~al.(2023)Rong, Yu, Zhang, and Ou]{rong2023across}
Jingtao Rong, Xinyi Yu, Mingyang Zhang, and Linlin Ou.
\newblock Across-task neural architecture search via meta learning.
\newblock \emph{International Journal of Machine Learning and Cybernetics}, 14\penalty0 (3):\penalty0 1003--1019, 2023.

\bibitem[Sainath et~al.(2013)Sainath, Kingsbury, Sindhwani, Arisoy, and Ramabhadran]{sainath2013low}
Tara~N Sainath, Brian Kingsbury, Vikas Sindhwani, Ebru Arisoy, and Bhuvana Ramabhadran.
\newblock Low-rank matrix factorization for deep neural network training with high-dimensional output targets.
\newblock In \emph{2013 IEEE international conference on acoustics, speech and signal processing}, pp.\  6655--6659. IEEE, 2013.

\bibitem[Sandjakoska \& Bogdanova(2018)Sandjakoska and Bogdanova]{8587027}
Ljubinka Sandjakoska and Ana~Madevska Bogdanova.
\newblock Towards stochasticity of regularization in deep neural networks.
\newblock In \emph{2018 14th Symposium on Neural Networks and Applications (NEUREL)}, pp.\  1--4, 2018.
\newblock \doi{10.1109/NEUREL.2018.8587027}.

\bibitem[Scardapane et~al.(2017)Scardapane, Comminiello, Hussain, and Uncini]{scardapane2017group}
Simone Scardapane, Danilo Comminiello, Amir Hussain, and Aurelio Uncini.
\newblock Group sparse regularization for deep neural networks.
\newblock \emph{Neurocomputing}, 241:\penalty0 81--89, 2017.

\bibitem[Schmidhuber(1987)]{schmidhuber:1987:srl}
Jurgen Schmidhuber.
\newblock Evolutionary principles in self-referential learning. on learning now to learn: The meta-meta-meta...-hook.
\newblock Diploma thesis, Technische Universitat Munchen, Germany, 14 May 1987.

\bibitem[Schwarz \& Teh(2022)Schwarz and Teh]{schwarz2022metalearning}
Jonathan Schwarz and Yee~Whye Teh.
\newblock Meta-learning sparse compression networks.
\newblock \emph{Transactions on Machine Learning Research}, 2022.
\newblock ISSN 2835-8856.

\bibitem[Shaw et~al.(2019)Shaw, Wei, Liu, Song, and Dai]{shaw2019meta}
Albert Shaw, Wei Wei, Weiyang Liu, Le~Song, and Bo~Dai.
\newblock Meta architecture search.
\newblock \emph{Advances in Neural Information Processing Systems}, 32, 2019.

\bibitem[Shen et~al.(2020)Shen, Chen, Heaton, Chen, Liu, Yin, and Wang]{shen2020learning}
Jiayi Shen, Xiaohan Chen, Howard Heaton, Tianlong Chen, Jialin Liu, Wotao Yin, and Zhangyang Wang.
\newblock Learning a minimax optimizer: A pilot study.
\newblock In \emph{International Conference on Learning Representations}, 2020.

\bibitem[Subramanian et~al.(2023)Subramanian, Ganapathiraman, and Gamal]{subramanian2023learned}
Shreyas Subramanian, Vignesh Ganapathiraman, and Aly~El Gamal.
\newblock {LEARNED} {LEARNING} {RATE} {SCHEDULES} {FOR} {DEEP} {NEURAL} {NETWORK} {TRAINING} {USING} {REINFORCEMENT} {LEARNING}, 2023.
\newblock URL \url{https://openreview.net/forum?id=0Zhwu1VaOs}.

\bibitem[Sun et~al.(2020{\natexlab{a}})Sun, Shao, Li, Liu, Yan, Qiu, and Huang]{sun2020learning}
Tianxiang Sun, Yunfan Shao, Xiaonan Li, Pengfei Liu, Hang Yan, Xipeng Qiu, and Xuanjing Huang.
\newblock Learning sparse sharing architectures for multiple tasks.
\newblock In \emph{Proceedings of the AAAI conference on artificial intelligence}, volume~34, pp.\  8936--8943, 2020{\natexlab{a}}.

\bibitem[Sun et~al.(2020{\natexlab{b}})Sun, Panda, Feris, and Saenko]{sun2020adashare}
Ximeng Sun, Rameswar Panda, Rogerio Feris, and Kate Saenko.
\newblock Adashare: Learning what to share for efficient deep multi-task learning.
\newblock \emph{Advances in Neural Information Processing Systems}, 33:\penalty0 8728--8740, 2020{\natexlab{b}}.

\bibitem[Sun et~al.(2023)Sun, Shen, Yin, Mao, Molchanov, and Alvarez]{regrow1}
Xinglong Sun, Maying Shen, Hongxu Yin, Lei Mao, Pavlo Molchanov, and Jose~M. Alvarez.
\newblock Towards dynamic sparsification by iterative prune-grow lookaheads, 2023.

\bibitem[Thrun \& Pratt(1998)Thrun and Pratt]{Thrun98}
S.~Thrun and L.Y. Pratt (eds.).
\newblock \emph{Learning To Learn}.
\newblock Kluwer Academic Publishers, Boston, MA, 1998.

\bibitem[Thrun \& Pratt(2012)Thrun and Pratt]{thrun2012learning}
Sebastian Thrun and Lorien Pratt.
\newblock \emph{Learning to learn}.
\newblock Springer Science \& Business Media, 2012.

\bibitem[Tian et~al.(2022)Tian, Zhao, and Huang]{TIAN2022203}
Yingjie Tian, Xiaoxi Zhao, and Wei Huang.
\newblock Meta-learning approaches for learning-to-learn in deep learning: A survey.
\newblock \emph{Neurocomputing}, 494:\penalty0 203--223, 2022.
\newblock ISSN 0925-2312.

\bibitem[Upadhyay et~al.(2023{\natexlab{a}})Upadhyay, Chhipa, Phlypo, Saini, and Liwicki]{paper2}
Richa Upadhyay, Prakash~Chandra Chhipa, Ronald Phlypo, Rajkumar Saini, and Marcus Liwicki.
\newblock Multi-task meta learning: learn how to adapt to unseen tasks.
\newblock In \emph{2023 International Joint Conference on Neural Networks (IJCNN)}, pp.\  1--10, 2023{\natexlab{a}}.
\newblock \doi{10.1109/IJCNN54540.2023.10191400}.

\bibitem[Upadhyay et~al.(2023{\natexlab{b}})Upadhyay, Phlypo, Saini, and Liwicki]{paper3}
Richa Upadhyay, Ronald Phlypo, Rajkumar Saini, and Marcus Liwicki.
\newblock Less is more {\textendash} towards parsimonious multi-task models using structured sparsity.
\newblock In \emph{Conference on Parsimony and Learning (Proceedings Track)}, 2023{\natexlab{b}}.

\bibitem[Upadhyay et~al.(2024)Upadhyay, Phlypo, Saini, and Liwicki]{paper1}
Richa Upadhyay, Ronald Phlypo, Rajkumar Saini, and Marcus Liwicki.
\newblock Sharing to learn and learning to share; fitting together meta, multi-task, and transfer learning: A meta review.
\newblock \emph{IEEE Access}, 12:\penalty0 148553--148576, 2024.
\newblock \doi{10.1109/ACCESS.2024.3478805}.

\bibitem[Wang et~al.(2021)Wang, Qin, Zhang, and Fu]{wang2021neural}
Huan Wang, Can Qin, Yulun Zhang, and Yun Fu.
\newblock Neural pruning via growing regularization.
\newblock In \emph{International Conference on Learning Representations}, 2021.

\bibitem[Wen et~al.(2016)Wen, Wu, Wang, Chen, and Li]{wen2016learning}
Wei Wen, Chunpeng Wu, Yandan Wang, Yiran Chen, and Hai Li.
\newblock Learning structured sparsity in deep neural networks.
\newblock \emph{Advances in neural information processing systems}, 29, 2016.

\bibitem[Wichrowska et~al.(2017)Wichrowska, Maheswaranathan, Hoffman, Colmenarejo, Denil, Freitas, and Sohl-Dickstein]{wichrowska2017learned}
Olga Wichrowska, Niru Maheswaranathan, Matthew~W Hoffman, Sergio~Gomez Colmenarejo, Misha Denil, Nando Freitas, and Jascha Sohl-Dickstein.
\newblock Learned optimizers that scale and generalize.
\newblock In \emph{International conference on machine learning}, pp.\  3751--3760. PMLR, 2017.

\bibitem[Wortsman et~al.(2019)Wortsman, Ehsani, Rastegari, Farhadi, and Mottaghi]{wortsman2019learning}
Mitchell Wortsman, Kiana Ehsani, Mohammad Rastegari, Ali Farhadi, and Roozbeh Mottaghi.
\newblock Learning to learn how to learn: Self-adaptive visual navigation using meta-learning.
\newblock In \emph{Proceedings of the IEEE/CVF conference on computer vision and pattern recognition}, pp.\  6750--6759, 2019.

\bibitem[Xiong et~al.(2022)Xiong, Lan, Chen, Wang, and Hsieh]{xiong2022learning}
Yuanhao Xiong, Li-Cheng Lan, Xiangning Chen, Ruochen Wang, and Cho-Jui Hsieh.
\newblock Learning to schedule learning rate with graph neural networks.
\newblock In \emph{International Conference on Learning Representation (ICLR)}, 2022.

\bibitem[Yu et~al.(2017)Yu, Koltun, and Funkhouser]{yu2017dilated}
Fisher Yu, Vladlen Koltun, and Thomas Funkhouser.
\newblock Dilated residual networks.
\newblock In \emph{IEEE Conference on Computer Vision and Pattern Recognition (CVPR)}, 2017.

\bibitem[Yuan \& Lin(2005)Yuan and Lin]{groupLasso}
Ming Yuan and Yi~Lin.
\newblock {Model Selection and Estimation in Regression with Grouped Variables}.
\newblock \emph{Journal of the Royal Statistical Society Series B: Statistical Methodology}, 68\penalty0 (1):\penalty0 49--67, 12 2005.
\newblock ISSN 1369-7412.
\newblock \doi{10.1111/j.1467-9868.2005.00532.x}.

\bibitem[Zhang et~al.(2024)Zhang, Yue, Wang, Fang, Sui, Wang, Liang, Cheng, Pan, and Chen]{regrow2}
Guibin Zhang, Yanwei Yue, Kun Wang, Junfeng Fang, Yongduo Sui, Kai Wang, Yuxuan Liang, Dawei Cheng, Shirui Pan, and Tianlong Chen.
\newblock Two heads are better than one: Boosting graph sparse training via semantic and topological awareness, 2024.

\bibitem[Zhou et~al.(2021{\natexlab{a}})Zhou, Ma, Zhu, Liu, Zhang, Yuan, Sun, and Li]{zhou2021learning}
Aojun Zhou, Yukun Ma, Junnan Zhu, Jianbo Liu, Zhijie Zhang, Kun Yuan, Wenxiu Sun, and Hongsheng Li.
\newblock Learning n: m fine-grained structured sparse neural networks from scratch.
\newblock \emph{arXiv preprint arXiv:2102.04010}, 2021{\natexlab{a}}.

\bibitem[Zhou et~al.(2021{\natexlab{b}})Zhou, Zhang, Xu, and Zhang]{zhou2021effective}
Xiao Zhou, Weizhong Zhang, Hang Xu, and Tong Zhang.
\newblock Effective sparsification of neural networks with global sparsity constraint.
\newblock In \emph{Proceedings of the IEEE/CVF Conference on Computer Vision and Pattern Recognition}, pp.\  3599--3608, 2021{\natexlab{b}}.

\bibitem[Zhu \& Gupta(2017)Zhu and Gupta]{zhu2017prune}
Michael Zhu and Suyog Gupta.
\newblock To prune, or not to prune: exploring the efficacy of pruning for model compression.
\newblock \emph{arXiv preprint arXiv:1710.01878}, 2017.

\end{thebibliography}
